\documentclass[conference,compsoc]{IEEEtran}

\usepackage{hyperref}
\usepackage{amssymb}
\usepackage{bm}
\usepackage{graphicx}
\usepackage{amsmath}
\usepackage{booktabs}
\usepackage{makecell}
\usepackage{threeparttable}
\usepackage{algorithm}
\usepackage{algorithmic}
\usepackage{subfig}
\usepackage{xcolor}
\definecolor{color1}{RGB}{255,92,92}
\definecolor{color2}{RGB}{64,218,255}
\definecolor{color3}{RGB}{255,221,20}
\definecolor{color4}{RGB}{150,200,150}

\newtheorem{definition}{Definition}[section]

\begin{document}

\title{How to Defend Against Large-scale Model Poisoning Attacks in Federated Learning: A Vertical Solution}

\author{\IEEEauthorblockN{Jinbo Wang}
\IEEEauthorblockA{University of Electronic \\Science and Technology of China\\
Chengdu, China 610000\\
Email: 1694300437@qq.com}
\and

\IEEEauthorblockN{Ruijin Wang* \\ Corresponding author}
\IEEEauthorblockA{University of Electronic \\Science and Technology of China\\
Chengdu, China 610000\\
Email: ruijinwang@uestc.edu.cn}
\and

\IEEEauthorblockN{Fengli Zhang}
\IEEEauthorblockA{University of Electronic \\Science and Technology of China\\
Chengdu, China 610000\\
Email: fzhang@uestc.edu.cn}
}

\maketitle

\begin{abstract}
	
Federated learning (FL) is vulnerable to model poisoning attacks due to its distributed nature. The current defenses start from all user gradients (model updates) in each communication round and solve for the optimal aggregation gradients (horizontal solution). This horizontal solution will completely fail when facing large-scale ($>$50\%) model poisoning attacks. In this work, based on the key insight that the convergence process of the model is a highly predictable process, we break away from the traditional horizontal solution of defense and innovatively transform the problem of solving the optimal aggregation gradients into a vertical solution problem. We propose VERT, which uses global communication rounds as the vertical axis, trains a predictor using historical gradients information to predict user gradients, and compares the similarity with actual user gradients to precisely and efficiently select the optimal aggregation gradients. In order to reduce the computational complexity of VERT, we design a low dimensional vector projector to project the user gradients to a computationally acceptable length, and then perform subsequent predictor training and prediction tasks. Exhaustive experiments show that VERT is efficient and scalable, exhibiting excellent large-scale ($\geq$80\%) model poisoning defense effects under different FL scenarios. In addition, we can design projector with different structures for different model structures to adapt to aggregation servers with different computing power.

\end{abstract}


\IEEEpeerreviewmaketitle

\section{Introduction}

Federated Learning (FL)~\cite{1} is a novel distributed learning paradigm that enables users from different geographical locations to collaboratively train a global neural network model with better performance and generalization ability without leaking local privacy data. Specifically, FL systems generally include an aggregation server and multiple participating users. Firstly, the aggregation server distributes the global model to all participating users; Secondly, users use local data for model training and upload gradients to the aggregation server; Finally, the aggregation server performs weighted aggregation of all user gradients based on the gradient aggregation function (federated average algorithm, FedAvg~\cite{1}). FL achieves knowledge transfer and aggregation in this way without leaking user privacy data~\cite{2,3,4}.

Although the distributed nature of FL has natural advantages in privacy protection, it also opens the door for attackers' poisoning attacks~\cite{5,6,7,8}. The poisoning attacks in FL include data poisoning~\cite{5,6} and model poisoning~\cite{7,8}. In data poisoning, attackers manipulate training data to poison models or implant backdoors~\cite{9,10}. And model poisoning is a more powerful attack, where attackers can skip manipulating training data and directly modify model parameters to maximize the poisoning of the global model. In this work, we will mainly consider model poisoning attacks. Imagine a simple example where a FL system has only two participating users, and one of them is a compromised user controlled by an attacker. The compromised user only needs to upload malicious gradients in the opposite direction to the honest gradient~\cite{11}, which can perfectly interfere with the convergence of the global model.

Actually, the number of users participating in the FL system is usually large, which makes it possible for attackers to control large-scale compromised users and conducting model poisoning attacks at the same time. The existing defenses~\cite{12,13,14,15,16,17} typically start from a horizontal solution thinking and attempt to find a set of optimal aggregation gradients (honest gradients for aggregation). The so-called horizontal solution (Definition~\ref{definition4.1}) is to solve the optimal aggregation gradients based on the gradients of all users in the current communication round. For example, Krum~\cite{12} calculates the sum of Euclidean distances between each user gradient and other gradients, and selects the user gradient with the smallest distance as the optimal aggregation gradient. Median~\cite{13} utilizes all users gradients to calculate the median for each gradient element, and finally combines it into the optimal aggregation gradient. However, the horizontal solution is only applicable to a small number of compromised users ($\leq$50\%)~\cite{12}. When the proportion of compromised users exceeds 50\%, malicious gradients will dominate the calculation of the optimal aggregation gradient (malicious gradients dominance problem (Definition~\ref{definition4.2})). FLTrust~\cite{18} solves the malicious gradients dominance problem by introducing a trust root into the aggregation server. However, it requires the aggregation server to have a completely clean dataset, and the data distribution should be consistent with the user's local data distribution, which is not realistic in the real world.

In this work, we attempt to bypass the malicious gradients dominance problem caused by large-scale model poisoning attacks, in order to solve the optimal aggregation gradient in large-scale model poisoning attacks. Our key insight is that the training and optimization process of model parameters is a regular and predictable convergence process. Similarly, in FL system, honest users upload predictable gradients in each communication round~\cite{19,20}, while compromised users upload irregular and unpredictable malicious gradients. In view of this, we break the traditional horizontal solution and innovatively propose a vertical solution (Definition~\ref{definition4.3}) with communication rounds as the vertical axis, in order to bypass malicious gradients dominance problem. We propose VERT, which is a neural network autoregressive method~\cite{21,22,23} based on historical gradient information to predict user gradients. Specifically, we believe that the calculation of user gradients depends on the distribution of local data and the global model issued by the aggregation server. Therefore, we use historical user gradients and historical global gradients as training data for the autoregressive model (predictor) to predict user gradients. Finally, the prediction gradient will be compared with the actual user gradient for similarity, and the user gradient with high similarity will be selected as the optimal aggregation gradient. Considering that existing model architectures are generally complex, typically containing tens of millions of parameters. In order to reduce the computational complexity of the predictor, we additionally designed a low dimensional vector projector to project the gradients onto the low dimensional feature space, and then proceed with the subsequent predictor training and prediction process. This projector is scalable and can be designed with different output dimensions for aggregation servers with different computing power.

Due to the vertical solution perspective of VERT, the optimal aggregation gradients is solved based on historical user gradients and historical global gradients, without involving gradients of other compromised users. Therefore, even under large-scale ($\geq$80\%) model poisoning attacks, VERT can still solve for the optimal aggregation gradient without being affected by malicious gradients dominance problem. Due to the above natures of VERT, we also find that it can still effectively against large-scale poisoning attacks in non independent and identically distributed (non-IID) user data. In addition, VERT does not require any unrealistic assumptions (including the attacker's background knowledge, capabilities, number of compromised users, and special abilities of the aggregation server), and we only need to deploy it on the aggregation server to achieve robustness against large-scale model poisoning attacks. Below, we summarize our contributions:

\begin{itemize}
	\item Based on the key insight that the optimization process of model parameters is a highly predictable process, we believe that the gradient of participating users in the FL system is also predictable, and demonstrates this through experiments.
	
	\item We break away from the traditional horizontal solution thinking and propose VERT, a large-scale model poisoning defense that uses communication rounds as the vertical axis and solves optimal aggregation gradients based on historical gradients information.
	
	\item We introduce a neural network-based autoregressive model (predictor) to learn the historical user gradients changes and predict user gradient. Simultaneously design a low dimensional vector projector to project gradients onto a low dimensional feature space, reducing the computational complexity of the predictor.
	
	\item We implement various experimental scenarios on three visual datasets, including different data distributions (IID and non-IID), multiple model poisoning attacks, and different neural network models. The results indicate that VERT can effectively defend against large-scale model poisoning attacks in various experimental scenarios.
	
	\item We release the implementation \href{https://github.com/bo-lab520/VERT}{code} of VERT for readers to replicate the experimental results.
	
\end{itemize}

\section{Background}

This section first describes the technical background of VERT, including FL and model poisoning attacks, and defines the relevant characters and their meanings. Secondly, explains the predictability of user gradients in FL systems.

\subsection{Federated Learning and Model poisoning Attacks}

We are considering a typical FL system, which includes an aggregation server $S$ and a set of participating users $C$. Assuming the number of participating users is $K$, there is $K=|C|$. Each participating user $k$ has a local dataset $D_k=\{x_i^k, y_i^k\}_{i=1}^{N_k}$, where $N_k$ is the amount of data $N_k=|D_k|$. The purpose of FL system is to transfer and aggregate knowledge by aggregating users local models, ultimately forming an optimal global model with minimal loss on local data of all participating users. Assuming the loss function of the participating users is $l$, the purpose of the aggregation server is to calculate the optimal global model parameter $\bm{w^*}=arg\min\limits_{\bm{w}}\mathcal{L}(\bm{w})=\sum_{k=1}^{K}p_k\mathbb{E}_{x_i^k\in D_k}l(\bm{w},(x_i^k,y_i^k))$, where $p_k$ is the probability of user $k$ participating in FL, $\sum_{k=1}^{K}p_k=1$, $\mathbb{E}$ is the expected function, $\bm{w}\in \mathbb{R}^d$.

Now, let's elaborate on the operation of FL system. Assuming a total of $T$  communication rounds between the aggregation server and the participating users. Without loss of generality, we choose the $t$-th ($1\leq t\leq T$) round. (1) The aggregation server selects some users to participate in $t$-th round of training according to a certain strategy (random or targeted selection), and the selected users set is denoted as $C_t$. The global model $\bm{w^t}$ is issued to $C_t$. (2) The selected user $k\in C_t$ minimizes the local loss to optimize the local model $\bm{w_k^t}=arg\min\limits_{\bm{w_k^t}}\mathbb{E}_{x_i^k\in D_k}l(\bm{w_k^t},(x_i^k,y_i^k))$, calculates the user gradient $\bm{g_k^t}=\bm{w_k^}t-\bm{w^t}$, and uploads $\bm{g_k^t}$ to the aggregation server, where $\bm{g_k^t}\in \mathbb{R}^d$. (3) The aggregation server calculates the global gradient according to the aggregation function (FedAvg) $\bm{g^t}=\frac{1}{|C_t|}\sum_{k\in C_t}^{}\bm{g_k^t}$, updates the $t+1$-th round of global model $\bm{w^{t+1}}=\bm{w^t}+\eta \bm{g^t}$, where $\eta$ is the global learning rate. The above steps (1)-(3) will be iteratively executed in $T$ rounds to ultimately obtain the global model $\bm{w^T}$.

Due to the large number of participating users, FL systems often face model poisoning attacks. This work mainly considers the situation of extreme large-scale model poisoning attacks. Assuming an attacker compromised multiple participating users, the compromised users set is $C_{poi}$ ($|C_{poi}|/K>50\%$). The attacker's goal is to make the global model unavailable by uploading large-scale malicious gradients, i.e. maximizing the loss $max\mathcal{L}(\bm{w})$. Taking $t$-th round as an example, the $C_t$ selected by the aggregation server includes both honest users and compromised users. Honest users upload $\bm{g_k^t}$ normally, while compromised users upload malicious gradients $\bm{\hat{g}_k^t}$ (randomly generated or other poisoning attacks). At this point, the global gradient is $\bm{g^t}=\frac{1}{|C_t|}\sum_{k\in C_t}^{}\bm{g_k^t}$, where $\bm{g_k^t}$ as follows.

\begin{equation}
	\bm{g_k^t}=
	\left\{
	\begin{array}{ll}
		\bm{g_k^t}, k\notin C_{poi} \\
		\\
		\bm{\hat{g}_k^t}, k\in C_{poi}
	\end{array}.
	\right.
\end{equation}

Due to the presence of large-scale malicious gradients, $\bm{g^t}$ will deviate from the correct convergence direction, resulting in the unavailability of the global model.

\subsection{The predictability of user gradients in FL system}

Neural networks tend to converge after multiple rounds of iterative training on the same dataset, which has been theoretically confirmed by many works~\cite{24,25,26}. This means that the model parameters are always updated towards the optimal point (minimizing loss) during the training process, that is, the parameter changes follow a certain pattern. We extend our thinking to consider the predictability of user gradients in FL systems. Participating users always update the local model towards the local optimal point, so the direction of user gradients is also predictable. Assuming that the user gradient of user $k$ in the $t$-th round is $\bm{g_k^t}=[g_{k,1}^t,g_{k,2}^t,...,g_{k,d}^t]$, and $d$ is the number of the global model parameters. We believe that the sequence $\{g_{k,j}^1,g_{k,j}^2,...,g_{k,j}^T\}$ of each gradient element $g_{k,j}^t$ ($1\leq j\leq d$) in all $T$ rounds is predictable. On the other hand, malicious gradients uploaded by compromised users lack predictability.

Assuming we have an autoregressive model $f_{AR}$ and use the user gradients from $m$ rounds of history as training data to predict the user gradient in $t$-th round. We use cosine function to measure the error between the prediction gradient and the actual gradient $cos(f_{AR}(\{\bm{g_k^{t-m-1}},...,\bm{g_k^{t-1}}\}),\bm{g_k^t})$. The prediction error for compromised users is much higher than honest users. The specific experimental results can be found in Section~\ref{section6.2.1}. In addition, \cite{19} also demonstrated through average time-delayed mutual information that the predictability of honest gradients is much higher than malicious gradients.

\section{Problem Formulation}

This section introduces the optimization objective of VERT and defines the threat model of this paper.

\subsection{Optimization Objective}

The purpose of this paper is to design a solution that can select the optimal aggregation gradients under large-scale model poisoning attacks. Assuming that the users set to which the selected optimal aggregation gradients belong is $C_{opt}^t$, $C_{opt}^t\in C_t$, we need to achieve the following optimization objectives:

\begin{equation}\label{equation2}
	\left\{
	\begin{array}{ll}
		max|C_{opt}^t\cap (C_t-C_{poi})| \\
		\\
		min|C_{opt}^t\cap C_{poi}|
	\end{array}
	\right.
	, s.t. |C_{opt}^t|=\kappa.
\end{equation}

$\kappa$ represents the number of optimal aggregation gradients selected in each round. The above optimization objective aims to maximize the participation of honest users in gradient aggregation while avoiding compromised users participation, to minimize the loss $min\mathcal{L}(\bm{w})$.

\subsection{Threat Model}

We mainly consider the model poisoning attacks faced by FL systems, and attacker can simultaneously compromise over 50\% of participating users for large-scale model poisoning attacks. The following explains the threat model of this paper from three perspectives: the attacker's goal, knowledge, and capability.

\textbf{Attacker's goal:} Model poisoning attacks can be further divided into untargeted attacks~\cite{14, 27} and targeted attacks (backdoor attacks)~\cite{9,10,28,29,30}. When the attacker executes untargeted attacks, his goal is to interfere with the global model convergence, making the model unable to accurately predict on all categories. When executes targeted attack, the attacker expects to embed a pre-defined trigger into the global model with a fixed mapping to the target label. During the testing phase, any data that patched the pre-defined trigger will be predicted as the target label.

\textbf{Attacker's knowledge:} Consistent with previous work~\cite{19,20}, we assume that the attacker has all the knowledge about compromised users, including training data, local model, and training code. This means that the attacker can freely generate their desired malicious gradients. In addition, in order to conduct more powerful attacks, we assume that the attacker knows the total number of participating users, even knows the user gradient of other honest users.

\textbf{Attacker's capability:} In a FL system, the attacker can inject a large number of malicious users  into the system or compromise a large number of participating users, and can freely control these users to conduct model poisoning attacks. Due to the attacker has all the knowledge about compromised users, they can freely upload their desired malicious gradients or generate backdoor gradients by distort training data.

\section{Proposed Method}

This section mainly introduces the design concept of VERT. Firstly, from the perspective of vertical solution, we rethink the defense against large-scale model poisoning attacks in FL systems. Secondly, provide a detailed introduction to the workflow of VERT.

\subsection{Rethinking the Defense Against Large-scale Model Poisoning Attacks in FL Systems from a Vertical Perspective}\label{section4.1}

The distributed nature of FL system makes it possible for attackers to control large-scale compromised users for model poisoning attacks. In order to prevent poisoning attacks from affecting the convergence of the global model, the aggregation server needs to solve for the optimal aggregation gradients before performing gradients aggregation. Most of the existing defenses belong to the horizontal solution (Definition~\ref{definition4.1}), such as Krum~\cite{12}, Median~\cite{13}, etc.

\begin{definition}[Horizontal solution]\label{definition4.1}
	Solving the optimal aggregation gradients based on the gradients of all users in the current round. If the horizontal solution is denoted as $f_{horizontal}$, then the optimal aggregation gradients set for the $t$-th round is $C_{opt}^t=f_{horizontal}(\{\bm{g_k^t}|k\in C_t\})$.
\end{definition}

However, a basic premise of the horizontal solution is that the proportion of malicious gradients is less than 50\%~\cite{12}. When facing large-scale model poisoning attacks, the horizontal solution will lose its effectiveness due to the malicious gradients dominance problem (Definition~\ref{definition4.2}).

\begin{definition}[Malicious gradients dominance problem]\label{definition4.2}
	When the proportion of malicious gradients in the total gradients is too high, the optimal aggregation gradients in the horizontal solution will rely too much on malicious gradients, resulting in errors in the solution.
\end{definition}

\begin{figure}[h]
	\centering
	\vspace{-2mm}
	\hspace{-0mm}
	\setlength{\abovecaptionskip}{-2mm}
	\includegraphics[width=90mm]{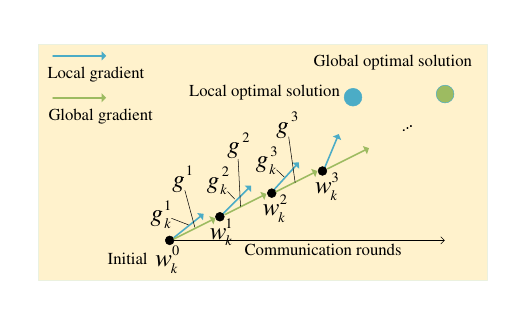}
	\caption{The direction of user gradients in all communication rounds in the FL system.}
	\label{figure1}
\end{figure}

The malicious gradients dominance problem is unavoidable, however, we can bypass this peoblem from a vertical perspective. Based on the key insight that the convergence process of neural networks is a highly predictable process, we believe that the user gradients in FL system can also be predicted, as shown in Figure~\ref{figure1}. The direction of user gradients depends on two key points. Firstly, the user gradients always updates the local model towards the local optimal solution, as the user trains the local model based on the same local data. Secondly, the user gradients is calculated based on the previous round's global model, and the global model is calculated by the global gradient, which always updates the global model towards the global optimal solution. Therefore, we can reasonably assume that the user gradients $g_k^t$ in the $t$-th round can be inferred from $\bm{g_k^{t-1}}$ and $\bm{g^{t-1}}$. If the inference function is $f_{infer}$, then there is $\bm{g_k^t}=f_{infer}(\bm{g_k^{t-1}},\bm{g^{t-1}})$.

Therefore, we believe that the vertical solution (Definition~\ref{definition4.3}) based on historical gradients information, with communication rounds as the vertical axis, can effectively handle large-scale model poisoning attacks, as the vertical solution is not constrained by other malicious gradients.

\begin{definition}[Vertical solution]\label{definition4.3}
	Calculate the probability of thecurrent round  user gradient is honest gradient based on the user gradients of the historical $m$ rounds, and ultimately solve for the optimal aggregation gradients. If the vertical solution is denoted as $f_{vertical}$, then the optimal aggregate gradients set for the $t$-th round is $C_{opt}^t=top-\kappa (f_{vertical}(\bm{g_k^t},\{\bm{g_k^{t-m}},...,\bm{g_k^{t-1}}\})|k\in C^t\})$, where g can obtain the k gradients with the highest probability in the set.
\end{definition}

\subsection{Vertical Solution: VERT}

\begin{figure*}[h]
	\centering
	\vspace{-0mm}
	\hspace{-0mm}
	\setlength{\abovecaptionskip}{-0mm}
	\includegraphics[width=180mm]{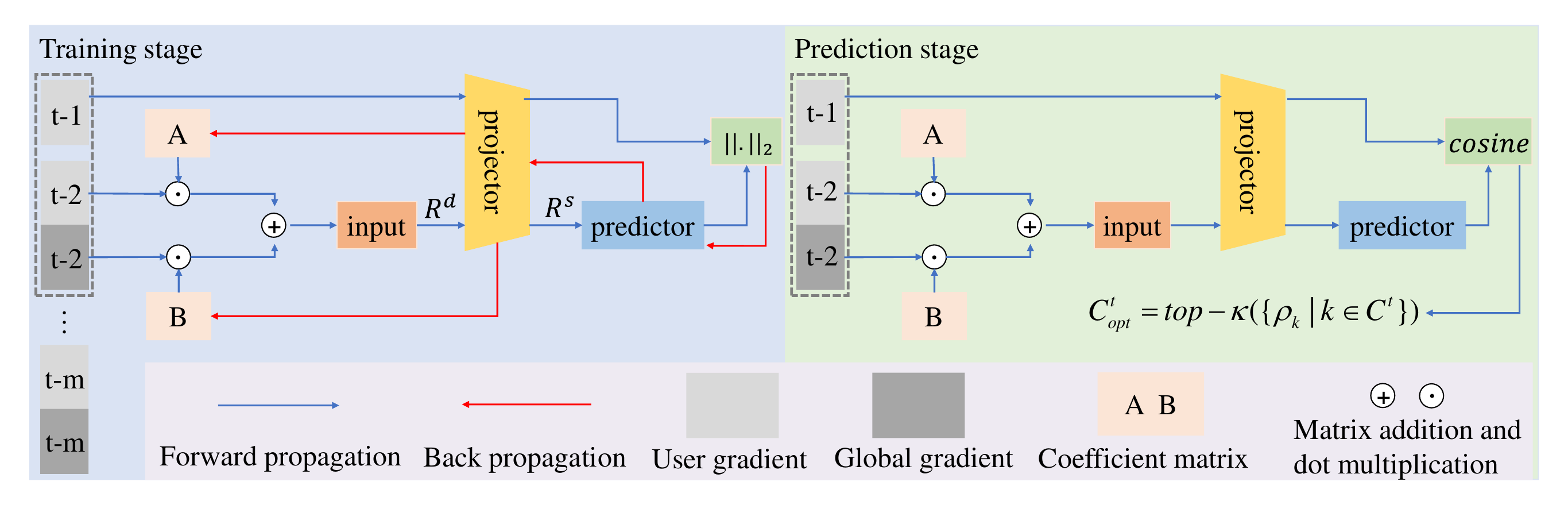}
	\caption{VERT framework. During the training phase, VERT utilizes historical gradients information to train the coefficient matrix, projector, and predictor, enabling the predictor to predict user gradients precisely. In the prediction stage, VERT inputs the gradient information from the previous round into the projector and predictor, calculates the cosine similarity between the prediction gradient and the actual gradient.}
	\label{figure2}
\end{figure*}

In order to give readers a preliminary understanding of VERT, we provide a framework diagram of VERT in the training and prediction stages, as shown in Figure~\ref{figure2}. Section~\ref{section4.2.1} to~\ref{section4.2.3} provide more detailed explanations.

\subsubsection{Integration of User Gradient and Global Gradient}\label{section4.2.1}

In Section~\ref{section4.1}, we have already explained that the user gradient $\bm{g_k^t}$ in $t$-th round can be inferred from $\bm{g_k^{t-1}}$ and $\bm{g^{t-1}}$, i.e., $\bm{g_k^t}=f_{infer}(\bm{g_k^{t-1}},\bm{g^{t-1}})$. However, directly calculating $f_{infer}$ is not feasible. Therefore, we design a predictor $f_{pred}$ based on a fully connected neural network layer, and train $f_{pred}$ using historical gradients information and autoregressive training to make $\bm{\tilde{g}_k^t}=f_{pred}(\bm{g_k^{t-1},\bm{g^{t-1}}})\approx f_{infer}(\bm{g_k^{t-1}},\bm{g^{t-1}})$.

The input of the predictor $f_{pred}$ mentioned above has two gradients, while general inference based on neural networks only has one input. A simple integration method for this is to directly add $\bm{g_k^{t-1}}$ and $\bm{g^{t-1}}$ proportionally $\bm{g_{input}^{t-1}}=\alpha \bm{g_k^{t-1}}+\beta \bm{g^{t-1}}$, or concatenate them together $\bm{g_{input}^{t-1}}=cat(\bm{g_k^{t-1}},\bm{g^{t-1}})$. However, these integration methods cannot dynamically adjust the relative importance between two gradients. Therefore, we propose an optimizable coefficient matrix $\bm{A}\in \mathbb{R}^d$, $\bm{B}\in \mathbb{R}^d$. The integration method is as follows:

\begin{equation}
	\bm{g_{input}^{t-1}}=\bm{A}\odot \bm{g_k^{t-1}}+\bm{B}\odot \bm{g^{t-1}}.
\end{equation}

where $\odot$ represents the multiplication of vector corresponding terms, to dynamically activate gradient elements. In the code, we set the $\mathtt{requires\_grad}$ of the coefficient matrix $\bm{A}$ and $\bm{B}$ to True, so that during the training process of $f_{pred}$, the gradients will be backpropagated to $\bm{A}$ and $\bm{B}$ to achieve an optimizable coefficient matrix.

\subsubsection{Prediction of User Gradient}\label{section4.2.2}

\textit{Training predictor:} Last Section~\ref{section4.2.1} solved the input problem of predictor $f_{pred}$, and this sub-section introduces how to train $f_{pred}$ using historical gradients information. In order to make $f_{pred}$ more accurate in predicting the user gradient $\bm{\hat{g}_k^{t}}$ of $t$-th round, we need to first train $f_{pred}$ using all historical user gradients and global gradients. For any historical round $t_{his}$, the feature data for training $f_{pred}$ are $\bm{g_k^{t_{his}}}$ and $\bm{g^{t_{his}}}$, and the corresponding label is the user gradient $\bm{g_k^{t_{his}+1}}$ in the next round $t_{his}+1$. The optimization objective we need to achieve is to minimize the distance between the prediction gradient and the actual gradient, that is, to minimize the loss $||f_{pred}(\bm{A}\odot \bm{g_k^{t_{his}}}+\bm{B}\odot \bm{g^{t_{his}}})-\bm{g_k^{t_{his}+1}}||_2$. When training using all historical gradients information, the optimization objective is as follows:

\begin{equation}
	min\sum_{t_{his}=1}^{t-2}||f_{pred}(\bm{A}\odot \bm{g_k^{t_{his}}}+\bm{B}\odot \bm{g^{t_{his}}})-\bm{g_k^{t_{his}+1}}||_2.
\end{equation}

Considering the large number of global communication rounds and the limited predictive power of too remote historical gradients information on $f_{pred}$, we only use the gradients information from the historical $m$ round for the $t$-th round of $f_{pred}$ training. The optimization objective can be rewritten as:

\begin{equation}
	min\sum_{t_{his}=t-m}^{t-2}||f_{pred}(\bm{A}\odot \bm{g_k^{t_{his}}}+\bm{B}\odot \bm{g^{t_{his}}})-\bm{g_k^{t_{his}+1}}||_2.
\end{equation}

\textit{Predicting and solving the optimal aggregation gradients:} After completing the training of $f_{pred}$, we need to use it for the $t$-th round of user gradient prediction. The higher similarity between the prediction gradient and the actual gradient, the less possibility the user gradient is to be malicious. The prediction process also requires integrating $\bm{g_k^{t-1}}$ and $\bm{g^{t-1}}$ first, and then inputting it into $f_{pred}$. Therefore, the prediction gradient for user $k$ in the $t$-th round is $\bm{\tilde{g}_k^{t}}=f_{pred}(\bm{A}\odot \bm{g_k^{t-1}}+\bm{B} \odot \bm{g^{t-1}})$. Subsequently, based on cosine similarity, we calculate the similarity between the prediction gradient and the actual gradient $\rho_k=cos(\bm{\tilde{g}_k^{t}}, \bm{g_k^{t}})=\frac{\bm{\tilde{g}_k^{t}}\odot \bm{g_k^{t}}}{||\bm{\tilde{g}_k^{t}}||_2*||\bm{g_k^{t}}||_2}$. We make the above prediction for all participating users. Calculating their similarity, and ultimately select the $\kappa$ with the highest similarity as the optimal aggregation gradients set $C_{opt}^t$, i.e., $C_{opt}=top-\kappa(\{\rho_k|k\in C^t\})$, to achieve the optimization objective as shown in Formula~\ref{equation2}. 

For the subsequent aggregation of the optimal aggregation gradients, we can use typical FedAvg or other horizontal aggregation methods with poisoning robustness as successive aggregation method, such as Krum~\cite{12}, Median~\cite{13}.

\textit{Exception handling:} The selected user set $C_t$ in the $t$-th round of the FL system is generally a subset of all participating user sets $C$. Therefore, the user selected in round $t-1$-th may not exist in $t$-th round.
In order to better train the predictor, we replace the gradient of user $k$ ($k\in C_t$ and $k\notin C_{t-1}$) in $t-1$-th round with $\bm{g^{t-1}}$, and the prediction gradient of user $k$ in $t$-th round is $f_{pred}(\bm{A}\odot \bm{g^{t-1}}+\bm{B}\odot \bm{g^{t-1}})$. In addition, we also use global gradient to replace user gradients that are considered malicious gradients in the $t$-th round, avoiding interference from malicious gradients in the training of matrix coefficients $\bm{A}$, $\bm{B}$, and predictor $f_{pred}$.

\subsubsection{Reduce Memory Overhead: Low Dimensional Vector Projector}\label{section4.2.3}

In order for the output of the predictor to be able to calculate the $l_2$ norm distance from the actual gradient, it is necessary to maintain consistency between the input and output dimensions of the predictor, both of which are $d$. To better handle complex datasets and tasks~\cite{31,32}, current neural network architectures are typically deep and have large model parameters. For example, ResNet110 and ResNet116~\cite{33} have over 100 layers depths, while ResNet18 has over 10 million parameters. Taking ResNet18 as an example, if a simplest predictor with only one hidden layer is designed, the number of parameters would also exceed $10^{8}$. Assuming a parameter is represented by a floating point number (occupying 4 bytes), such a predictor would require 400,000GB of memory. For an aggregation server, maintaining such a predictor is not realistic.

To reduce the additional costs brought by VERT to the aggregation server, we propose a low dimensional vector projector, which is a linear projector designed based on a single-layer fully connected neural network and can project user gradients to low dimensional feature space. The projector is denoted as $f_{proj}$, with $\bm{p_k^t}=f_{proj}(\bm{g_k^t})$, where $\bm{p_k^t}\in \mathbb{R}^s$, $s\ll d$. Before training and predicting of the predictor, we need to project the gradient onto the $\mathbb{R}^s$ space using $f_{proj}$. Therefore, the optimization objective for training the predictor can be further updated to:

\begin{align}\label{formula6}
	\begin{split}
			min\sum_{t_{his}=t-m}^{t-2}||f_{pred}(f_{proj}(\bm{A}\odot \bm{g_k^{t_{his}}}+\bm{B}\odot \bm{g^{t_{his}}}))-\\
		f_{proj}(\bm{g_k^{t_{his}+1}})||_2.
	\end{split}
\end{align}

In the prediction stage, the prediction feature is $\bm{\tilde{p}_k^t}=f_{pred}(f_{proj}(\bm{A}\odot \bm{g_k^{t_{his}}}+\bm{B}\odot \bm{g^{t_{his}}}))$, and the similarity between the prediction feature and the actual feature is $\rho_k=cos(\bm{\tilde{p}_k^t}, f_{proj}(\bm{g_k^t}))$. Based on this similarity, select the optimal aggregation gradients. It should be noted that $f_{proj}$ does not need to be trained simultaneously with the predictor. In VERT, only the coefficient matrix and predictor need to be trained, and the purpose of $f_{proj}$ is only to project the user gradients onto a low dimensional feature space.

At this point, the overall of VERT has been introduced, and Appendix~\ref{appendixa} provides more detailed Pseudocode~\ref{algoritnm1}.

\section{Theoretical Analysis}\label{section5}

This section first analyzes the computational complexity of VERT theoretically, and then analyzes the optimization problem of Formula~\ref{formula6} when training the coefficient matrix and predictor.

\subsection{Computational Complexity Analysis}

The additional computational costs brought by VERT to the FL system mainly comes from the training and prediction of the coefficient matrix, projector, and predictor. For ease of calculation, we assume that both the projector and predictor are a fully connected neural network layer.

In the training stage, for a user, in the forward propagation phase, the coefficient matrix is used to integrate the user gradient and global gradient costs $2O(d)$, the projector is used to reduce the dimensionality of the gradient costs $2O(ds)$, the predictor is used to predict the feature costs $O(s^2)$, and finally the the $l_2$ norm loss costs $O(s)$. Assuming that VERT uses historical $m$ rounds gradients for training, the total costs is $m(2O(d)+2O(ds)+O(s^2))$. In the backpropagation phase, in addition to the weights of the projector, predictor, and coefficient matrix, it is necessary to differentiate the biases of the projector and predictor, with a total costs of $2O(d)+O(ds)+O(s^2)+2O(s)$. In the model update phase, we only need to update the parameters of the coefficient matrix and predictor without updating the projector, with a total costs of $2O(d)+O(s^2)+O(s)$. Therefore, for all selected users $C_t$ in $t$-th round, the total costs is $|C_t|((2m+4)O(d)+(2m+1)O(ds)+(m+2)O(s^2)+3O(s))$. In the prediction stage, without involving backward propagation, the total costs is $|C_t|(2O(d)+2O(ds)+O(s^2))$. In the end, VERT incurs an additional costs of $|C_t|((2m+6)O(d)+(2m+3)O(ds)+(m+3)O(s^2)+3O(s))$ in $t$-th round. Generally, the gradient dimension $d$ is much larger than other parameters, so the above equation mainly depends on $d$.

\subsection{Theoretical Analysis on Coefficient Matrix and Predictor}

We design VERT to train coefficient matrices and predictor for each participating user in the FL system using historical gradients information, in order to achieve accurate prediction of user gradients. Therefore, assuming the optimization objective is $\Phi(\bm{A};\bm{B};f_{pred};f_{proj})=\sum_{t_{his}=t-m}^{t-2}||f_{pred}(f_{proj}(\bm{A}\odot \bm{g_k^{t_{his}}}+\bm{B}\odot \bm{g^{t_{his}}}))-f_{proj}(\bm{g_k^{t_{his}+1}})||_2$, the optimal coefficient matrixs and predictor should meet the following optimization objective:

\begin{align}\label{formula7}
	\begin{split}
		\bm{A},\bm{B},f_{pred}=arg\min\limits_{\bm{A},\bm{B},f_{pred}}\Phi(\bm{A};\bm{B};f_{pred};f_{proj}).
	\end{split}
\end{align}

To calculate the optimal coefficient matrixs and predictor, we need to calculate the partial derivatives of the optimization objective on the $\bm{A}$, $\bm{B}$ and $f_{pred}$. When the partial derivatives are 0, the minimum value of the optimization objective can be achieved. Taking $\bm{A}$ as an example, there are:

\begin{align}
	\begin{split}
		\frac{\partial \Phi(\bm{A};\bm{B};f_{pred};f_{proj})}{\partial \bm{A}}=0.
	\end{split}
\end{align}

Finally, when $m$=2, we obtain:

\begin{align}
	\begin{split}
		\bm{A}=((\bm{W_{pred}}\bm{W_{proj}})^{-1}\bm{W_{proj}g_k^{t_{his}+1}}-\\
		\bm{B}\odot \bm{g^{t_{his}}})/\bm{g_k^{t_{his}}}. 
	\end{split}
\end{align}

The specific calculation process can be found in Appendix~\ref{appendixb}. Similarly, we can calculate:

\begin{align}
	\begin{split}
		 \bm{B}=((\bm{W_{pred}}\bm{W_{proj}})^{-1}\bm{W_{proj}g_k^{t_{his}+1}}-\\
		 \bm{A}\odot \bm{g_k^{t_{his}}})/\bm{g^{t_{his}}}.
	\end{split}
\end{align}

\section{Experiments}

This section introduces the experimental results. Firstly, we introduce the basic experimental setup and related scenarios. Secondly, the experimental results of VERT and other defenses are presented and analyzed.

\subsection{Experimental Setup}\label{section6.1}

\textbf{Datasets:} We use three typical visual image datasets, including MNIST, CIFAR10, and CIFAR100. MNIST is a 28$\times$28 grayscale images of handwritten digits dataset, including 10 categroies, with a total of 60,000 training samples and 10,000 testing samples. CIFAR10 is a 32$\times$32$\times$3 RGB images dataset, including 10 categroies, with a total of 50,000 training samples and 10,000 testing samples. CIFAR100 includes 100 categories, and other attributes consistent with CIFAR10.

\textbf{Models:} We train three CNN models for each datasets. The CNN of MNIST consists of two convolutional layers and two fully connected layers. The CNN of CIFAR10 includes two convolutions and three fully connected layers. The CNN of CIFAR100 is ResNet110 network~\cite{33}.

\textbf{Attacks:} This work mainly discusses the robustness of VERT to large-scale model poisoning attacks in FL system, we use four types of model poisoning attacks, namely Gaussian noise attack (GN)~\cite{19},  Model replacement attack (MR)~\cite{10}, Min-Max distance attack (AGR)~\cite{11}, and A little is enough attack (ALIE)~\cite{34}, as detailed in the Appendix~\ref{appendixe}.

\textbf{Defenses:} We choose three typical horizontal defenses (FedAvg~\cite{1}, Krum~\cite{12}, Median~\cite{13}) and two vertical defenses (FLDetector~\cite{20}, FLANDERS~\cite{19}, VERT) to defend against the large-scale model poisoning attacks mentioned above (see Section~\ref{section7} for details).

\textbf{Simulation Implementation of FL System}: We implement \footnote{The MNIST and CIFAR10 datasets run on NVIDIA GeForce GTX 1660Ti graphics card (6GB VRAM), while CIFAR100 runs on RTX 3090 (24GB).} the basic setup of the FL system using Python language and Pytorch deep learning framework, including an aggregation server (running defenses code for model poisoning attacks and optimal aggregation gradients code) and multiple participating users (running local model training or poisoning attacks code in the model), with a total of 200 communication rounds between the two parties. In order to reflect the large-scale model poisoning attack, we simulated 100 participating users ($|C|$=100) in the FL system, and the ratio of compromised users is $pr$=80\%, 90\%. In each communication round, the aggregation server selects 80 or 90 users ($|C_t|$=80, 90), and the probability of the selected compromised users performing model poisoning attacks is 90\%. To be closer to the real scene, we choose to assign IID or non-IID training data to each participating user. Specifically, training data is allocated to different participating users using the Dirichlet distribution function $DirN(\beta)$, where IID ($\beta$=100) and non-IID ($\beta$=0.6) can be simulated by adjusting the concentration parameter $\beta$.

\textbf{Other training hyperparameter setup:} For the aggregation server to execute the VERT to train coefficient matrix and predictor, we set the optimizer to Adam~\cite{35}, the learning rate to 0.001, the training data to historical gradients of $m$=10 rounds, and the number of training epochs to 5. Set $\kappa$=15, 8 respectively for the cases of $|C_t|$=80, 90 ($pr$=80\%, 90\%). In addition, we designed different projectors and predictors structures for three datasets, as shown in TABLE~\ref{table1}. For participating users in training local models, we set the optimizer to SGD~\cite{36}, the learning rate to 0.001, the local training epochs to 2, and batch size to 50.

\begin{table}[htbp]
	\centering
	\caption{Structural design of projectors and predictors on different datasets.}
	\begin{tabular}{|c|c|c|}
		\toprule
		\textbf{Datasets} & \textbf{Projector} & \textbf{Predictor} \\
		\midrule
		\textbf{MNIST} & \makecell{MLP + softmax \\ I/O: $d\times s$ \\ $d$=21,840, $s$=128} & \makecell{Relu(MLP)$\times$ 3 + softmax \\ I/O: $s\times s$, $s\times s$, $s\times s$} \\
		\midrule
		\textbf{CIFAR10} & \makecell{MLP + softmax \\ I/O: $d\times s$ \\ $d$=62,006, $s$=256} & \makecell{Relu(MLP)$\times$ 3 + softmax \\ I/O: $s\times s$, $s\times s$, $s\times s$} \\
		\midrule
		\textbf{CIFAR100} & \makecell{MLP + softmax \\ I/O: $d\times s$ \\ $d$=1,744,963, $s$=256} & \makecell{Relu(MLP)$\times$ 3 + softmax \\ I/O: $s\times s$, $s\times s$, $s\times s$} \\
		\bottomrule
	\end{tabular}%
	\begin{tablenotes}
		\footnotesize
		\item[1] Note: MLP is fully connected network, Relu is activation function, I/O represents the input/output size of MLP.
	\end{tablenotes}
	\label{table1}%
\end{table}%

\subsection{Experimental Results}\label{section6.2}

We evaluate the effectiveness of VERT from three aspects. Firstly, we evaluate the performance of VERT in predicting user gradients and selecting the optimal aggregation gradients. Secondly, we demonstrate the highest global model accuracy that different defenses can achieve in the face of large-scale model poisoning attacks. Finally, we will represent the computational cost of different defense methods in terms of runtime. The specific experimental results are as follows.

\subsubsection{Prediction Performance of VERT}\label{section6.2.1}

\begin{figure}[h]
	\centering
	\subfloat[MNIST,$|C_t|$=80,$pr$=80\%,$\kappa$=15]{
		\includegraphics[scale=0.18]{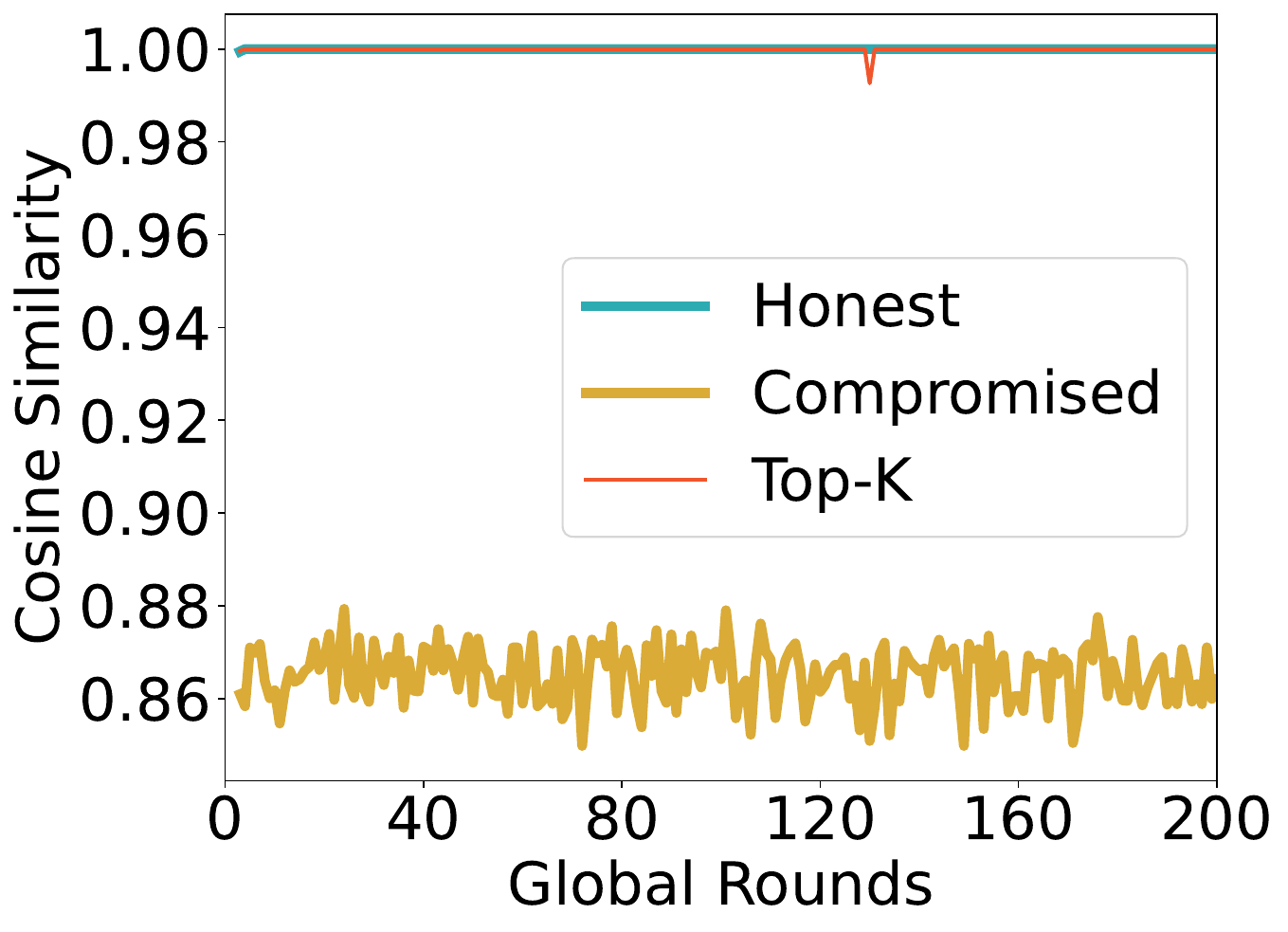}}
	\subfloat[MNIST,90,90\%,8]{
		\includegraphics[scale=0.18]{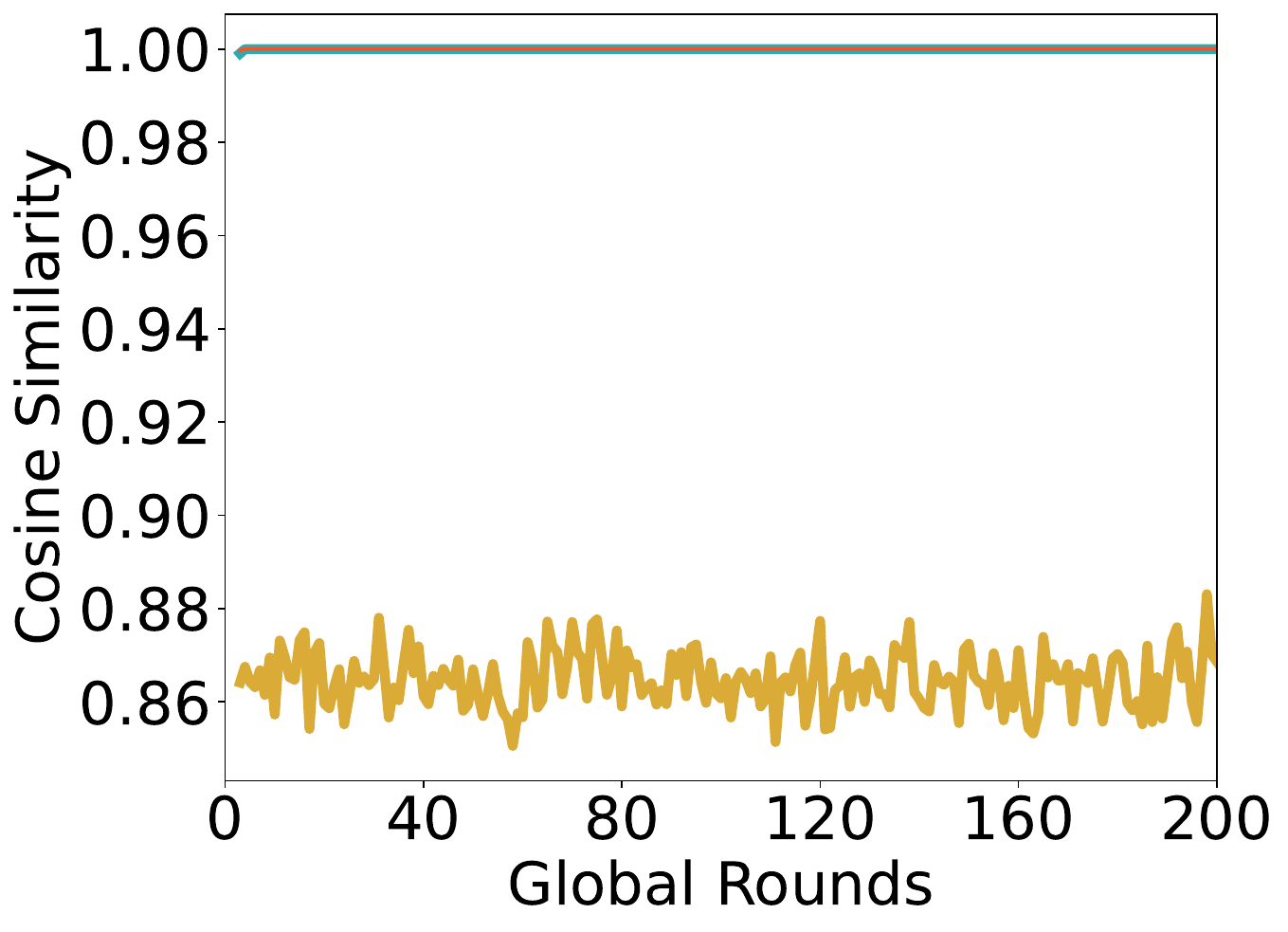}}
	\\
	\subfloat[CIFAR10,80,80\%,15]{
		\includegraphics[scale=0.18]{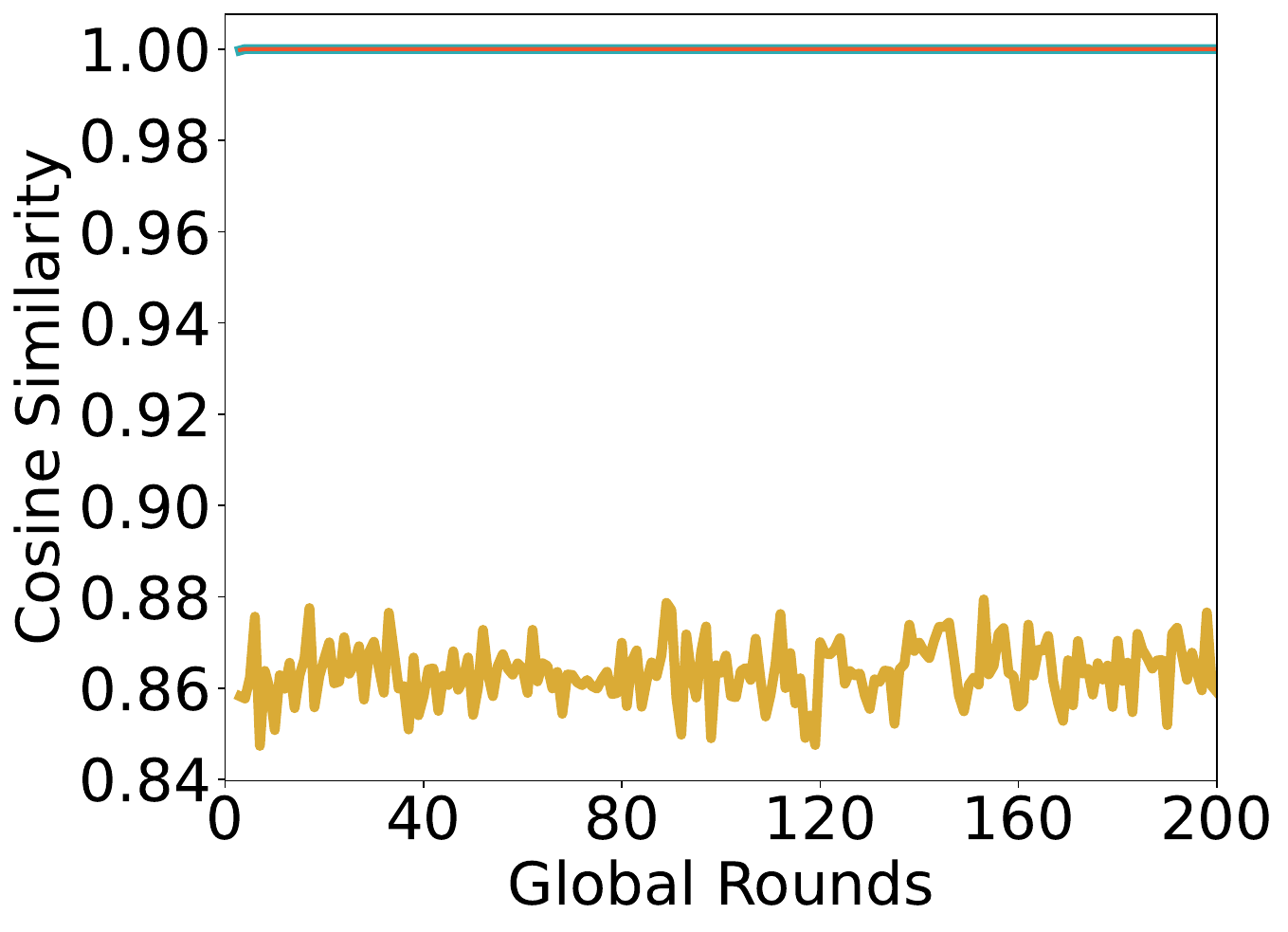}}
	\subfloat[CIFAR10,90,90\%,8]{
		\includegraphics[scale=0.18]{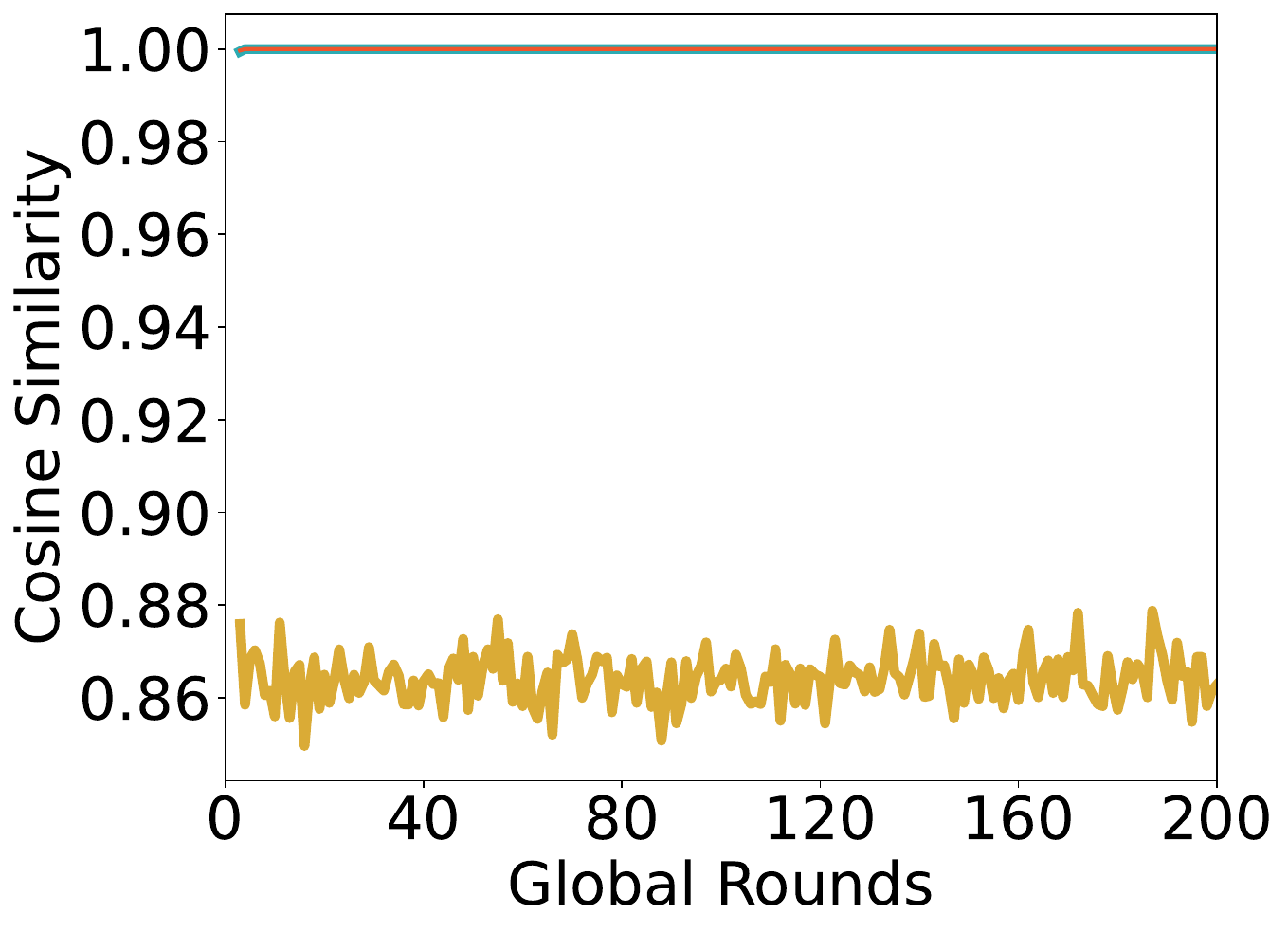}}
	\\
	\subfloat[CIFAR100,80,80\%,15]{
		\includegraphics[scale=0.18]{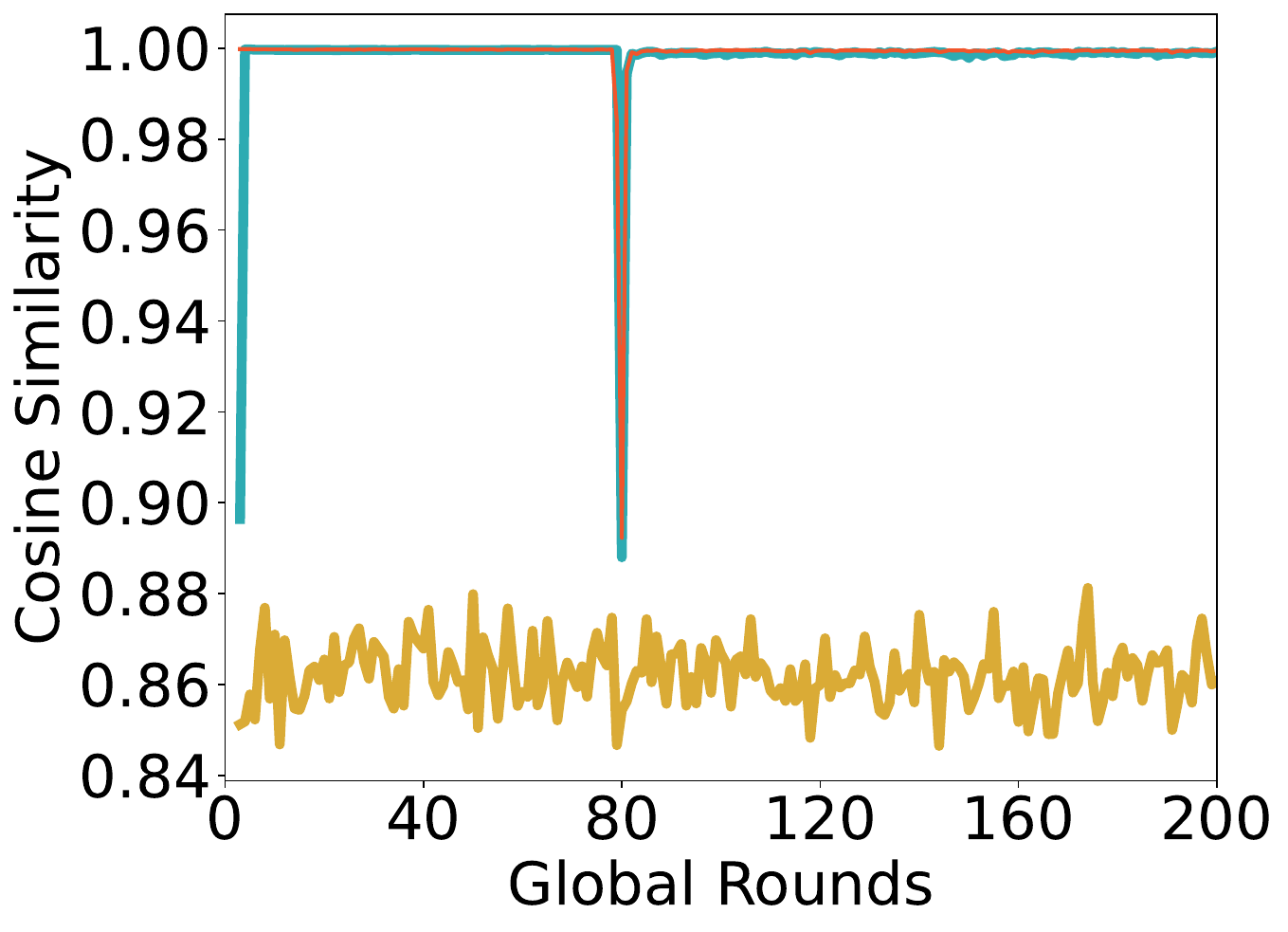}}
	\subfloat[CIFAR100,90,90\%,8]{
		\includegraphics[scale=0.18]{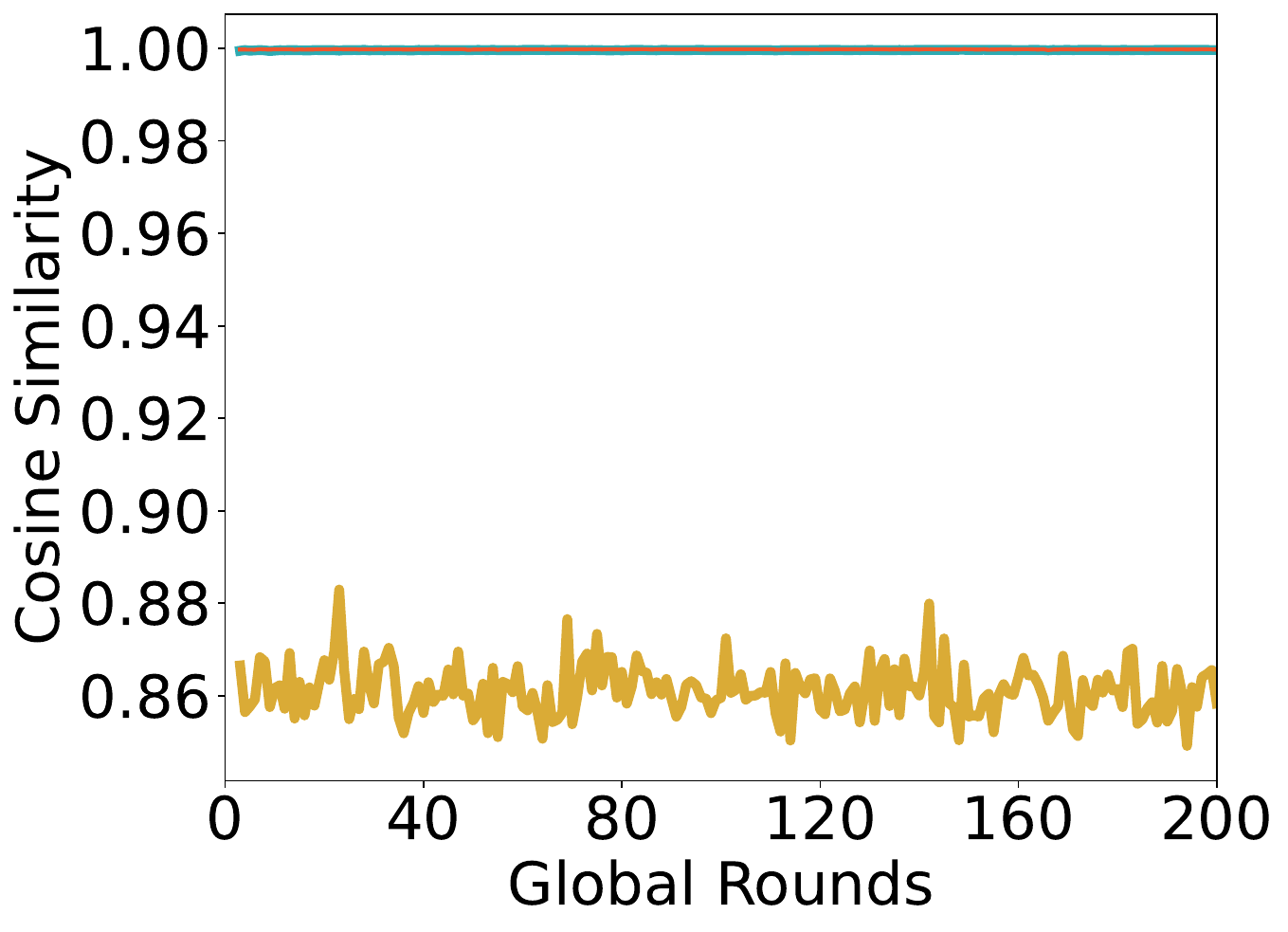}}
	\caption{The performance of VERT in predicting user gradients in the face of large-scale GN attacks in IID scenarios.}
	\label{figure3}
\end{figure}

The key of VERT is whether it can accurately predict user gradients and select honest user gradients as the optimal aggregation gradients. We use cosine similarity to measure the similarity between prediction gradients and real gradients, and select the $top-\kappa$ gradients with the highest similarity as the optimal aggregation gradients. Since training the predictor requires at least 2 rounds of historical gradients, the aggregation server executes FedAvg in the 1, 2-th rounds and starts VERT from the 3-th round (model poisoning attacks start from 3-th round). Taking GN attack as an example, Figure~\ref{figure3} shows the average similarity between honest users, compromised users, and the optimal aggregation gradients selected by VERT in each round in the IID scenario. It can be seen that on the three datasets, the similarity between the prediction gradient and the actual gradient for honest users is extremely high, and the cosine similarity between the two is always close to 1. This not only strongly proves that VERT can effectively predict user gradients, but also proves the thinking of Section~\ref{section4.1} that user gradients have predictability. On the other hand, malicious gradients uploaded by compromised users have lower predictability, with the similarity maintained between 0.86 and 0.88. This allows VERT to select honest user gradients as the optimal aggregation gradients through $top-\kappa$ in the context of large-scale model poisoning attacks, without interference from malicious gradients, ultimately forming a high accurate global model (see Section~\ref{section6.2.2}).

We believe that more people share the same doubts as us, that is, the similarity between VERT's predictions of malicious gradients seems to be very high (0.86-0.88). In this regard, we believe that the distribution characteristics satisfied by malicious gradients may be captured by the predictor, resulting in the predictor being able to generate prediction gradients with similar distribution to malicious gradients. However, the accuracy of predicting malicious gradients is still lower than that of honest gradients.

The predictive performance of VERT for other large-scale poisoning attacks in non-IID and IID scenarios can be seen in the Appendix~\ref{appendixc}.

\subsubsection{Defense Effectiveness}\label{section6.2.2}

\begin{table*}[h]
	\centering
	\caption{The defense effectiveness of different defenses against large-scale model poisoning attacks in IID scenarios.}
	\begin{tabular}{|c|c|c|c|c|c|c|}
		\toprule
		\textbf{Methods} & \textbf{FedAvg} & \textbf{Krum} & \textbf{Median} & \textbf{FLDetector} & \textbf{VERT+Krum} & \textbf{VERT+FedAvg} \\
		\midrule
		\multicolumn{7}{|c|}{MNIST, $|C_t|$ = 80, 90, $pr$ = 80\%, 90\%} \\
		\midrule
		\textbf{GN} & 16.45\%, 16.38\% & \colorbox{color2}{51.04\%}, \colorbox{color2}{49.64\%} & 27.55\%, 31.94\% & 11.23\%, 11.54\% & 42.73\%, 15.24\% & \colorbox{color1}{58.29\%}, \colorbox{color1}{81.11\%} \\
		\midrule
		\textbf{MR} & 12.11\%, 11.35\% & \colorbox{color2}{42.14\%}, \colorbox{color1}{58.79\%} & 14.20\%, 12.72\% & 10.01\%, 12.36\% & \colorbox{color1}{50.76\%}, \colorbox{color2}{50.87\%} & 18.99\%, 12.48\% \\
		\midrule
		\textbf{AGR} & 9.58\%, 9.80\% & 9.80\%, 9.82\%  & 9.82\%, 9.82\% & 9.80\%, 9.80\% & \colorbox{color1}{59.61\%}, \colorbox{color1}{63.76\%} & \colorbox{color2}{10.28\%}, \colorbox{color2}{11.54\%} \\
		\midrule
		\textbf{ALIE} & 11.35\%, 10.32\% & 10.28\%, 10.32\% & 10.09\%, 10.09\% & 9.80\%, 9.80\% & \colorbox{color1}{62.09\%}, \colorbox{color1}{42.51\%} & \colorbox{color2}{12.35\%}, \colorbox{color2}{19.60\%} \\
		\midrule
		\multicolumn{7}{|c|}{CIFAR10, $|C_t|$ = 80, 90, $pr$ = 80\%, 90\%} \\
		\midrule
		\textbf{GN} & 12.43\%, 11.95\% & 13.37\%, 19.41\% & 17.18\%, 12.15\% & 13.62\%, 11.28\% & \colorbox{color2}{20.68\%}, \colorbox{color2}{20.83\%} & \colorbox{color1}{32.43\%}, \colorbox{color1}{28.83\%} \\
		\midrule
		\textbf{MR} & 22.17\%, \colorbox{color2}{25.48\%} & 12.59\%, 20.16\% & \colorbox{color2}{24.94\%}, 21.47\% & 10.09\%, 12.57\% & 19.33\%, 18.05\% & \colorbox{color1}{34.25\%}, \colorbox{color1}{36.49\%} \\
		\midrule
		\textbf{AGR} & 10.00\%, 10.00\% & 10.00\%, 10.00\% & 10.00\%, 10.00\% & 10.00\%, 10.00\% & \colorbox{color1}{16.29\%}, \colorbox{color1}{14.99\%} & \colorbox{color2}{10.16\%}, 10.00\% \\
		\midrule
		\textbf{ALIE} & 10.00\%, 10.00\% & 10.00\%, 10.00\% & 10.00\%, 10.00\% & 10.00\%, 10.00\% & \colorbox{color1}{15.66\%}, \colorbox{color1}{18.08\%} & 10.00\%, \colorbox{color2}{17.30\%} \\
		\midrule
		\multicolumn{7}{|c|}{CIFAR100, $|C_t|$ = 80, 90, $pr$ = 80\%, 90\%} \\
		\midrule
		\textbf{GN} & 1.94\%, 1.67\% & \colorbox{color2}{3.46\%}, \colorbox{color1}{3.21\%} & 1.88\%, 1.75\% & 1.31\%, 1.57\% & \colorbox{color1}{3.85\%}, \colorbox{color2}{3.07\%} & 2.46\%, 2.97\% \\
		\midrule
		\textbf{MR}& 1.06\%, 1.00\% & 2.38\%, 2.89\% & 1.19\%, 1.11\% & 1.00\%, 0.98\% & \colorbox{color1}{4.12\%}, \colorbox{color1}{3.91\%} & \colorbox{color2}{3.71\%}, \colorbox{color2}{3.69\%} \\
		\midrule
		\textbf{AGR} & 3.29\%, 3.07\% & 2.60\%, 2.39\% & 2.43\%, 2.07\% & 1.13\%, 1.19\% & \colorbox{color2}{3.63\%}, \colorbox{color1}{4.08\%} & \colorbox{color1}{3.82\%}, \colorbox{color2}{3.65\%} \\
		\midrule
		\textbf{ALIE} & 1.18\%, 1.04\% & 1.00\%, 1.00\% & 1.00\%, 1.00\% & 1.00\%, 1.00\% & \colorbox{color1}{3.64\%}, \colorbox{color2}{3.42\%} & \colorbox{color2}{3.27\%}, \colorbox{color1}{3.50\%} \\
		\bottomrule
	\end{tabular}
	\begin{tablenotes}
		\footnotesize
		\item[1] Note: The data in each cell is the highest global model accuracy, on the left is the result of $pr$=80\%, and the right is 90\%.
	\end{tablenotes}
	\label{table2}
\end{table*}

The ultimate goal of VERT is to achieve defense against large-scale model poisoning attacks in FL system. We test the defense effectiveness (reflected by the highest accuracy of the global model) of different defenses against $pr$=80\%, 90\%, and TABLE~\ref{table2} shows the experimental results under IID. VERT+Krum and VERT+FedAvg represent the use of Krum or FedAvg for successive aggregation method after VERT selects the optimal aggregation gradients. We do not provide the defense effectiveness of FLANDERS because it requires significant computational and memory overhead (Original paper uses a graphics card with 964GB of memory), and our device cannot support its full operation.

It can be seen that VERT achieves the best defense effectiveness on all three datasets. On MNIST, when facing 80\% and 90\% large-scale GN attacks, VERT+FedAvg can achieve the \colorbox{color1}{\textbf{highest global model accuracy}} of 58.29\% and 81.11\%, which is 7.25\% and 31.47\% higher than the \colorbox{color2}{\textbf{suboptimal accuracy}} of Krum. When facing MR attacks, the effect of VERT+FedAvg is not significant because MR achieves global model replacement through gradient amplification. Therefore, even if the optimal aggregation gradients selected by VERT contains only one malicious gradient, it can still result in poor aggregation results. On the contrary, Krum reaches a peak of 58.79\%. However, the premise for Krum to perform well is that the aggregation server needs to know the number of compromised users, which is unfair to other defenses and also unrealistic. Nevertheless, VERT+Krum can still achieve 50.76\% and 50.87\% respectively, without assuming that the aggregation server knows any knowledge about the attacker. The reason why VERT+Krum is effective is that the optimal aggregation gradients selected by VERT only contains a few or even no malicious gradients, while Krum as a successive aggregation method can effectively avoid a few malicious gradients. When facing the more powerful attacks of AGR and ALIE, all defenses except for VERT cannot resist large-scale attacks. Even though VERT can select honest user gradients, the few malicious gradients selected still prevent VERT+FedAvg from training the global model well. In contrast, VERT+Krum achieves better model accuracy by further filter gradients, reaching 59.61\% and 63.76\% under AGR, which is more than 50\% higher than traditional horizontal defenses. Under ALIE, the accuracy also reaches 62.09\% and 42.51\%, which is more than 30\%-50\% higher than traditional horizontal defense.

On CIFAR10, the defense effectiveness of VERT is more significant, significantly higher than all other defenses. When facing 80\% and 90\% GN attacks, VERT+FedAvg is 15.25\% and 9.42\% higher than horizontal defenses. In the face of MR attacks, VERT is 9.31\% and 11.01\% higher than horizontal defense. When facing AGR and ALIE attacks, similar to MNIST, VERT+Krum performs better than VERT+FedAvg. From this, it can be seen that the aggregation server should choose different successive aggregation methods to be used in combination with VERT based on actual scenarios in order to achieve the best defense effectiveness. On CIFAR100, due to the complexity of the dataset and the small number of honest users (20\%, 10\%), the accuracy of all defenses is low. However, VERT can still bring more improvements to the global model compared to other defenses.

FLDetector is also a type of vertical defense, but its defense effectiveness is poor because it requires multiple rounds of normal historical gradients without poisoning attacks to estimate the Hessian matrix (set to 50 rounds in the original paper). In the configuration of this paper, in order to be closer to real-world attack scenarios, we conduct large-scale model poisoning attacks at the beginning stage of the FL system. At this time, the FLDetector will not be able to correctly estimate the Hessian matrix, resulting in poor prediction performance of user gradients.

The performance of different defenses in non-IID scenarios can be seen in the Appendix~\ref{appendixd}.

\subsubsection{Runtime Costs}

Defense effectiveness (global model accuracy) is one of the metrics for evaluating the effectiveness of a defense, and another key metric is computational cost. We need to design a defense with low computational cost and good defense effectiveness. Here, we will use the runtime in the experiment to represent the computational cost of different defenses.

We choose FLANDERS, which is most similar to VERT (both use regression methods to predict user gradients), for runtime comparison. Section~\ref{section5} has theoretically analyzed that the computational complexity of VERT approaches $O(ds)$ ($s\ll d$), while FLANDERS is $O(d^2)$. Here, the runtime is used to more intuitively demonstrate that VERT has a smaller computational costs. Since the code executed by the aggregation server for defense is not related to poisoning attacks and data distribution, so we take GN attack and IID scenario as an example, and TABLE~\ref{table3} shows the average runtime of VERT and FLANDERS in all rounds. For fairness, we set the same training hyperparameters for FLANDERS as for VERT, as detailed in Section~\ref{section6.1}.

It is worth noting that due to the huge additional memory costs required by FLANDERS (for CIFAR10, it needs 15GB RAM, and CIFAR100 needs 11,300GB), our device cannot support its operation. Therefore, we indirectly calculate the approximate runtime of FLANDERS based on the runtime of VERT and the number of parameters that need to be optimized during the training process of VERT and FLANDERS. For example, if the number of parameters that need to be optimized for VERT is $num_{vert}$ and the runtime is $rt$, the number of parameters that need to be optimized for FLANDERS is $num_{fld}$, then the estimated runtime for FLANDERS is $\frac{num_{vert}}{num_{fld}}\times rt$. After calculation, the $num_{vert}$ on MNIST, CIFAR10, and CIFAR100 are (92,832, 320,620, 3,686,534)$\times |C_t|$ , respectively.  The $num_{fld}$ are 476,995,600, 3,844,754,036, and 3,044,895,881,369. Therefore, the runtime is shown in TABLE~\ref{table3}, where \colorbox{color3}{yellow} represents the estimated runtime and \colorbox{color4}{green} represents the actual runtime. It can be seen that FLANDERS, which is also a vertical defense, requires much more running time than VERT. Although its runtime is estimated through calculation, it also indicates to some extent that FLANDERS has poor availability. Section~\ref{section4.2.3} also explains this from the perspective of memory costs.

\begin{table}[h]
	\centering
	\caption{Comparison of defense efficiency between VERT and FLANDERS.}
	\begin{tabular}{|c|c|c|}
		\toprule
		\textbf{Methods} & \textbf{FLANDERS+FedAvg} & \textbf{VERT+FedAvg} \\
		\midrule
		\multicolumn{3}{|c|}{MNIST, $|C_t|$ = 80, 90} \\
		\midrule
		\textbf{Runtime} & \colorbox{color3}{265.91s}, \colorbox{color3}{263.19s} & \colorbox{color4}{4.14s}, \colorbox{color4}{4.61s} \\
		\midrule
		\multicolumn{3}{|c|}{CIFAR10, $|C_t|$ = 80, 90} \\
		\midrule
		\textbf{Runtime} & \colorbox{color3}{1,767.27s}, \colorbox{color3}{1,749.44s} & \colorbox{color4}{11.79s}, \colorbox{color4}{13.13s} \\
		\midrule
		\multicolumn{3}{|c|}{CIFAR100, $|C_t|$ = 80, 90} \\
		\midrule
		\textbf{Runtime} & \colorbox{color3}{1,031,715.85s}, \colorbox{color3}{1,022,068.29s} & \colorbox{color4}{99.93s}, \colorbox{color4}{111.37s} \\
		\bottomrule
	\end{tabular}
	\begin{tablenotes}
		\footnotesize
		\item[1] Note: The data in each cell is the runtime, on the left is the result of $|C_t|$=80, and the right is 90.
	\end{tablenotes}
	\label{table3}
\end{table}

\section{Related Work}\label{section7}

This section mainly introduces the recently proposed and popular defenses against model poisoning attacks on FL system. We consider both horizontal and vertical solution.

\textbf{(1) Horizontal solution:}

\begin{itemize}
	
	\item Krum~\cite{12}: The aggregation server calculates the Euclidean distance between each user gradient and the gradients of other users, and accumulates the sum of the distances. The user gradient with the smallest sum of distances is ultimately chosen as the optimal aggregation gradient. In addition, an extended method, Muilt-Krum~\cite{12}, obtains multiple optimal aggregation gradients through iterative Krum multiple times, and then performs FedAvg.
	
	\item Median~\cite{13}: Based on all user gradients, sort and select the median for each gradient element, and finally combine them to form the optimal aggregated gradient. Trimed Mean~\cite{13} is an extended version of Median that performs maximum and minimum pruning on sorted gradient elements, calculates the average, and finally combines them into the optimal aggregated gradient.
	
	\item PEFL~\cite{16}: Calculates the benchmark gradient according to Median~\cite{13}. Then, calculates the cosine similarity between each user gradient and the benchmark gradient, and normalize the cosine similarity as the gradient weight for weighted aggregation.
	
	\item FLtrust~\cite{18}: Assuming that the aggregation server has a completely clean and user data distribution consistent dataset (Trust root), the gradient obtained by training the model using trust root is used as the benchmark, and subsequent calculations are consistent with PEFL~\cite{16}.
	
\end{itemize}

The above methods will fail when facing large-scale model poisoning attacks, as malicious gradient dominance problem will make the calculation of the optimal aggregation gradients more dependent on malicious gradients. Even though FLTrust circumvents the malicious gradient dominance problem by introducing a trust root, its underlying assumption is unrealistic, which is the existence of a completely clean dataset that is consistent with the distribution of user data.

\textbf{(2) Vertical solution:}

We carefully researched recent related work and found that FLDetector~\cite{20} is the earliest vertical solution based on Cauchy's mean value theorem. It approximates the Hessian matrix using the L-BFGS~\cite{37} algorithm and ultimately predicts the user gradient. However, FLDetector's excessive reliance on mathematical formulas results in its inability to work well in some special situations, such as large-scale poisoning attacks and non-IID of user data. FLANDERS~\cite{19} is most similar to our work in proposing the use of autoregressive matrices to predict user gradients. However, the construction of this autoregressive matrix depends on the user gradient dimension $d$, and the number of parameters in the matrix coefficients exceeds $d^2$, while the computational complexity also exceeds $O(d^2)$. When facing large models (ResNet series, with parameter sizes in the tens of millions), their computational and memory resource costs is extremely high, and their scalability is poor.

\section{Conclusion}

We proposes a novel vertical defense method, VERT, for large-scale model poisoning attacks in FL system. Unlike traditional horizontal defenses, VERT trains a gradient predictor using historical user gradients and global gradients, and uses the similarity between the prediction gradient and the actual gradient as the criterion to determine whether it participates in gradients aggregation. In addition, to further reduce the costs of aggregation server to train predictor, we design a low dimensional vector projector that projects user gradients into a computationally acceptable range  of low dimensional features, which can reduce the original computational complexity from $O(d^2)$ to $O(ds)$ ($s\ll d$). Ultimately, VERT achieves high prediction accuracy and effectively defends against large-scale model poisoning attacks.

%

\bibliographystyle{IEEEtran}
\bibliography{references}

\newpage

\appendix

\subsection{Pseudocode of VERT}\label{appendixa}

Below we provide the pseudocode~\ref{algoritnm1} for VERT.

\begin{algorithm}
	\caption{Execution  of VERT.}
	\label{algoritnm1}
	\begin{algorithmic}[1]
		\STATE \textbf{Input}: Gradients uploaded by the selected user in the $t$-th round $\{\bm{g_k^t}|k\in C_t\}$, the selected user's historical $m$ rounds gradients $\{\{\bm{g_k^{t_{his}}}\}_{t_{his}=t-m}^{t_{his}=t-1}|k\in C_t\}$, the aggregation server's historical $m$ rounds gradients $\{\bm{g^{t_{his}}}\}_{t_{his}=t-m}^{t_{his}=t-1}$, coefficient matrix $\bm{A}$ and $\bm{B}$, predictor $f_{pred}$, projector $f_{proj}$.
		\STATE \textbf{Output}: The optimal aggregation gradients set $G_{opt}$.
		
		\FOR{$k$ in $C_t$}
		\STATE \texttt{/* Training predictor and coefficient matrixes. */}
		\FOR{$t_{his}$ in $[t-m,t-2]$}
		\STATE Integrate historical user gradient and global gradient:
		\STATE $\bm{g_k^{t_{his}}}=
		\left\{
		\begin{array}{ll}
			\bm{g_k^{t_{his}}}, k\in C_{opt}^t \\
			\\
			\bm{g^{t_{his}}}, k\notin C_{opt}^t
		\end{array}
		\right.$. \texttt{/* Exception handling, see Section~\ref{section4.2.2}. */}
		\STATE $\bm{g_{input}^{t_{his}}}=\bm{A}\odot \bm{g_k^{t_{his}}}+\bm{B}\odot \bm{g^{t_{his}}}$.
		\STATE Project the input into a low dimensional feature space:
		\STATE $\bm{p_{input}^{t_{his}}}=f_{proj}(\bm{g_{input}^{t_{his}}})$.
		\STATE Train predictor and coefficient matrices:
		\STATE $\bm{A},\bm{B},f_{pred}=arg\min\limits_{\bm{A},\bm{B},f_{pred}}||f_{pred}(\bm{p_{input}^t})-f_{proj}(\bm{g_k^{t_{his}+1}})||_2$.
		\ENDFOR
		
		\STATE \texttt{/* Predicting user gradients and calculating similarity. */}
		\STATE Use cosine function to calculate the similarity between predicted gradient and actual gradient:
		\STATE $\bm{g_{input}^{t-1}}=\bm{A}\odot \bm{g_k^{t-1}}+\bm{B}\odot \bm{g^{t-1}}$.
		\STATE $\rho_k=cos(f_{pred}(f_{proj}(\bm{g_{input}^{t-1}})),f_{proj}(\bm{g_k^t}))$.
		\ENDFOR
		
		\STATE \texttt{/* Update global model. */}
		\STATE Select the $top-\kappa$  user gradients with the highest similarity as the optimal aggregation gradients:
		\STATE $C_{opt}^t=top-\kappa({\rho_k|k\in C_t})$.
		\STATE Calculate global gradient:
		\STATE $\bm{g^{t+1}}=\frac{1}{\kappa}\sum_{k\in C_{opt}^t}^{\bm{g_k^t}}$.
		\STATE Update global model:
		\STATE $\bm{w^{t+1}}=\bm{w^t}+\bm{g^{t+1}}$.
		
		\RETURN $\bm{w^{t+1}}$
	\end{algorithmic}
\end{algorithm}

\subsection{Theoretical Analysis on Coefficient Matrix and Predictor}\label{appendixb}

Calculate the partial derivative of the optimization objective on $\bm{A}$. Let the partial derivative is 0, then:

\begin{align}\label{formula9}
	\begin{split}
		\frac{\partial \Phi(\bm{A};\bm{B};f_{pred};f_{proj})}{\partial \bm{A}}=0.
	\end{split}
\end{align}

Let $X=f_{pred}(f_{proj}(\bm{A}\odot \bm{g_k^{t_{his}}}+\bm{B}\odot \bm{g^{t_{his}}}))$, $Y=f_{proj}(\bm{g_k^{t_{his}+1}})$, because the outputs of the predictor and projector are vectors of length $s$, i.e., $X,Y\in \mathbb{R}^{s}$. According to the vector norm calculation $||a-b||_2=(a-b)^\mathtt{T}(a-b)$, Formula~\ref{formula9} is further rewritten as:

\begin{align}
	\begin{split}
		\frac{\partial \sum_{t_{his}=t-m}^{t-2}\sqrt{X^\mathtt{T}X-X^\mathtt{T}Y-Y^\mathtt{T}X+Y^\mathtt{T}Y}}{\partial \bm{A}}=0.
	\end{split}
\end{align}

Where $\mathtt{T}$ represents transposition. Extract the summation formula to the outside of the derivative formula and take the derivative of the root sign:

\begin{align}\label{formula11}
	\begin{split}
		\sum_{t_{his}=t-m}^{t-2}\frac{\frac{\partial X^\mathtt{T}X-X^\mathtt{T}Y-Y^\mathtt{T}X+Y^\mathtt{T}Y}{\partial \bm{A}}}{2\sqrt{X^\mathtt{T}X-X^\mathtt{T}Y-Y^\mathtt{T}X+Y^\mathtt{T}Y}}=0.
	\end{split}
\end{align}

Firstly, we calculate $\frac{\partial X^\mathtt{T}X}{\partial \bm{A}}$:

\begin{align}\label{formula12}
	\begin{split}
		\frac{\partial X^\mathtt{T}X}{\partial \bm{A}}&=2X\frac{\partial X}{\partial \bm{A}}\\
		&=2X\frac{\partial f_{pred}(f_{proj}(\bm{A}\odot \bm{g_k^{t_{his}}}+\bm{B}\odot \bm{g^{t_{his}}}))}{\partial \bm{A}}\\
		&=2X\frac{\partial f_{pred}}{\partial f_{proj}}\frac{\partial f_{proj}}{\partial \bm{A}\odot \bm{g_k^{t_{his}}}+\bm{B}\odot \bm{g^{t_{his}}}}\frac{\partial \bm{A}\odot \bm{g_k^{t_{his}}}+\bm{B}\odot \bm{g^{t_{his}}}}{\partial \bm{A}}\\
		&=2X\frac{\partial f_{pred}}{\partial f_{proj}}\frac{\partial f_{proj}}{\partial \bm{A}\odot \bm{g_k^{t_{his}}}+\bm{B}\odot \bm{g^{t_{his}}}}\bm{g_k^{t_{his}}}.
	\end{split}
\end{align}

Note $\frac{\partial f_{pred}}{\partial f_{proj}}\frac{\partial f_{proj}}{\partial \bm{A}\odot \bm{g_k^{t_{his}}}+\bm{B}\odot \bm{g^{t_{his}}}}=f_{pred}'f_{proj}'$, then $\frac{\partial X^\mathtt{T}X}{\partial \bm{A}}=2Xf_{pred}'f_{proj}'\bm{g_k^{t_{his}}}$.

Secondly, we calculate $-\frac{\partial X^\mathtt{T}Y+Y^\mathtt{T}X-Y^\mathtt{T}Y}{\partial \bm{A}}$. Since $Y$ is unrelated to $\bm{A}$, therefore, $\frac{\partial Y^\mathtt{T}Y}{\partial \bm{A}}=0$. The original formula $=-\frac{\partial X^\mathtt{T}Y}{\partial \bm{A}}-\frac{\partial Y^\mathtt{T}X}{\partial \bm{A}}=-Y\frac{\partial X^\mathtt{T}}{\partial \bm{A}}-Y^\mathtt{T}\frac{\partial X}{\partial \bm{A}}$. According to Formula~\ref{formula12} $\frac{\partial X}{\partial \bm{A}}=f_{pred}'f_{proj}'\bm{g_k^{t_{his}}}$, therefore:

\begin{align}
	\begin{split}
		-Y\frac{\partial X^\mathtt{T}}{\partial \bm{A}}-Y^\mathtt{T}\frac{\partial X}{\partial \bm{A}}=-2Yf_{pred}'f_{proj}'\bm{g_k^{t_{his}}}.
	\end{split}
\end{align}

Therefore, Formula~\ref{formula11} can be rewritten:

\begin{align}
	\begin{split}
		\sum_{t_{his}=t-m}^{t-2}\frac{(2X-2Y)f_{pred}'f_{proj}'\bm{g_k^{t_{his}}}}{2\sqrt{X^\mathtt{T}X-X^\mathtt{T}Y-Y^\mathtt{T}X+Y^\mathtt{T}Y}}=0.
	\end{split}
\end{align}

When $m$=2, due to $\frac{1}{2\sqrt{X^\mathtt{T}X-X^\mathtt{T}Y-Y^\mathtt{T}X+Y^\mathtt{T}Y}}>0$, let:

\begin{align}\label{formula15}
	\begin{split}
		(2X-2Y)f_{pred}'f_{proj}'\bm{g_k^{t_{his}}}=0.
	\end{split}
\end{align}

Assuming  projector is $f_{proj}(g)=\bm{W_{proj}}g$, $\bm{W_{proj}}\in \mathbb{R}^{d\times s}$, predictor is $f_{pred}(p)=\bm{W_{pred}}p$, $\bm{W_{ped}}\in \mathbb{R}^{s\times s}$, then $X=\bm{W_{pred}}\bm{W_{proj}}(\bm{A}\odot \bm{g_k^{t_{his}}}+\bm{B}\odot \bm{g^{t_{his}}})$, $Y=\bm{W_{proj}g_k^{t_{his}+1}}$. Formula~\ref{formula15} can be rewritten:

\begin{align}
	\begin{split}
		(2\bm{W_{pred}}\bm{W_{proj}}(\bm{A}\odot \bm{g_k^{t_{his}}}+\bm{B}\odot \bm{g^{t_{his}}})-\\ 2\bm{W_{proj}g_k^{t_{his}+1}})f_{pred}'f_{proj}'\bm{g_k^{t_{his}}}=0.
	\end{split}
\end{align}

Let:

\begin{align}
	\begin{split}
		2\bm{W_{pred}}\bm{W_{proj}}(\bm{A}\odot \bm{g_k^{t_{his}}}+\bm{B}\odot \bm{g^{t_{his}}})-\\ 2\bm{W_{proj}}\bm{g_k^{t_{his}+1}}=0.
	\end{split}
\end{align}

Then:

\begin{align}
	\begin{split}
		\bm{A}=((\bm{W_{pred}}\bm{W_{proj}})^{-1}\bm{W_{proj}g_k^{t_{his}+1}}-\\
		\bm{B}\odot \bm{g^{t_{his}}})/\bm{g_k^{t_{his}}}. 
	\end{split}
\end{align}

Where $/$ is the division of vector corresponding terms.

\subsection{Model Poisoning Attacks}\label{appendixe}

Below, we provide a detailed introduction to the four types of model poisoning attacks used in the experiment.

\begin{itemize}
	
	\item Gaussian noise attack (GN)~\cite{19}: The attacker controls compromised users to sample noise from the $N(0,1)$ Gaussian distribution as malicious user gradients and upload them to the aggregation server.
	
	\item Model replacement attack (MR)~\cite{10}: Expand the local poisoning model based on the total number of participating users to calculate the malicious gradients, and replace the global model with the local poisoning model.
	
	\item Min-Max distance attack (AGR)~\cite{11}: Utilize honest user gradient $g$ to calculate unit interference gradient $\triangledown^p=-\frac{g}{||g||_2}$, optimize expansion factor $\lambda$ to calculate malicious gradient $g+\lambda \triangledown^p$.
	
	\item A little is enough attack (ALIE)~\cite{34}: Calculate the maximum interference range $z^{max}$ (set to 1.0 in the experiments) based on the total number of participating users and the number of compromised users, use malicious gradients of all compromised users to calculate the mean $\mu$ and variance $\sigma$, and finally configure the same malicious gradient $\mu +z^{max}\sigma$ for all compromised users. 
	
\end{itemize}

\subsection{Prediction Performance of VERT}\label{appendixc}

\begin{figure}[h]
	\centering
	\subfloat[MNIST,$|C_t|$=80,$pr$=80\%,$\kappa$=15]{
		\includegraphics[scale=0.18]{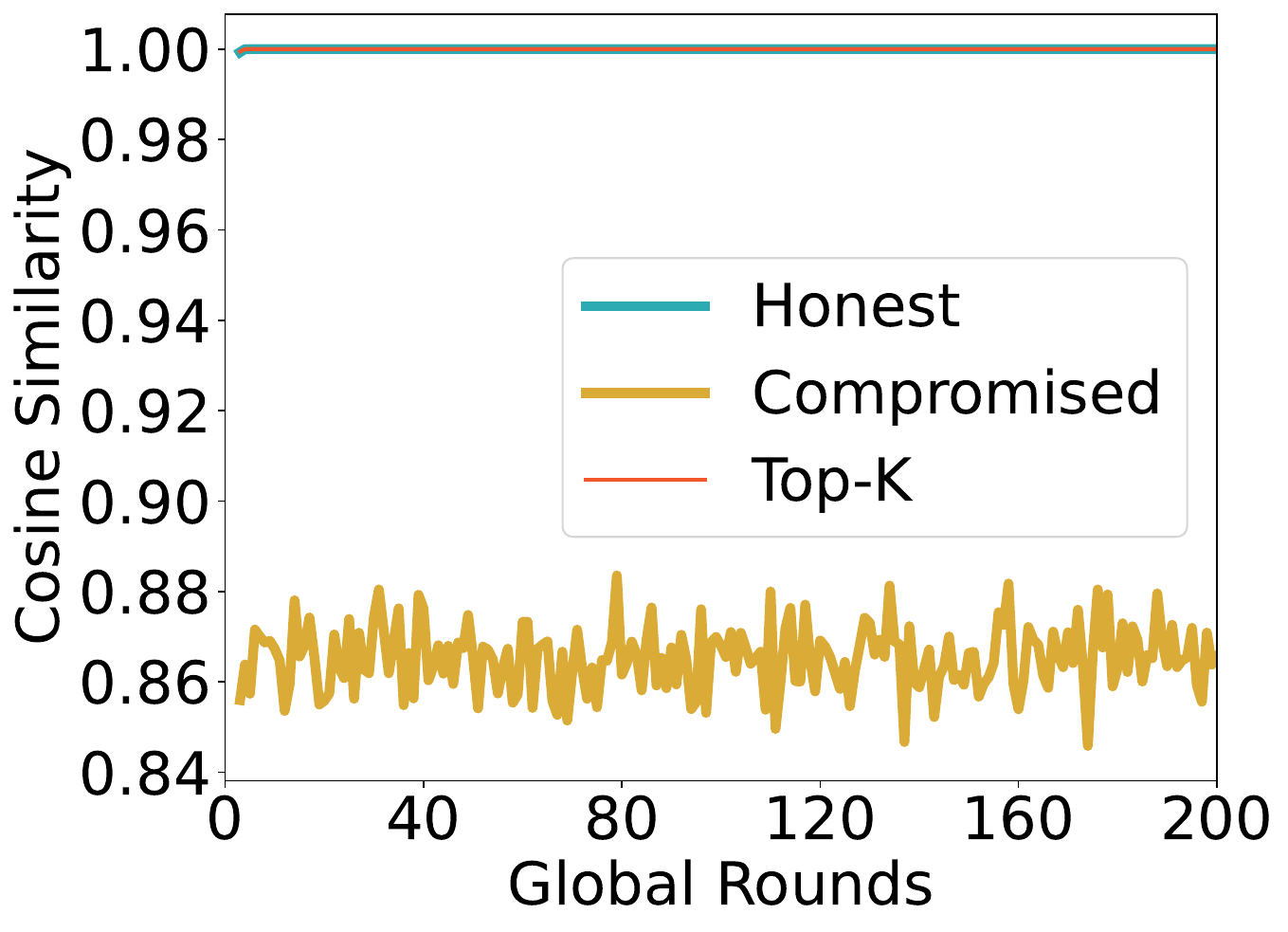}}
	\subfloat[MNIST,90,90\%,8]{
		\includegraphics[scale=0.18]{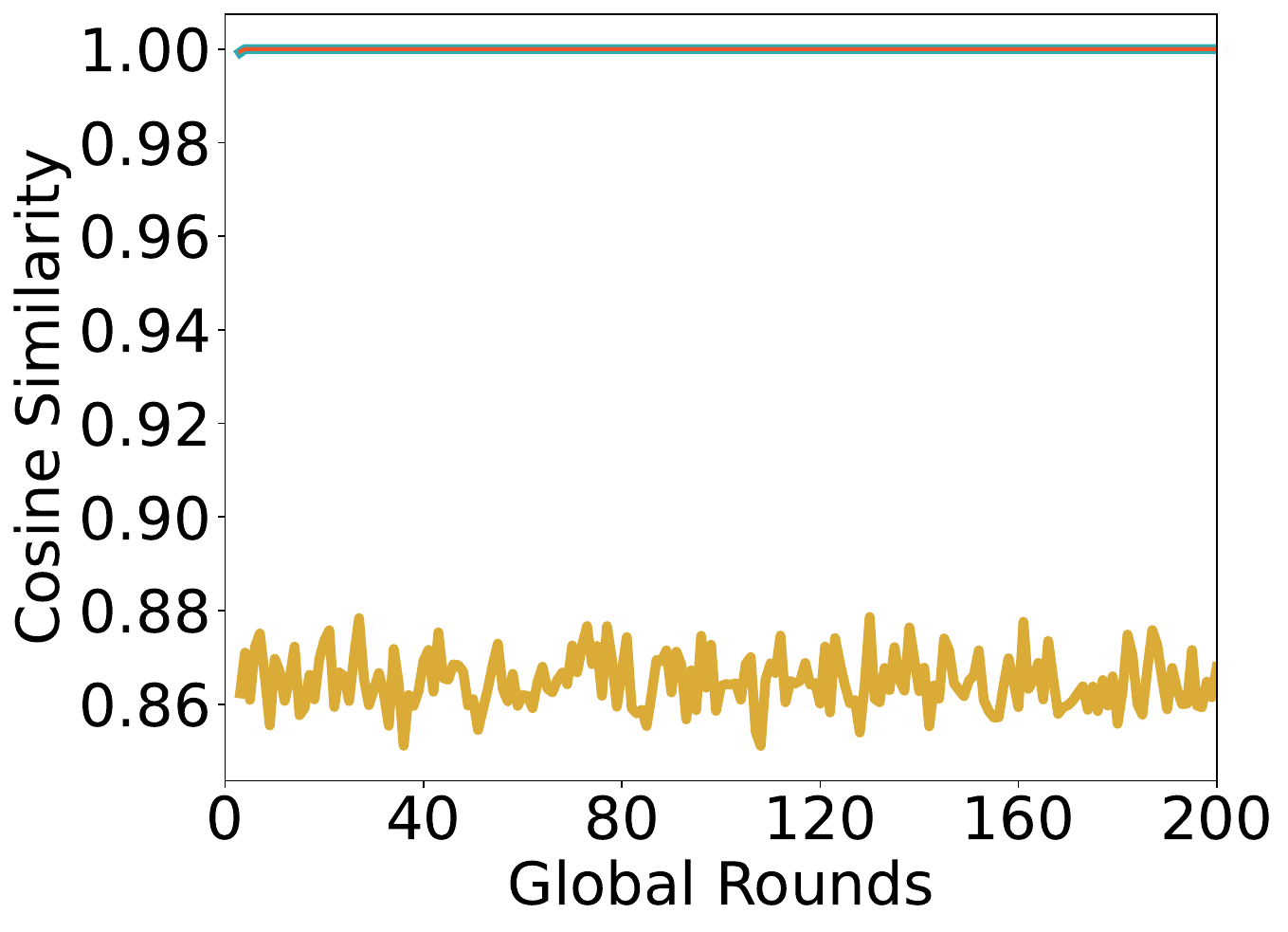}}
	\\
	\subfloat[CIFAR10,80,80\%,15]{
		\includegraphics[scale=0.18]{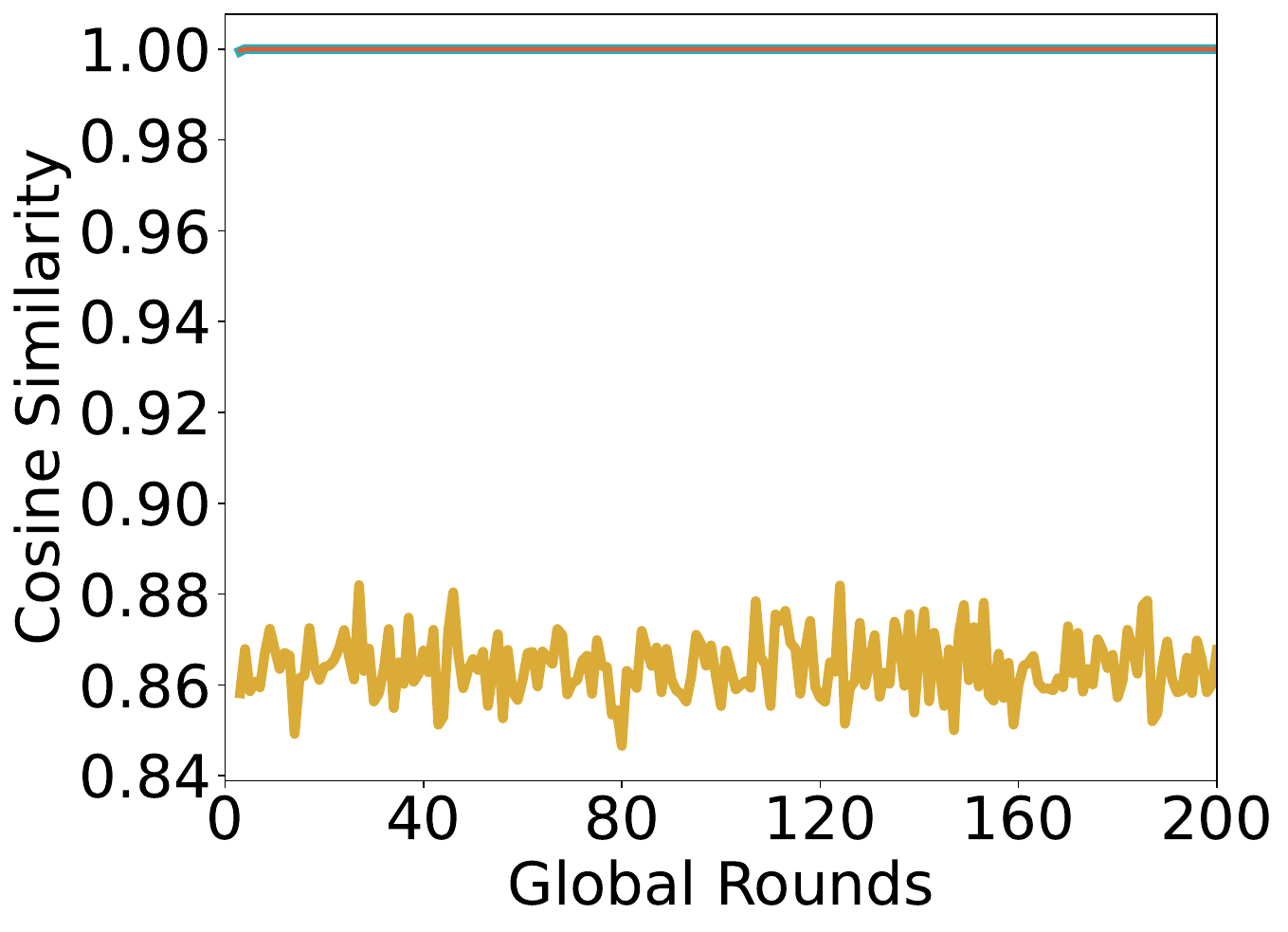}}
	\subfloat[CIFAR10,90,90\%,8]{
		\includegraphics[scale=0.18]{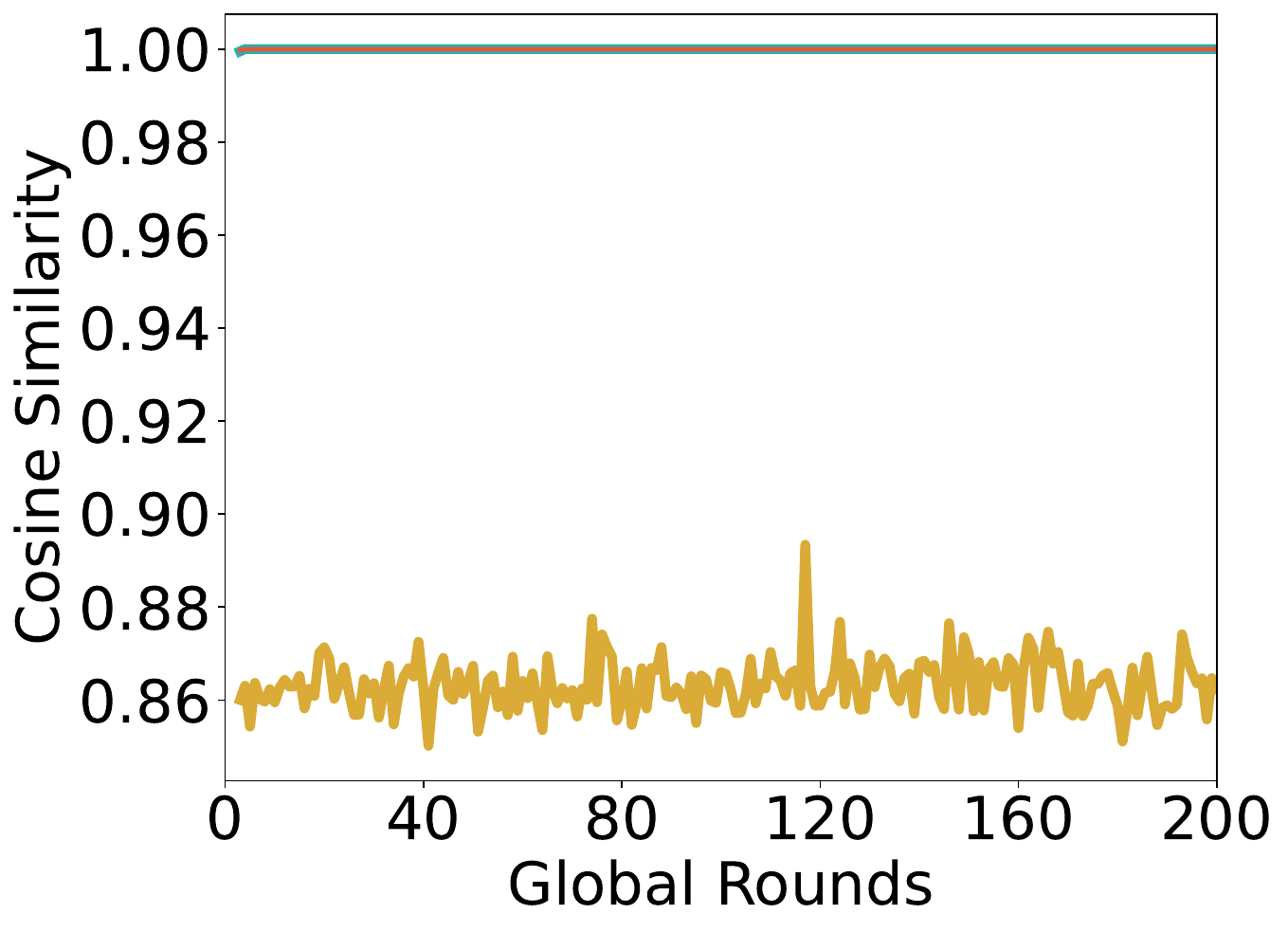}}
	\\
	\subfloat[CIFAR100,80,80\%,15]{
		\includegraphics[scale=0.18]{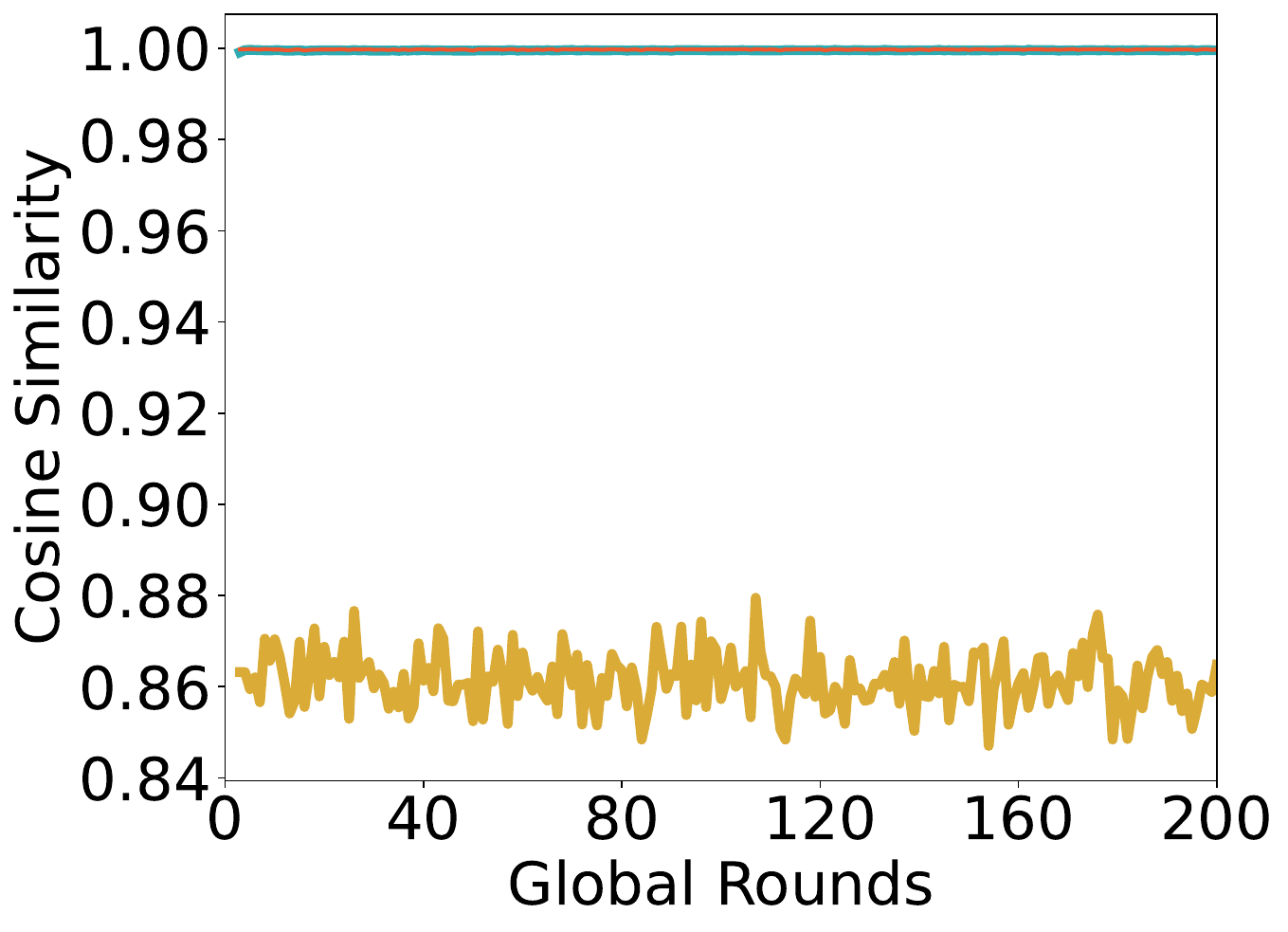}}
	\subfloat[CIFAR100,90,90\%,8]{
		\includegraphics[scale=0.18]{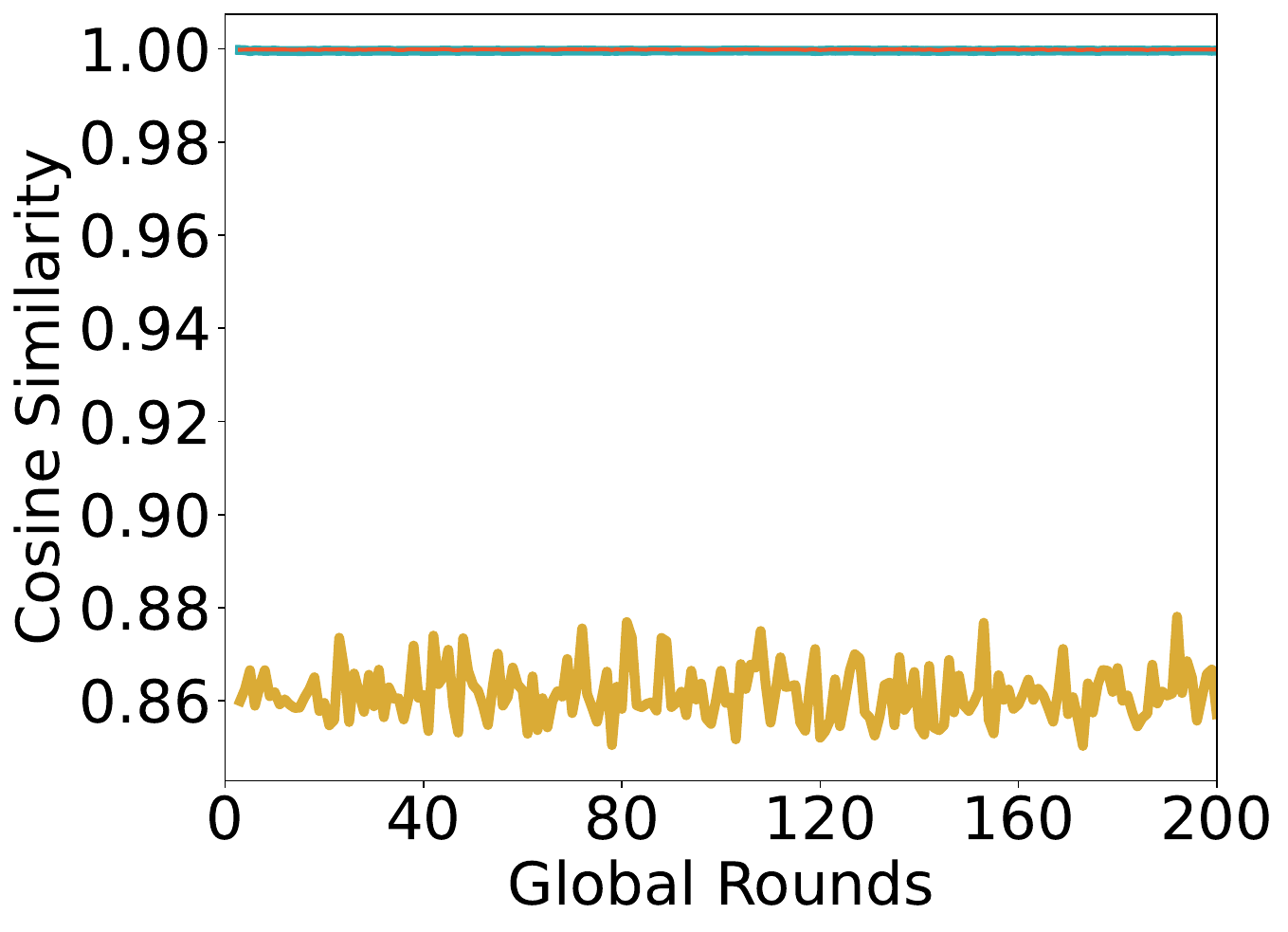}}
	\caption{The performance of VERT in predicting user gradients in the face of large-scale GN attacks in non-IID scenarios.}
	\label{figure4}
\end{figure}

\begin{figure}[h]
	\centering
	\subfloat[MNIST,$|C_t|$=80,$pr$=80\%,$\kappa$=15]{
		\includegraphics[scale=0.18]{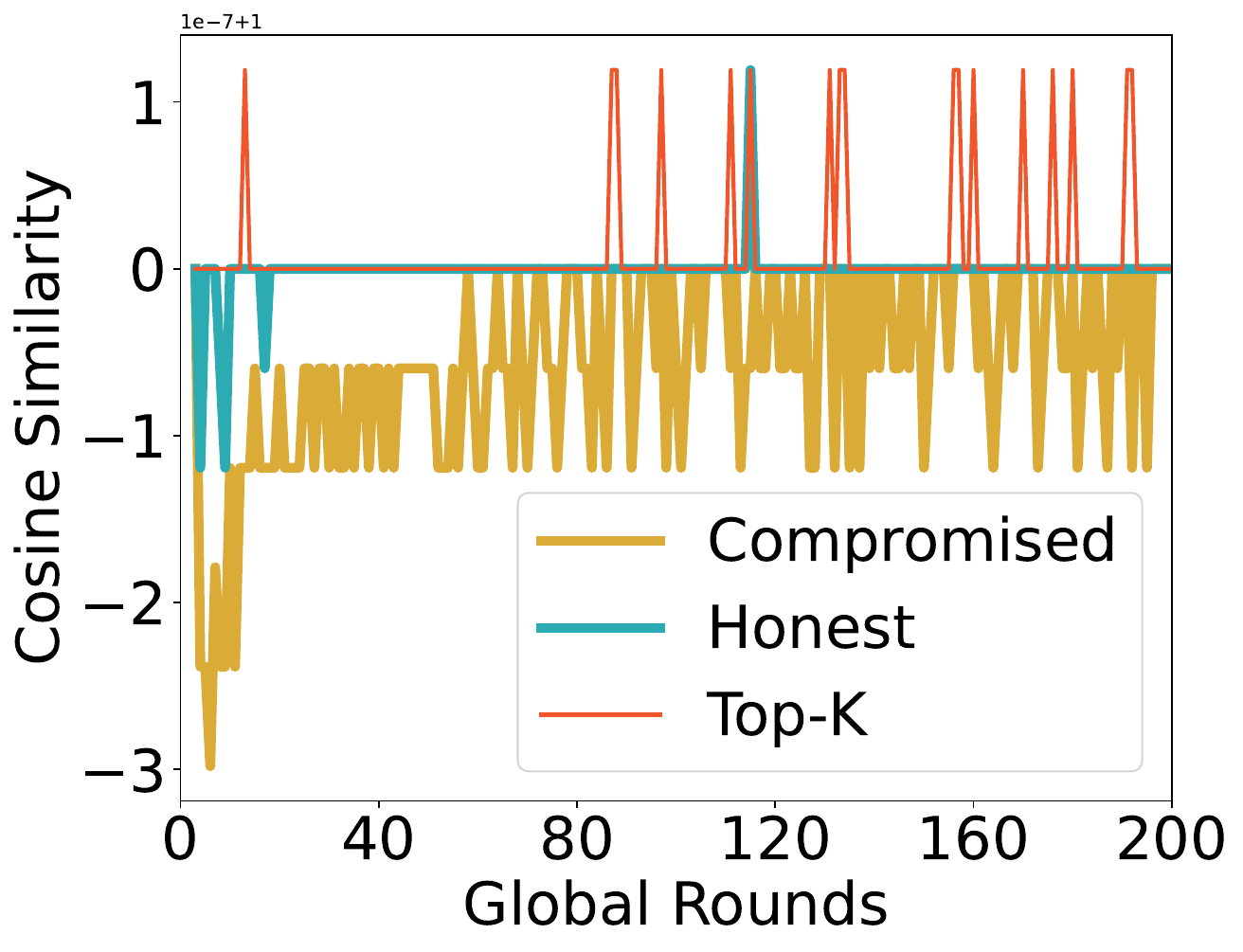}}
	\subfloat[MNIST,90,90\%,8]{
		\includegraphics[scale=0.18]{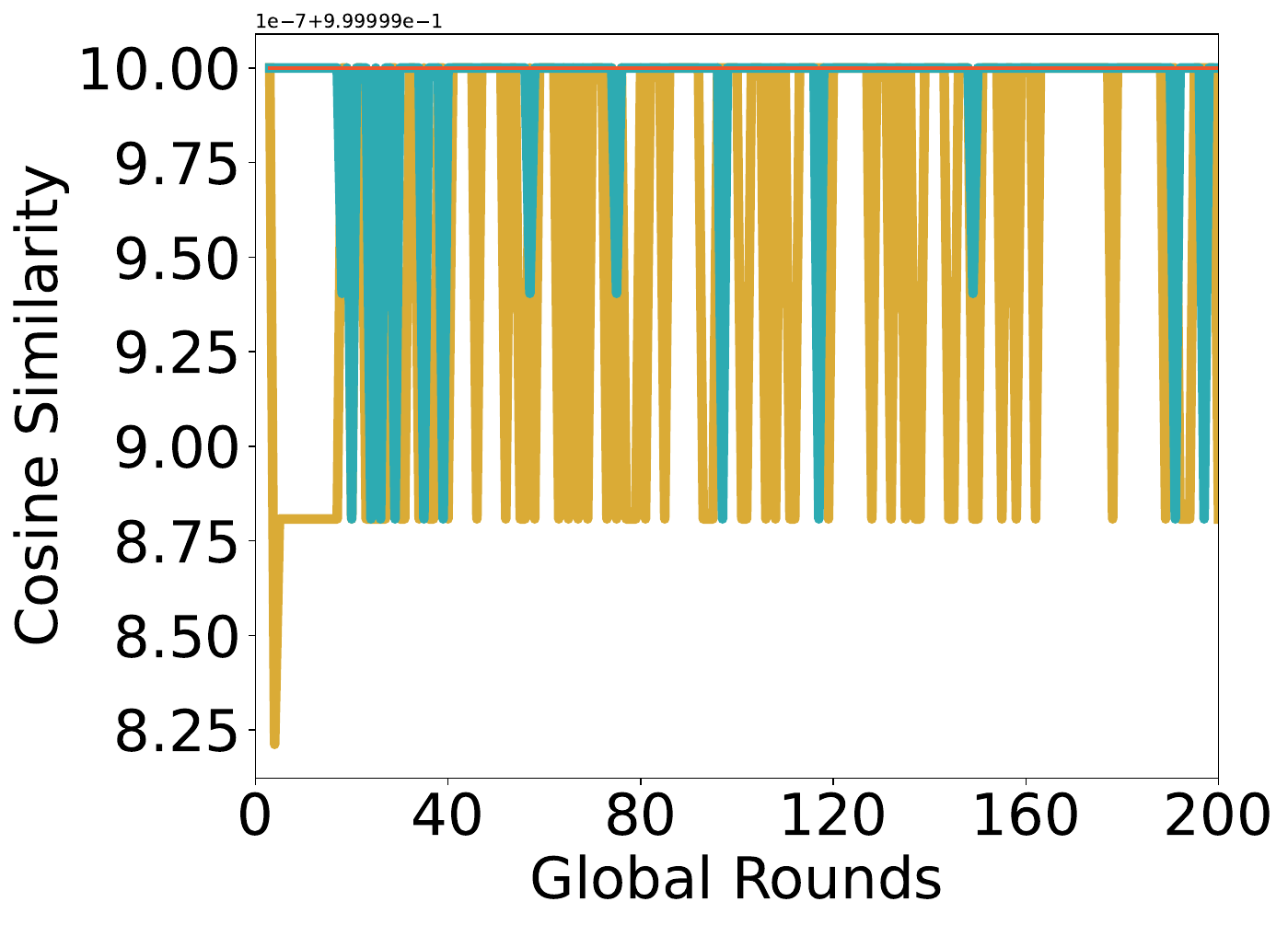}}
	\\
	\subfloat[CIFAR10,80,80\%,15]{
		\includegraphics[scale=0.18]{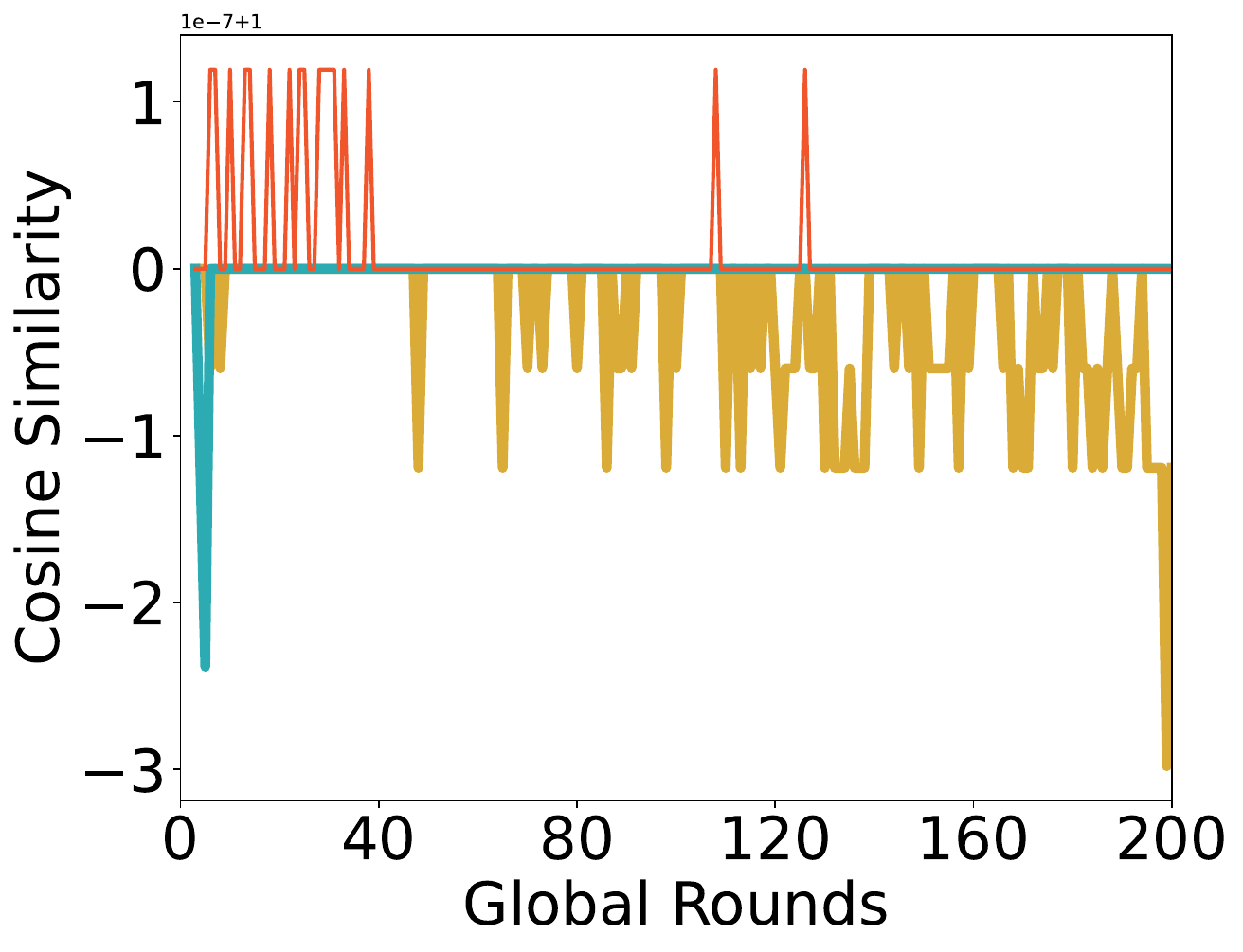}}
	\subfloat[CIFAR10,90,90\%,8]{
		\includegraphics[scale=0.18]{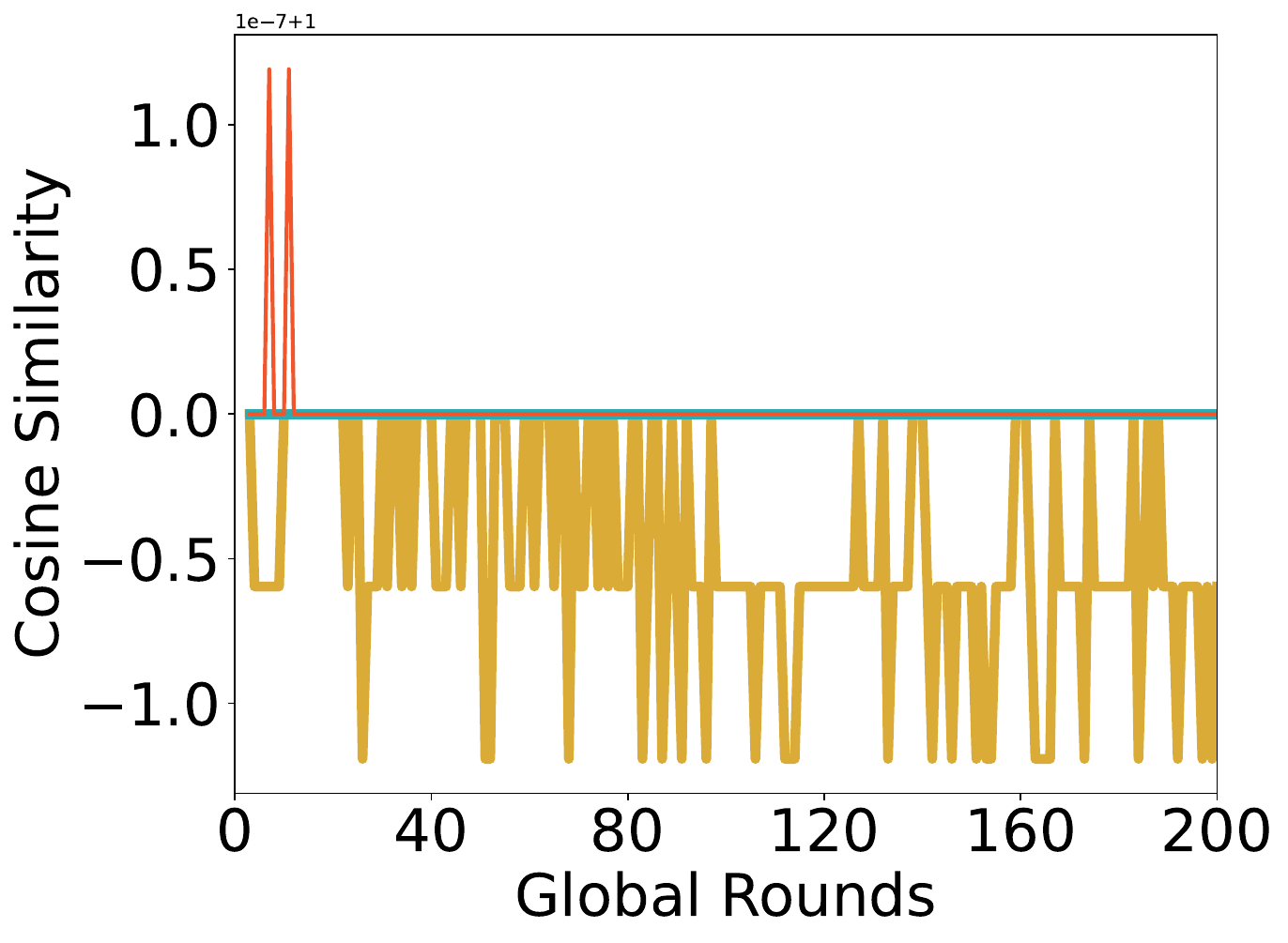}}
	\\
	\subfloat[CIFAR100,80,80\%,15]{
		\includegraphics[scale=0.18]{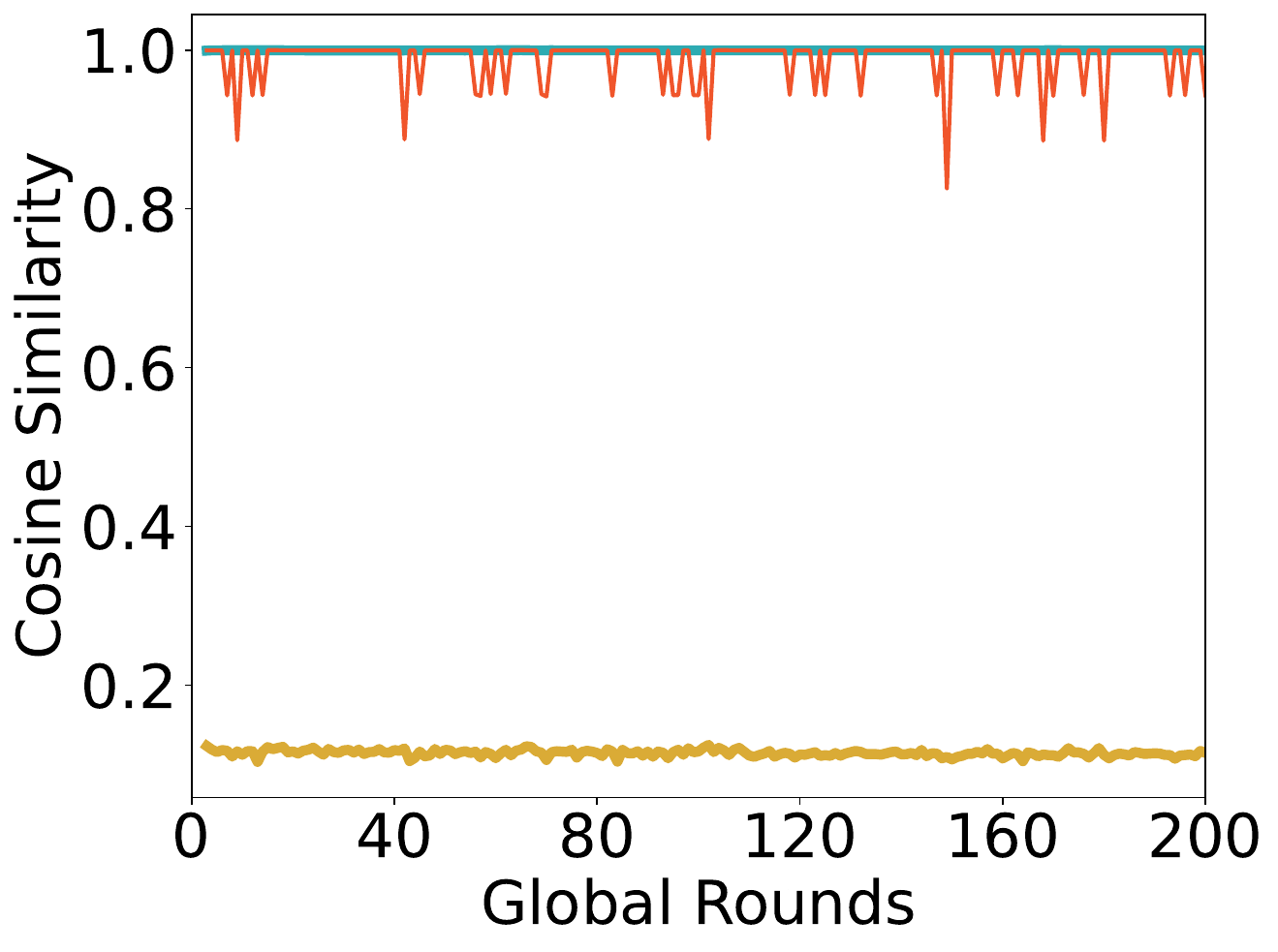}}
	\subfloat[CIFAR100,90,90\%,8]{
		\includegraphics[scale=0.18]{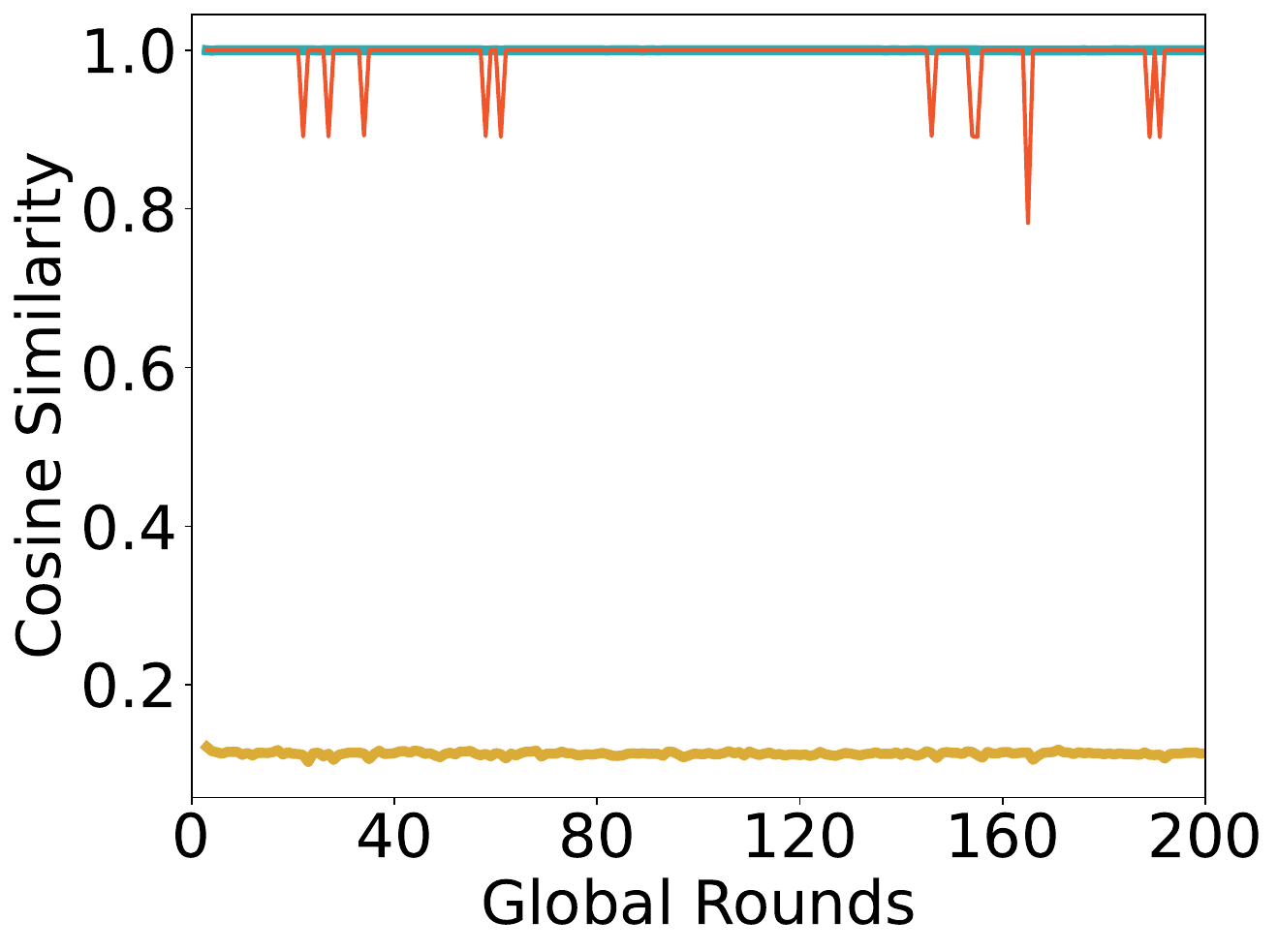}}
	\caption{The performance of VERT in predicting user gradients in the face of large-scale MR attacks in IID scenarios.}
	\label{figure5}
\end{figure}

\begin{figure}[h]
	\centering
	\subfloat[MNIST,$|C_t|$=80,$pr$=80\%,$\kappa$=15]{
		\includegraphics[scale=0.18]{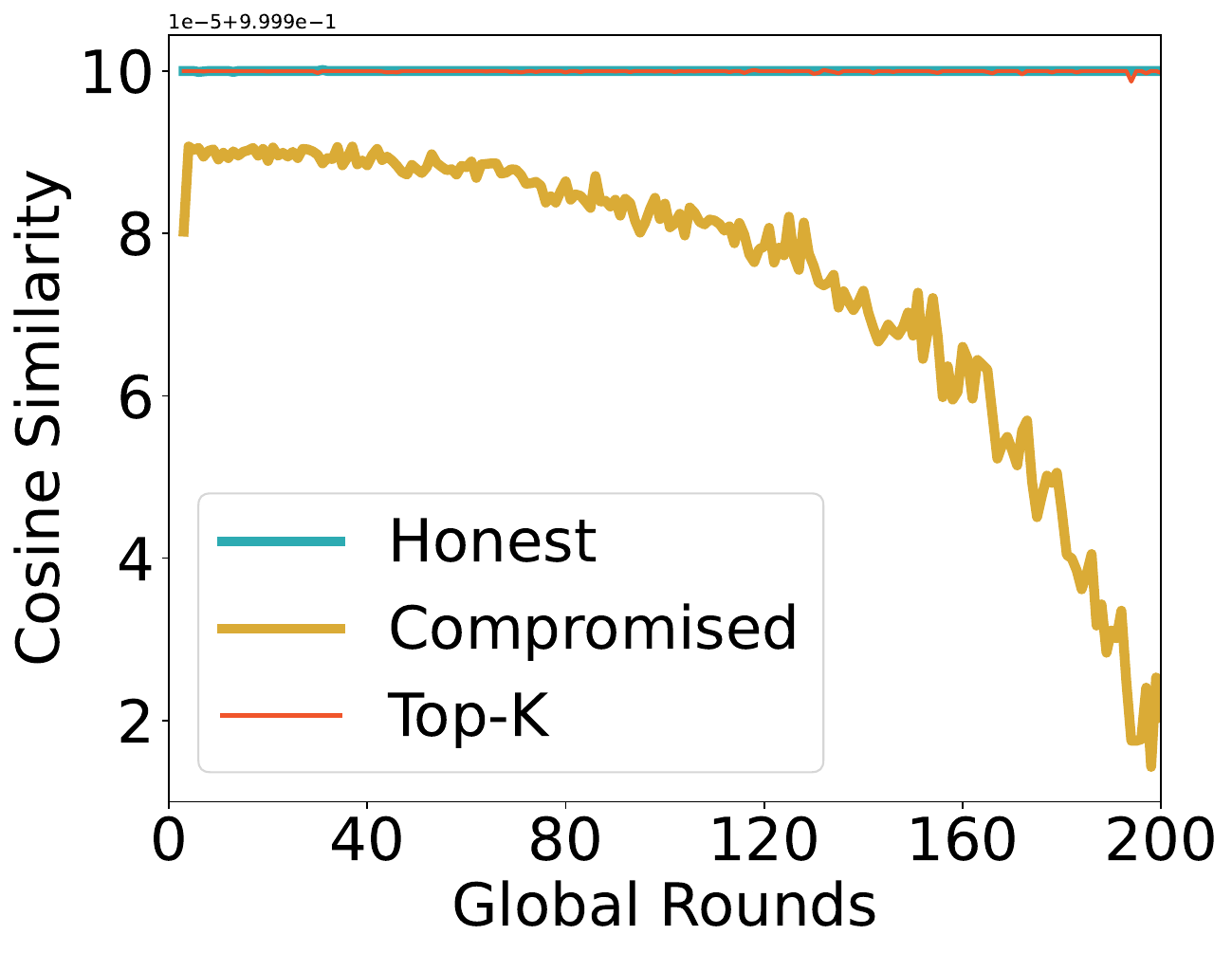}}
	\subfloat[MNIST,90,90\%,8]{
		\includegraphics[scale=0.18]{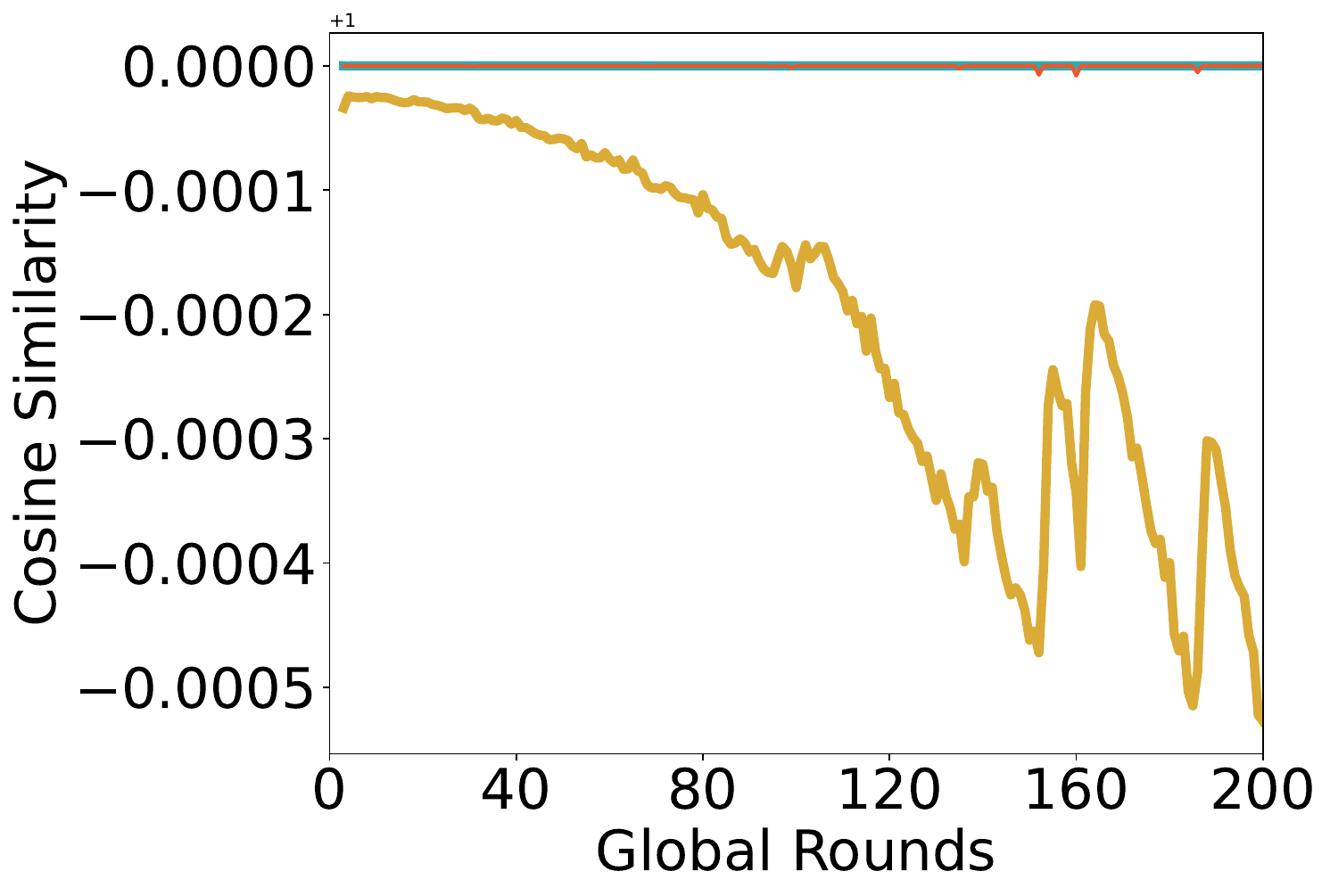}}
	\\
	\subfloat[CIFAR10,80,80\%,15]{
		\includegraphics[scale=0.18]{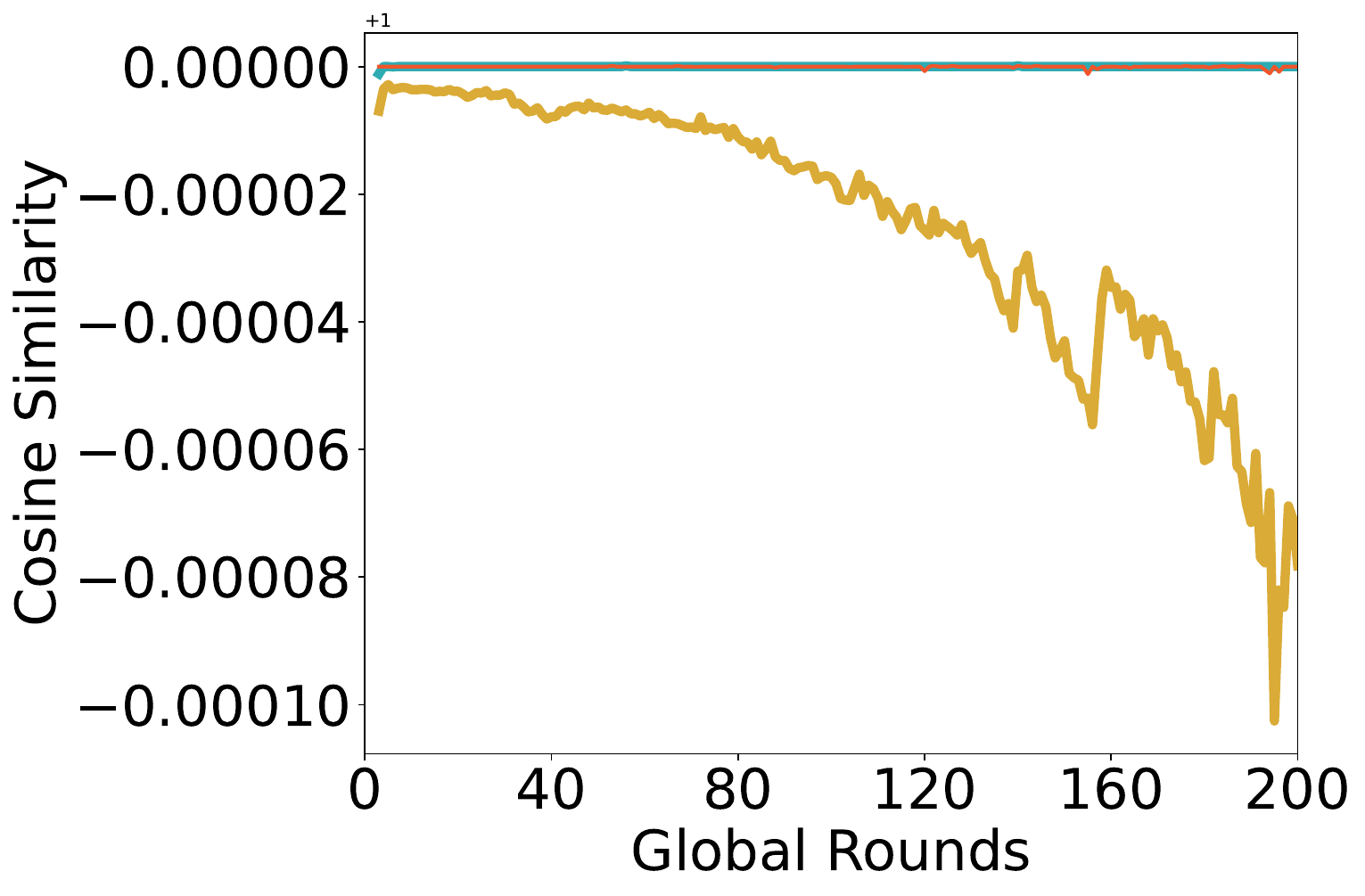}}
	\subfloat[CIFAR10,90,90\%,8]{
		\includegraphics[scale=0.18]{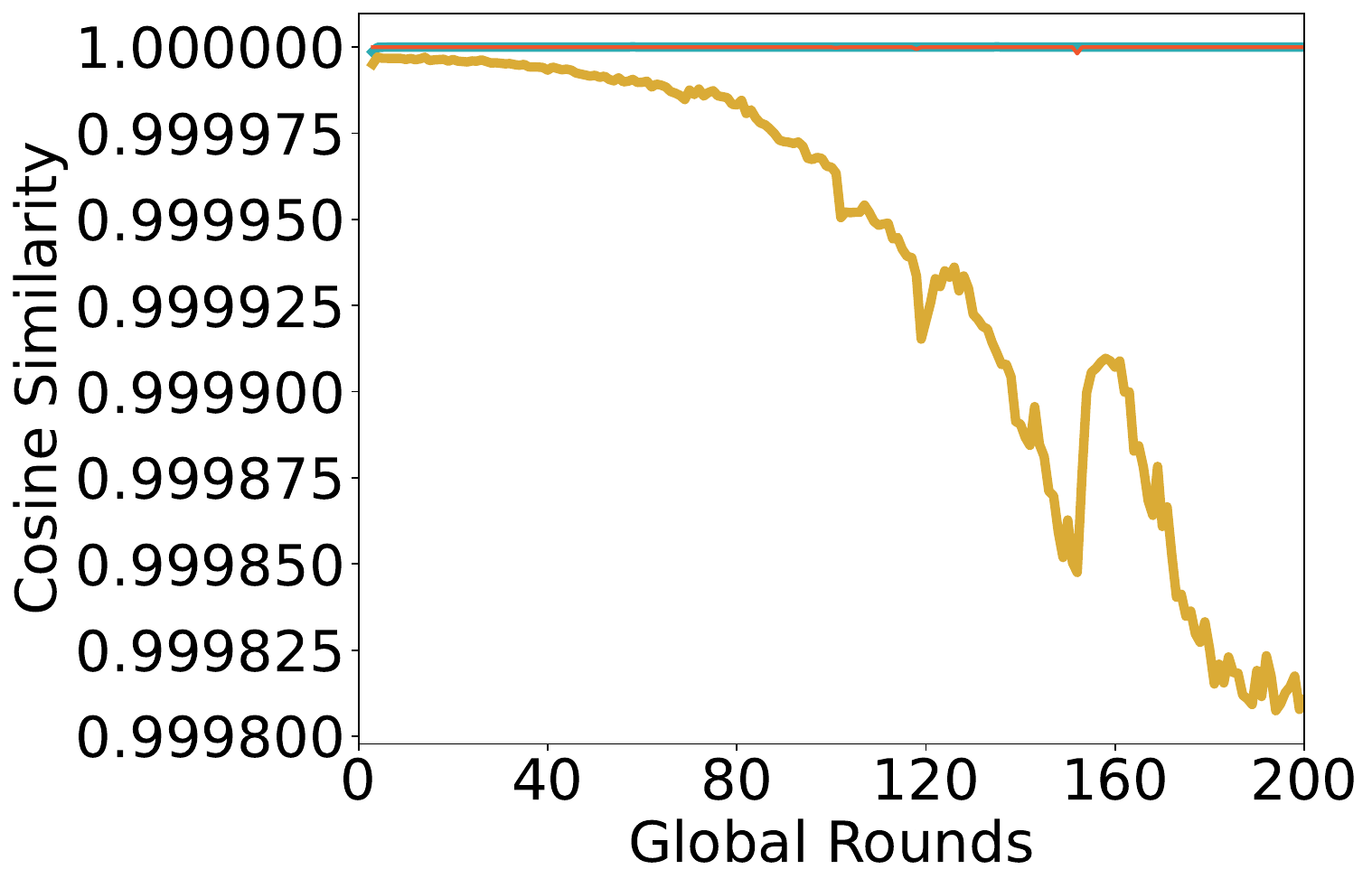}}
	\caption{The performance of VERT in predicting user gradients in the face of large-scale MR attacks in non-IID scenarios.}
	\label{figure6}
\end{figure}

\begin{figure}[h]
	\centering
	\subfloat[MNIST,$|C_t|$=80,$pr$=80\%,$\kappa$=15]{
		\includegraphics[scale=0.18]{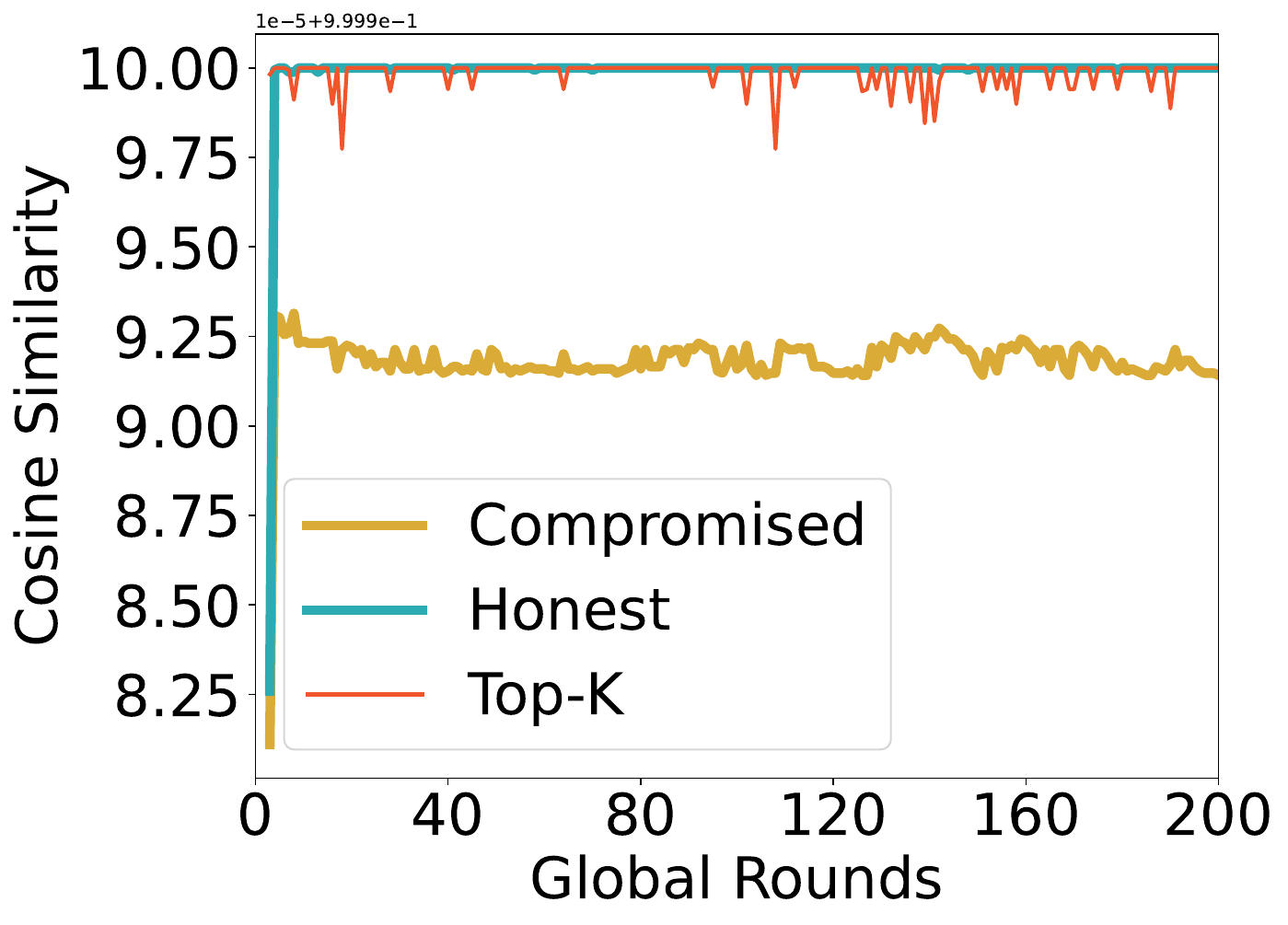}}
	\subfloat[MNIST,90,90\%,8]{
		\includegraphics[scale=0.18]{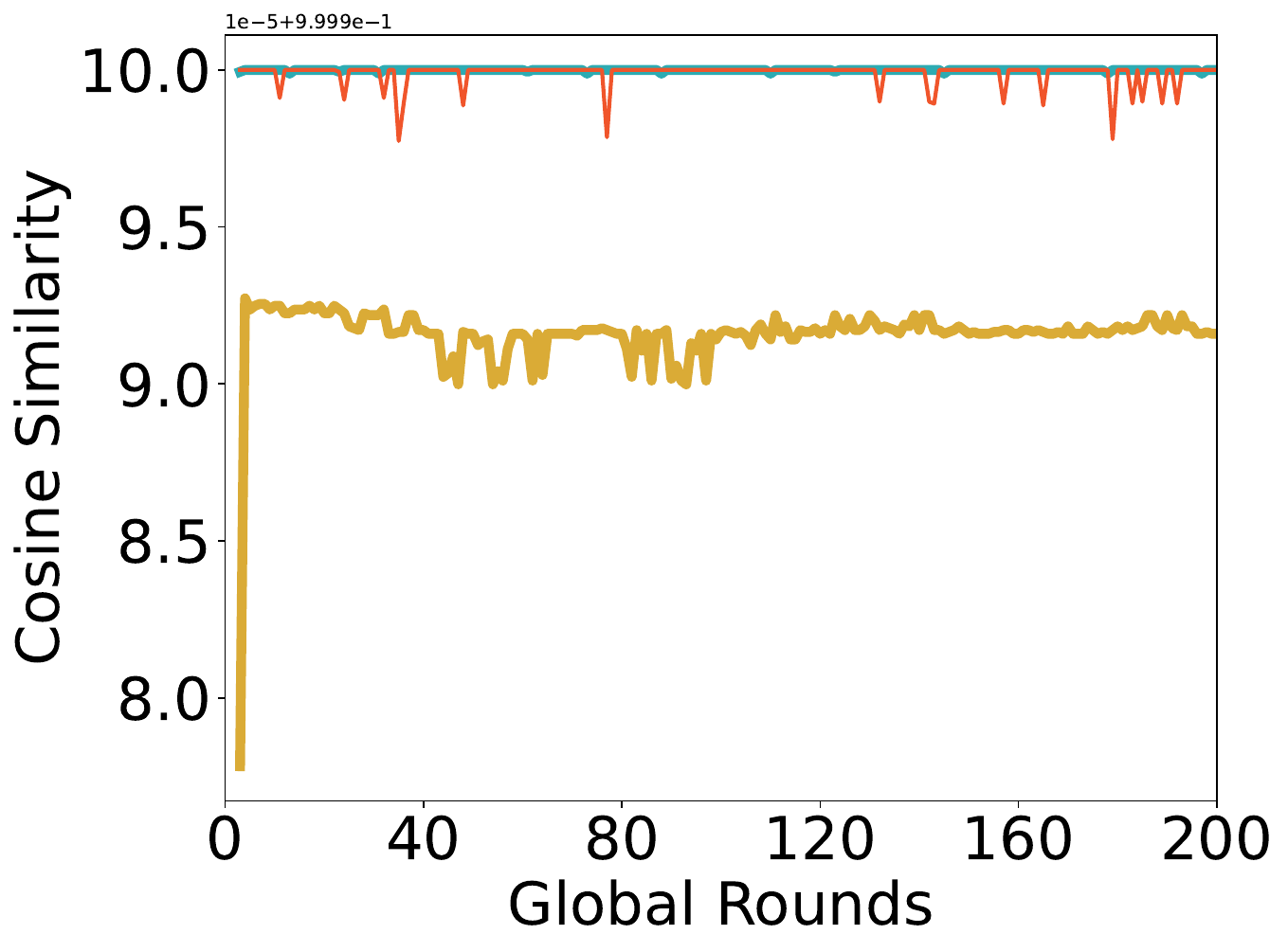}}
	\\
	\subfloat[CIFAR10,80,80\%,15]{
		\includegraphics[scale=0.18]{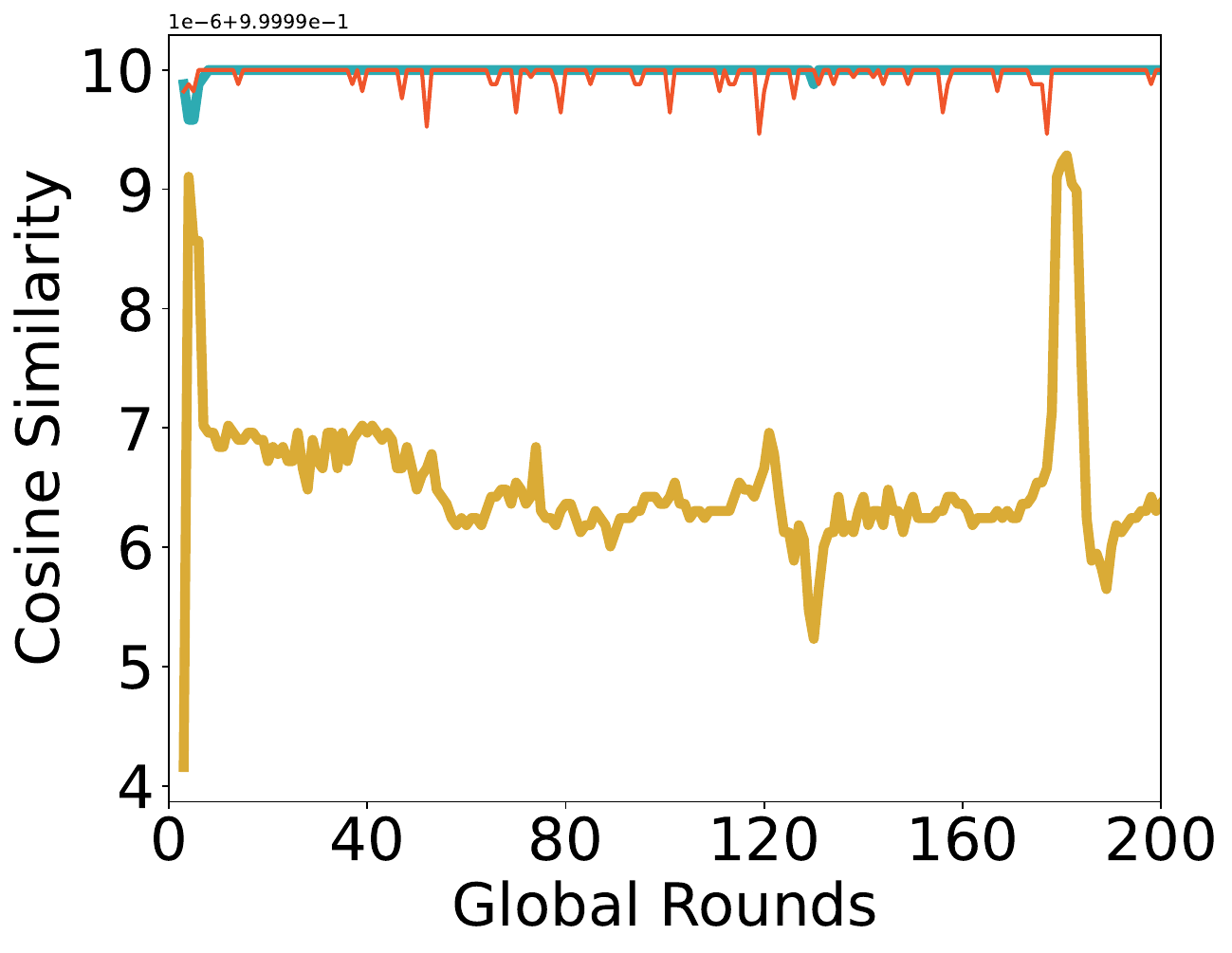}}
	\subfloat[CIFAR10,90,90\%,8]{
		\includegraphics[scale=0.18]{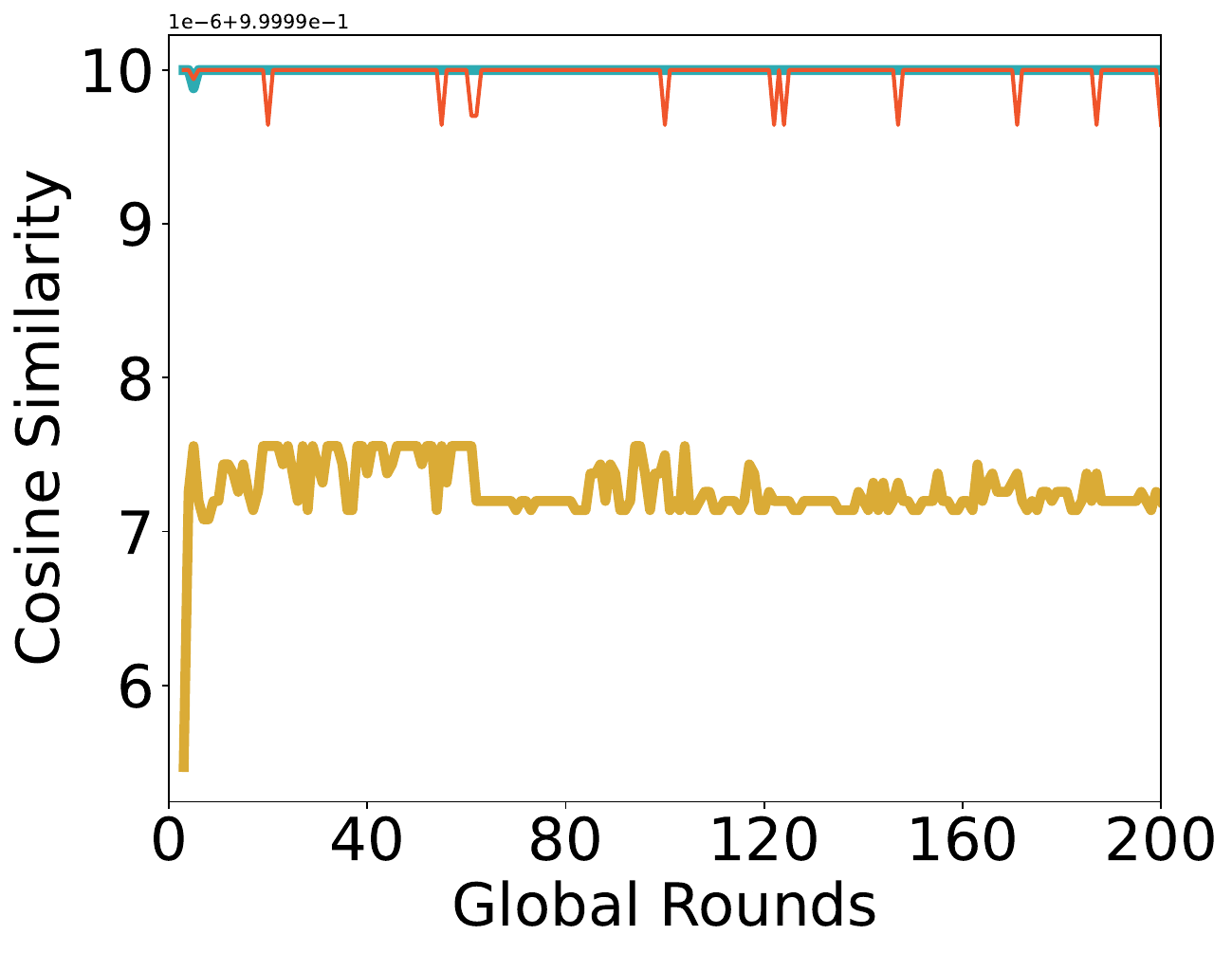}}
	\\
	\subfloat[CIFAR100,80,80\%,15]{
		\includegraphics[scale=0.18]{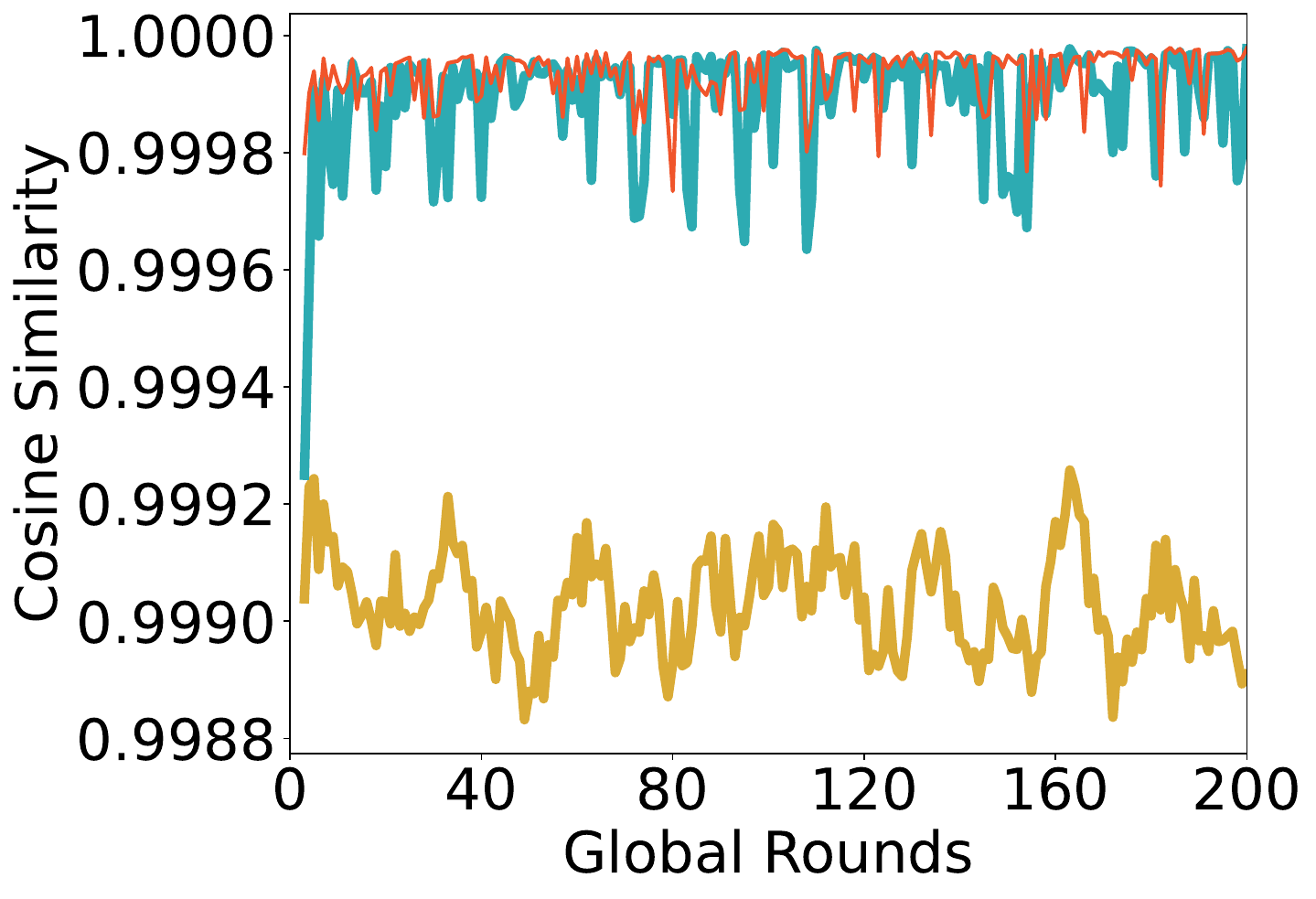}}
	\subfloat[CIFAR100,90,90\%,8]{
		\includegraphics[scale=0.18]{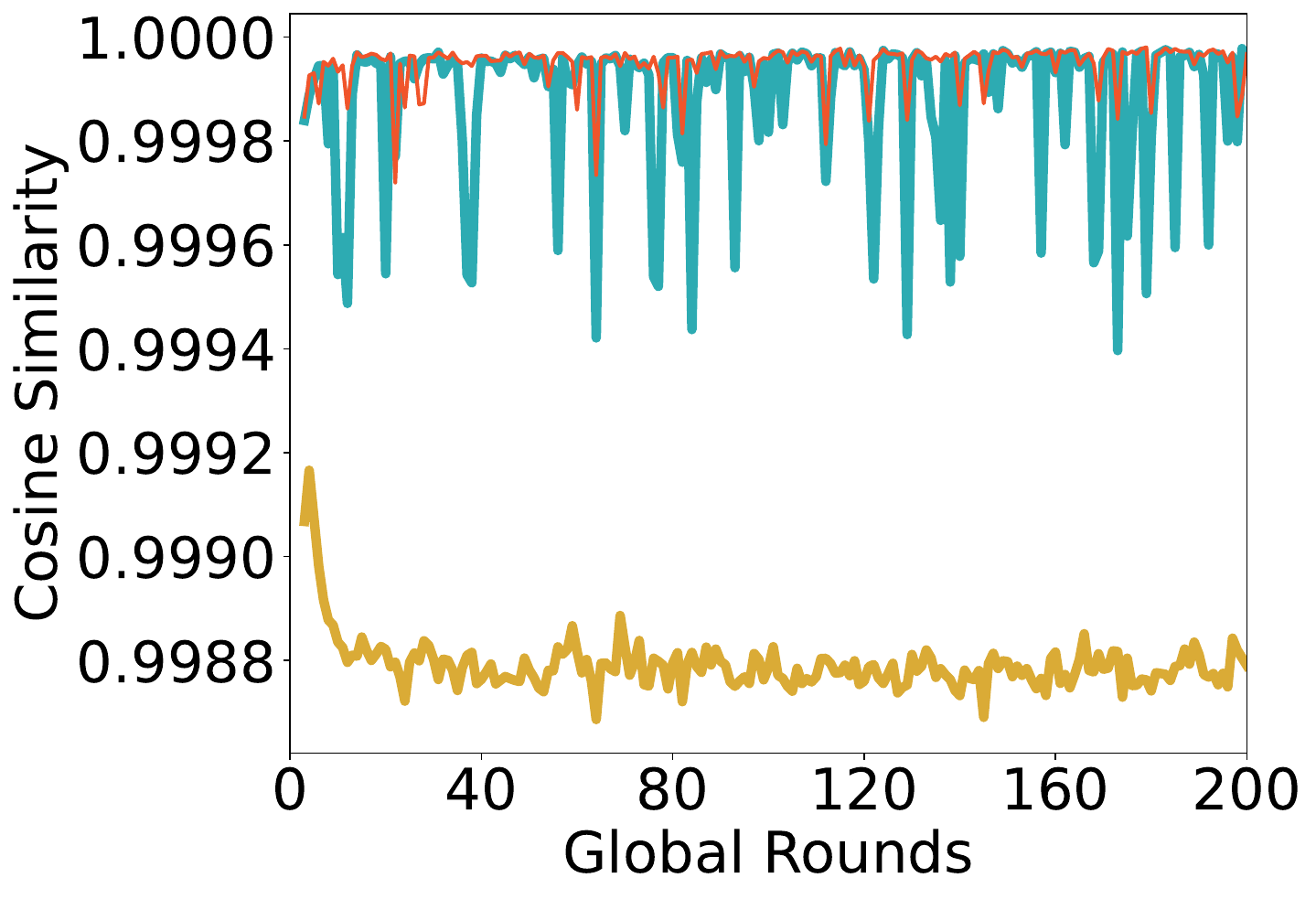}}
	\caption{The performance of VERT in predicting user gradients in the face of large-scale AGR attacks in IID scenarios.}
	\label{figure7}
\end{figure}

\begin{figure}[h]
	\centering
	\subfloat[MNIST,$|C_t|$=80,$pr$=80\%,$\kappa$=15]{
		\includegraphics[scale=0.18]{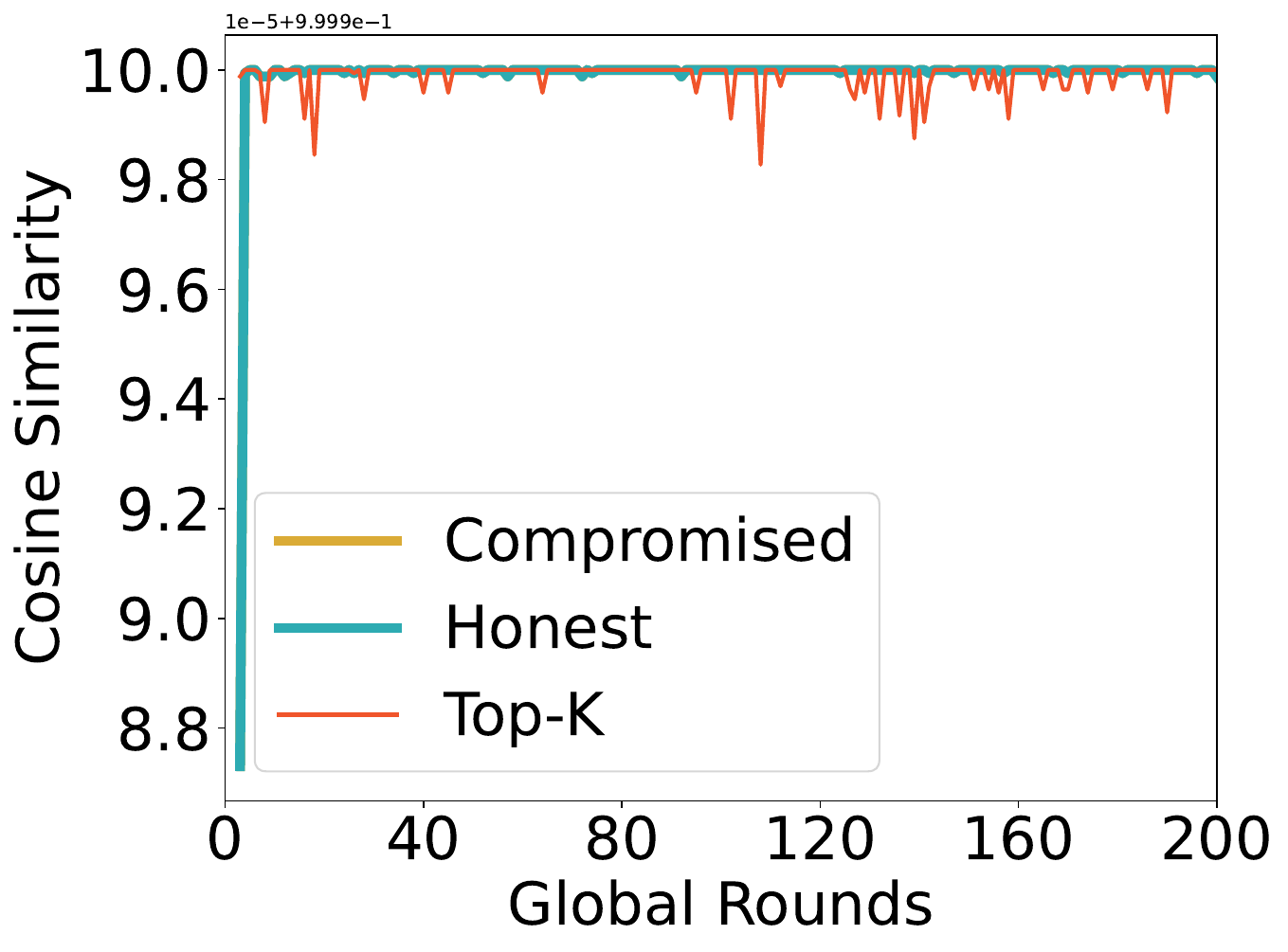}}
	\subfloat[MNIST,90,90\%,8]{
		\includegraphics[scale=0.18]{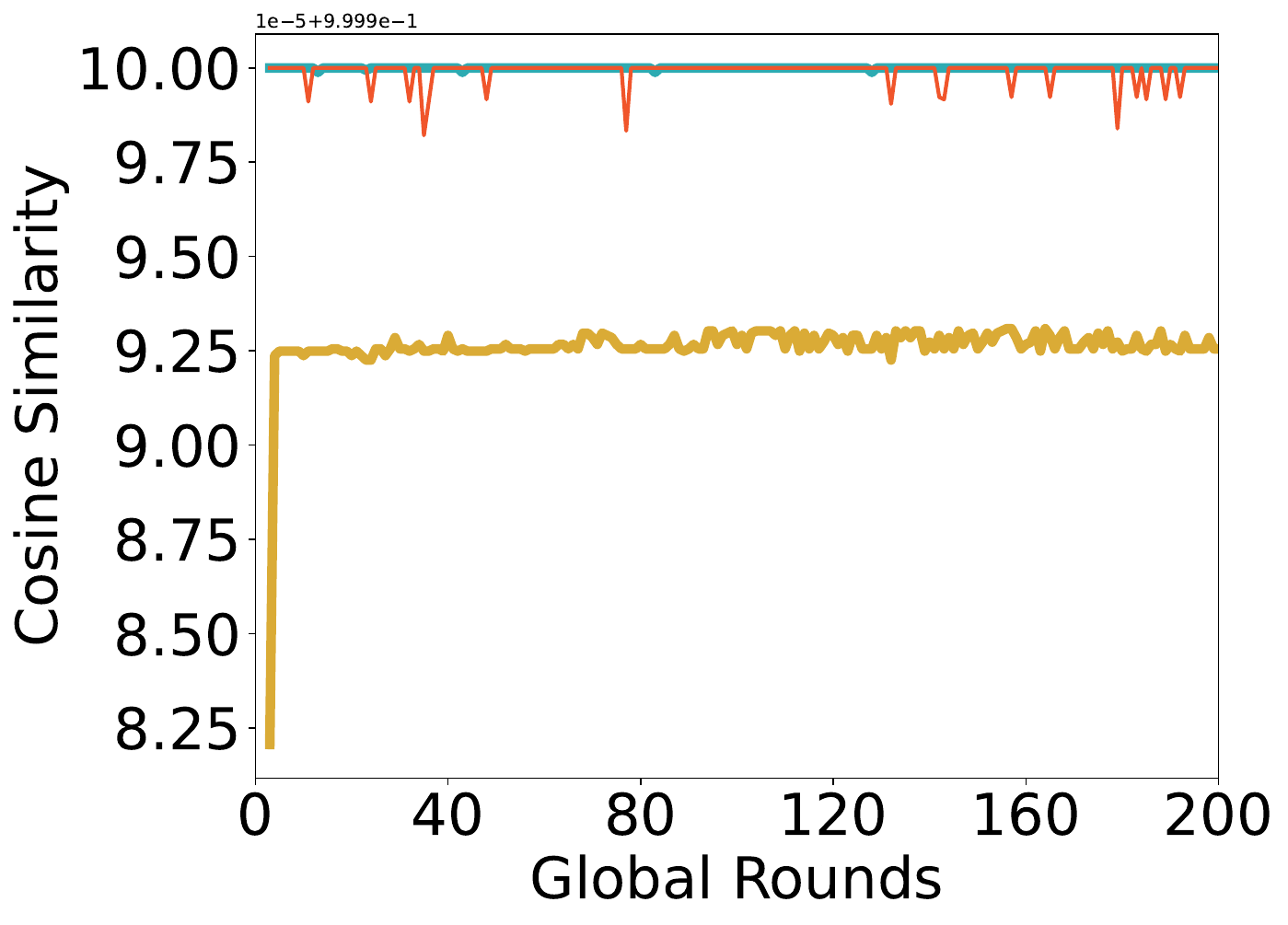}}
	\\
	\subfloat[CIFAR10,80,80\%,15]{
		\includegraphics[scale=0.18]{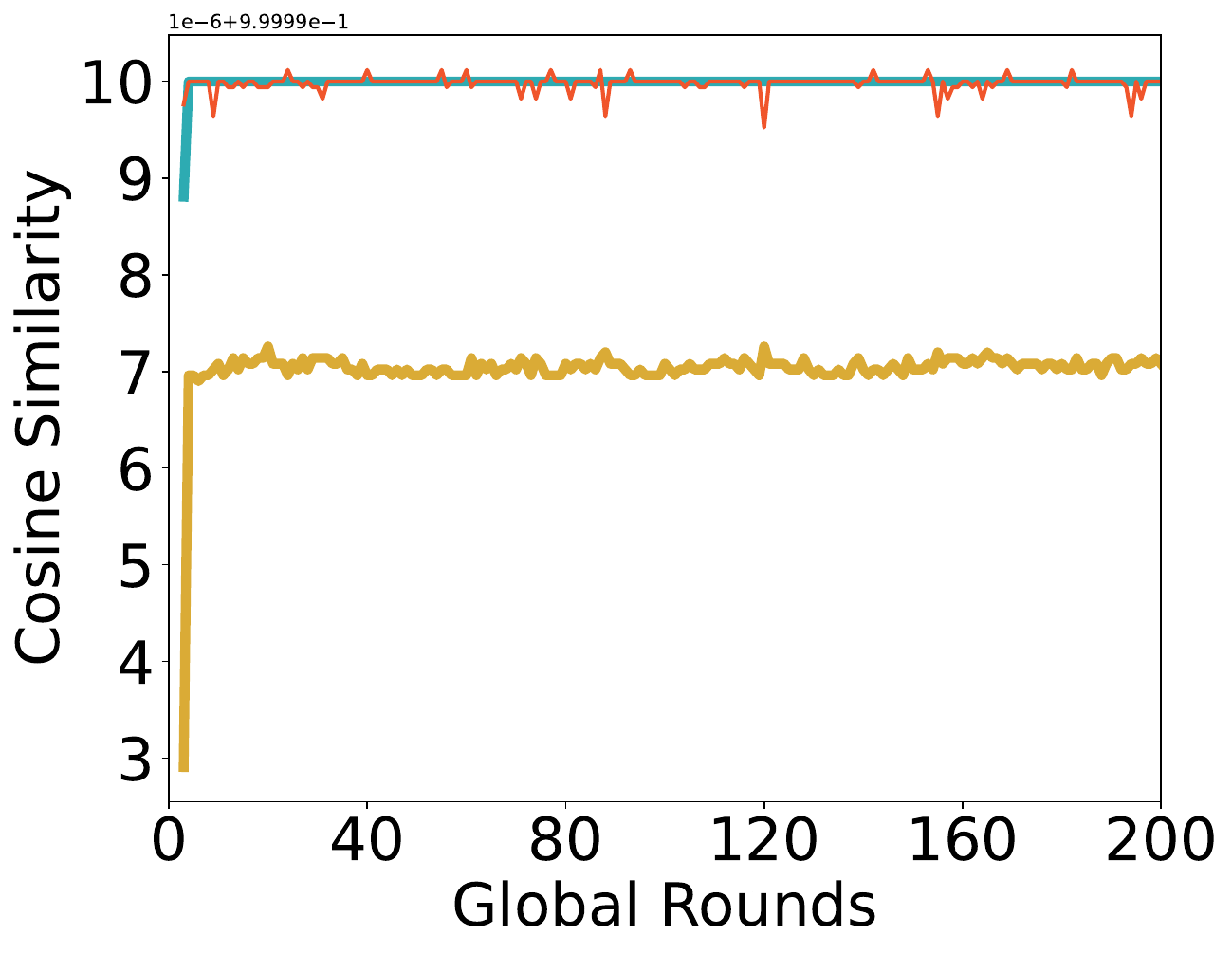}}
	\subfloat[CIFAR10,90,90\%,8]{
		\includegraphics[scale=0.18]{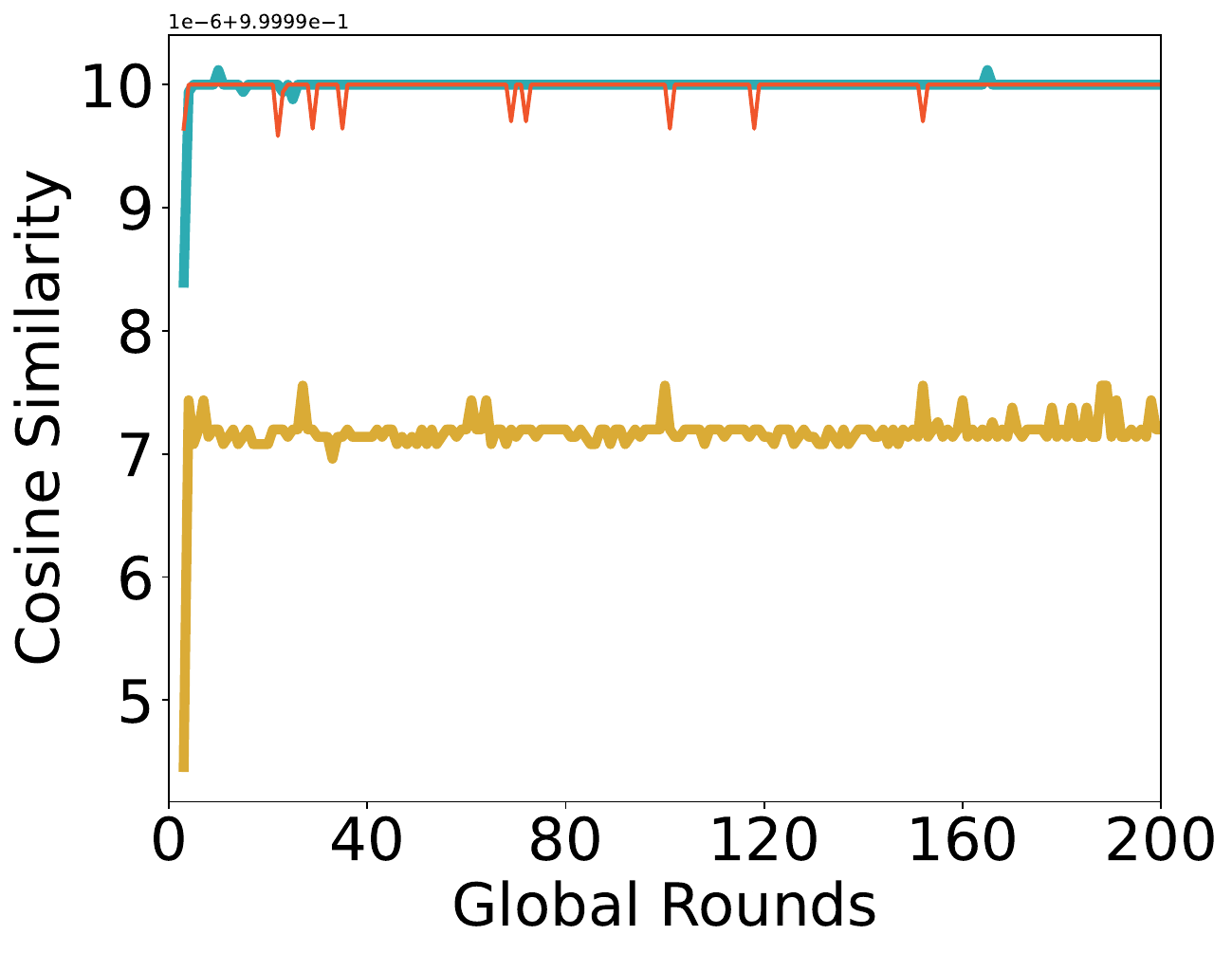}}
	\caption{The performance of VERT in predicting user gradients in the face of large-scale AGR attacks in non-IID scenarios.}
	\label{figure8}
\end{figure}

\begin{figure}[h]
	\centering
	\subfloat[MNIST,$|C_t|$=80,$pr$=80\%,$\kappa$=15]{
		\includegraphics[scale=0.18]{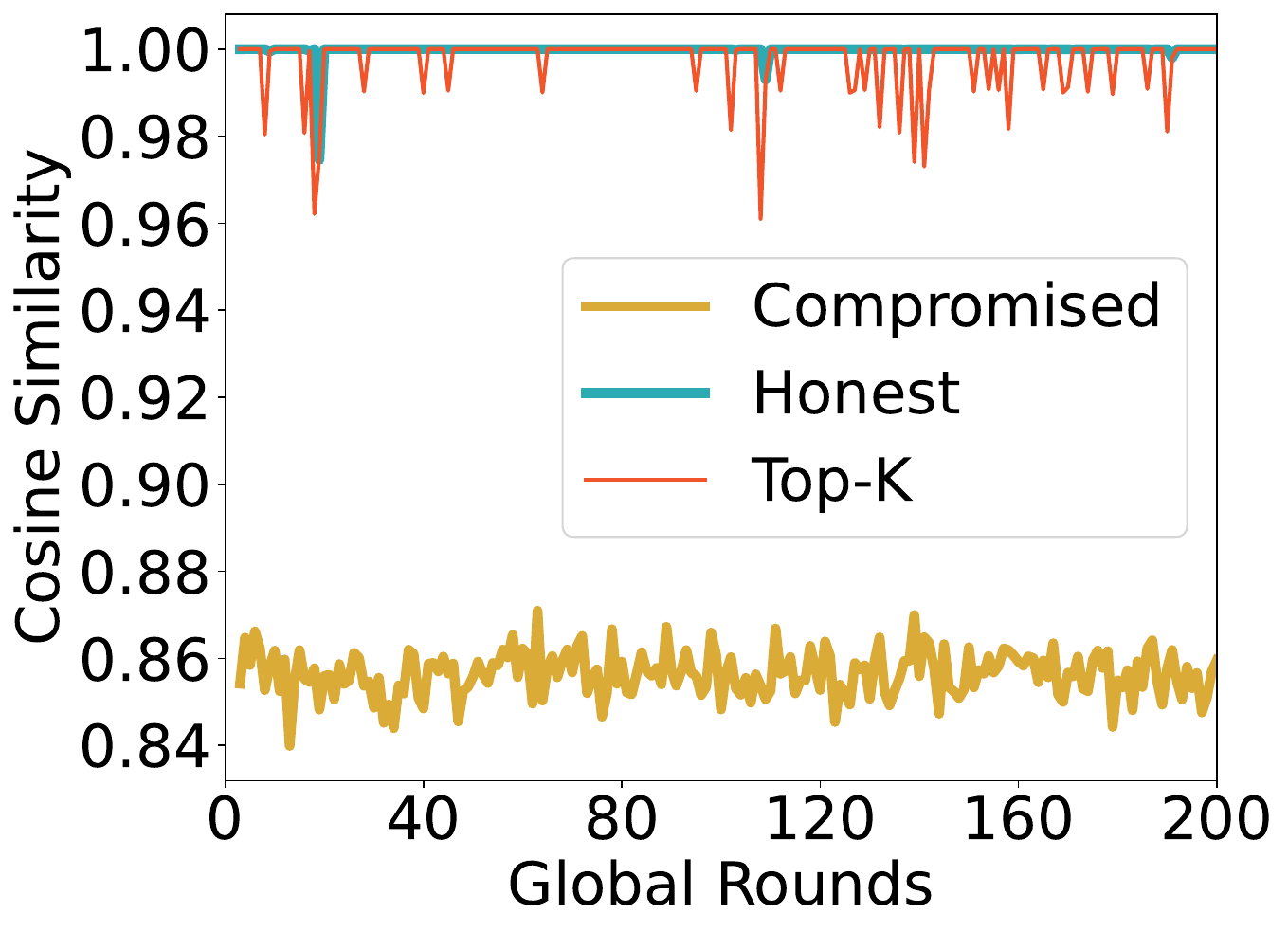}}
	\subfloat[MNIST,90,90\%,8]{
		\includegraphics[scale=0.18]{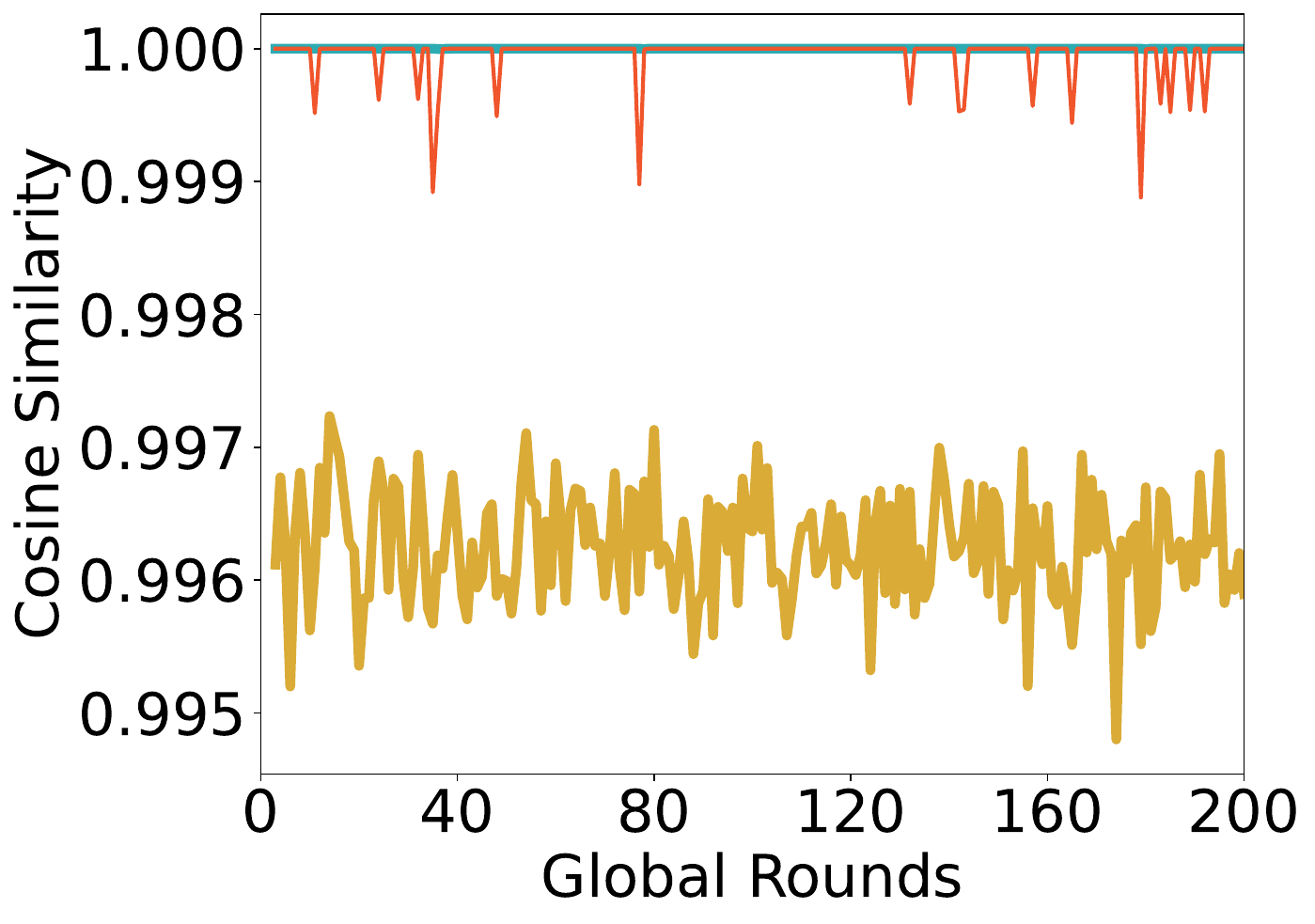}}
	\\
	\subfloat[CIFAR10,80,80\%,15]{
		\includegraphics[scale=0.18]{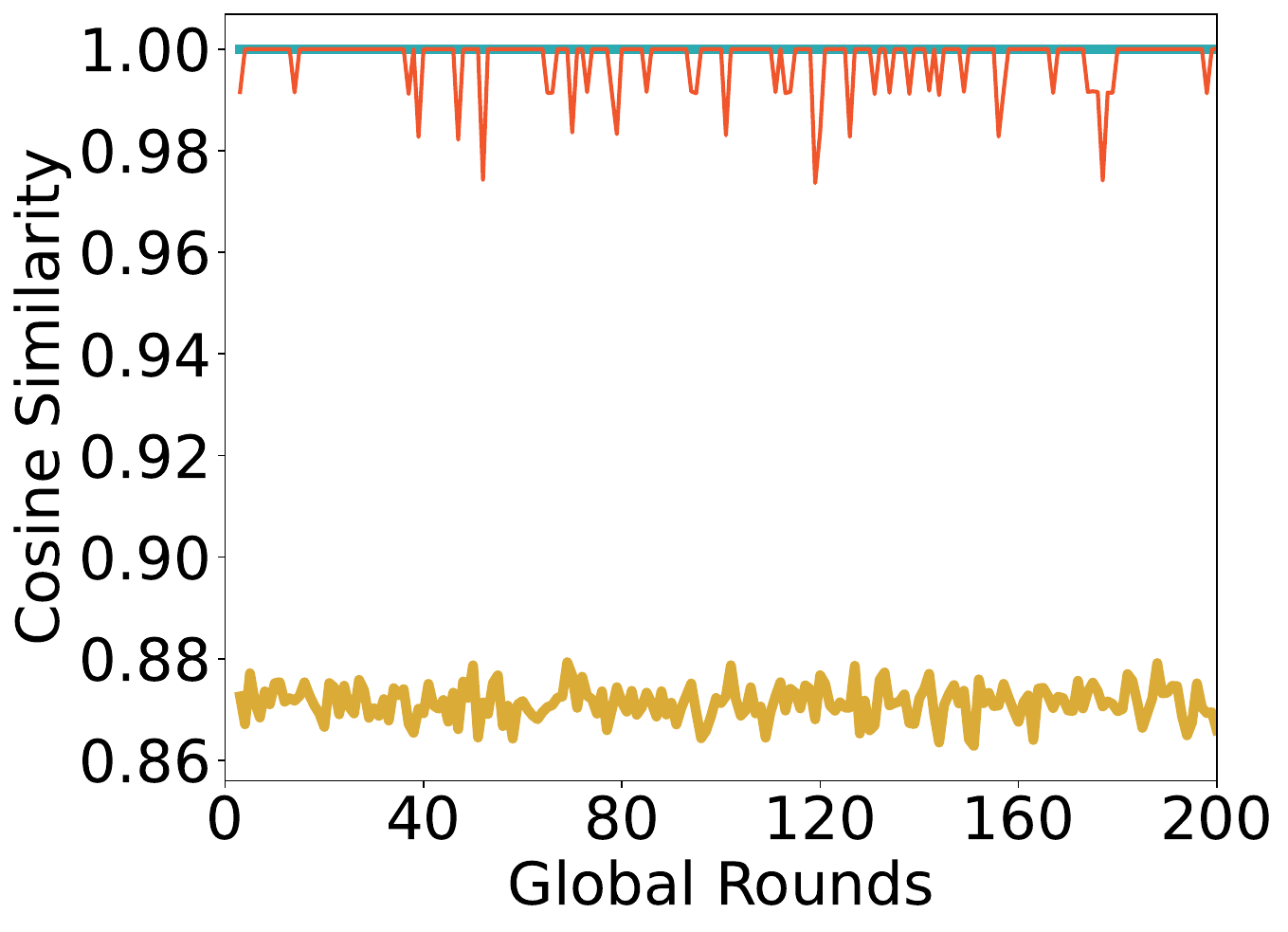}}
	\subfloat[CIFAR10,90,90\%,8]{
		\includegraphics[scale=0.18]{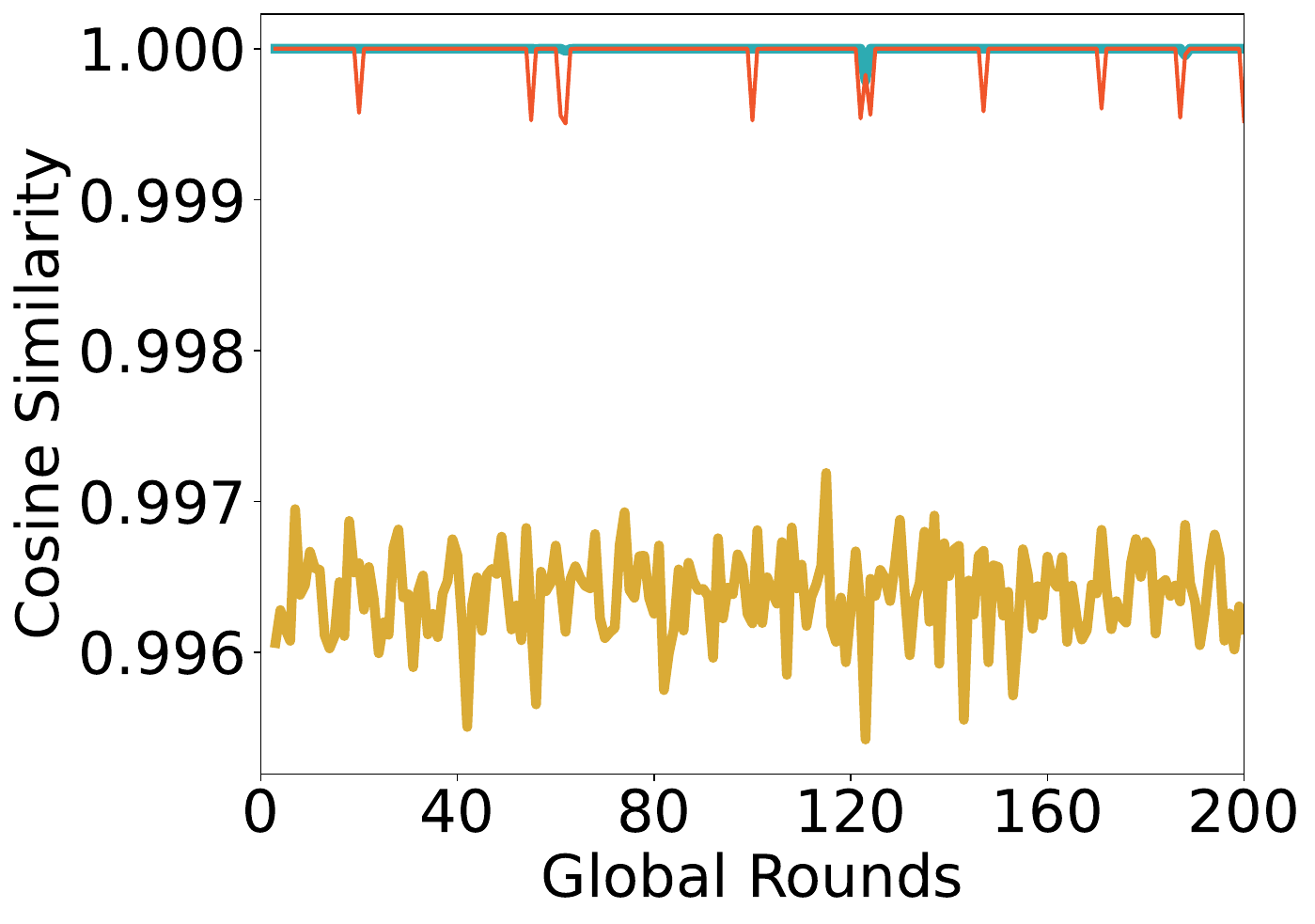}}
	\\
	\subfloat[CIFAR100,80,80\%,15]{
		\includegraphics[scale=0.18]{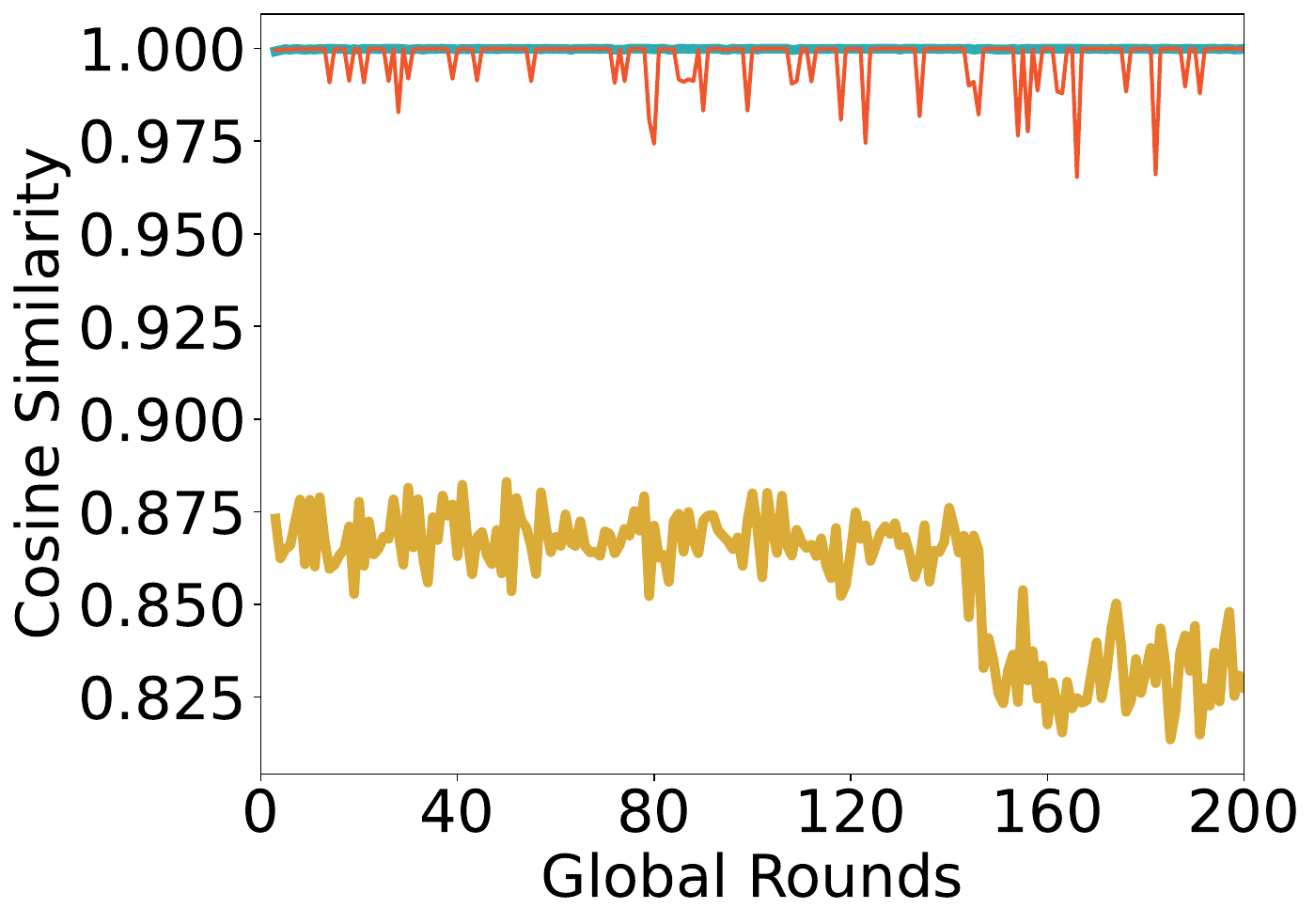}}
	\subfloat[CIFAR100,90,90\%,8]{
		\includegraphics[scale=0.18]{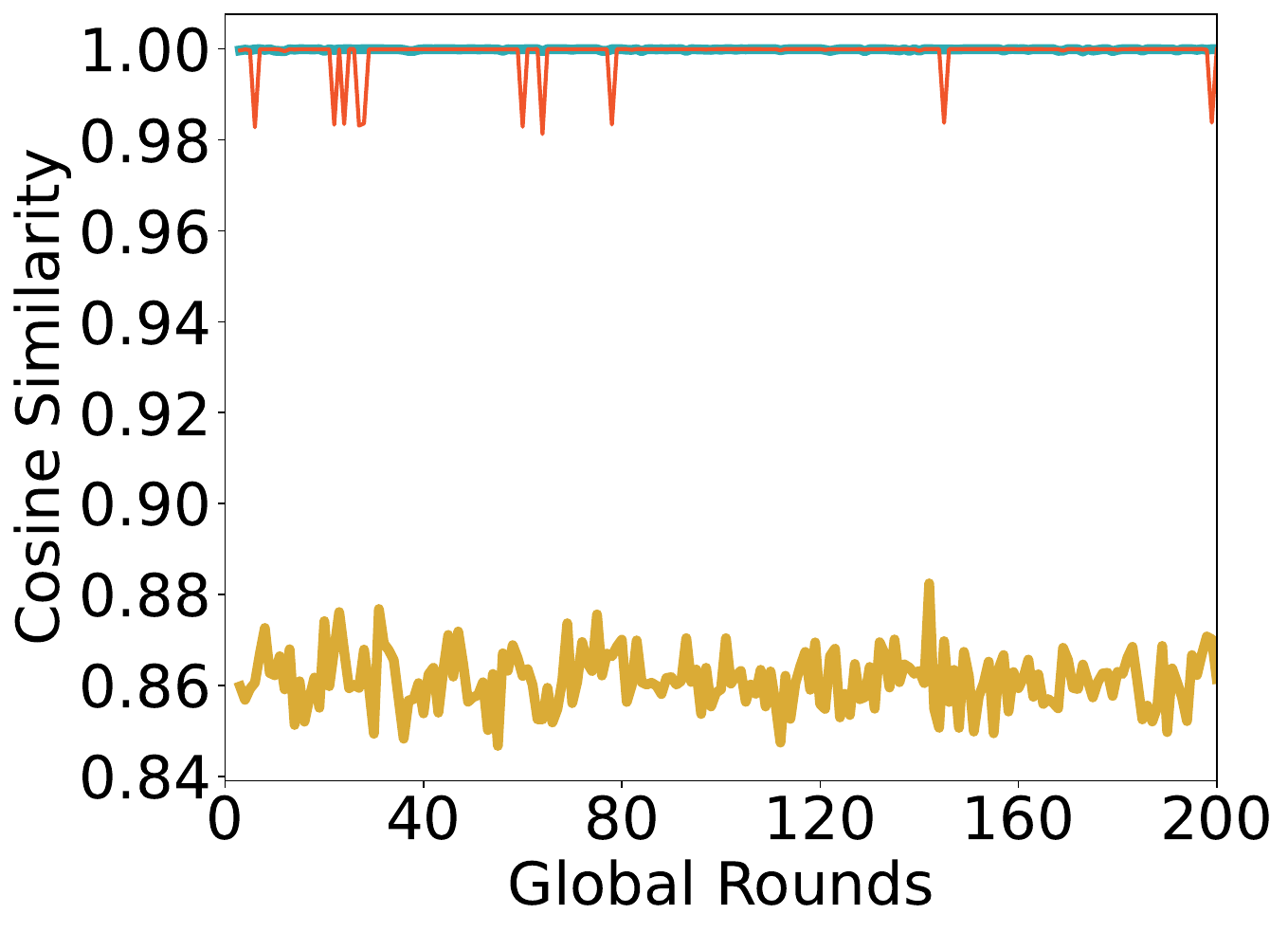}}
	\caption{The performance of VERT in predicting user gradients in the face of large-scale ALIE attacks in IID scenarios.}
	\label{figure9}
\end{figure}

\begin{figure}[h]
	\centering
	\subfloat[MNIST,$|C_t|$=80,$pr$=80\%,$\kappa$=15]{
		\includegraphics[scale=0.18]{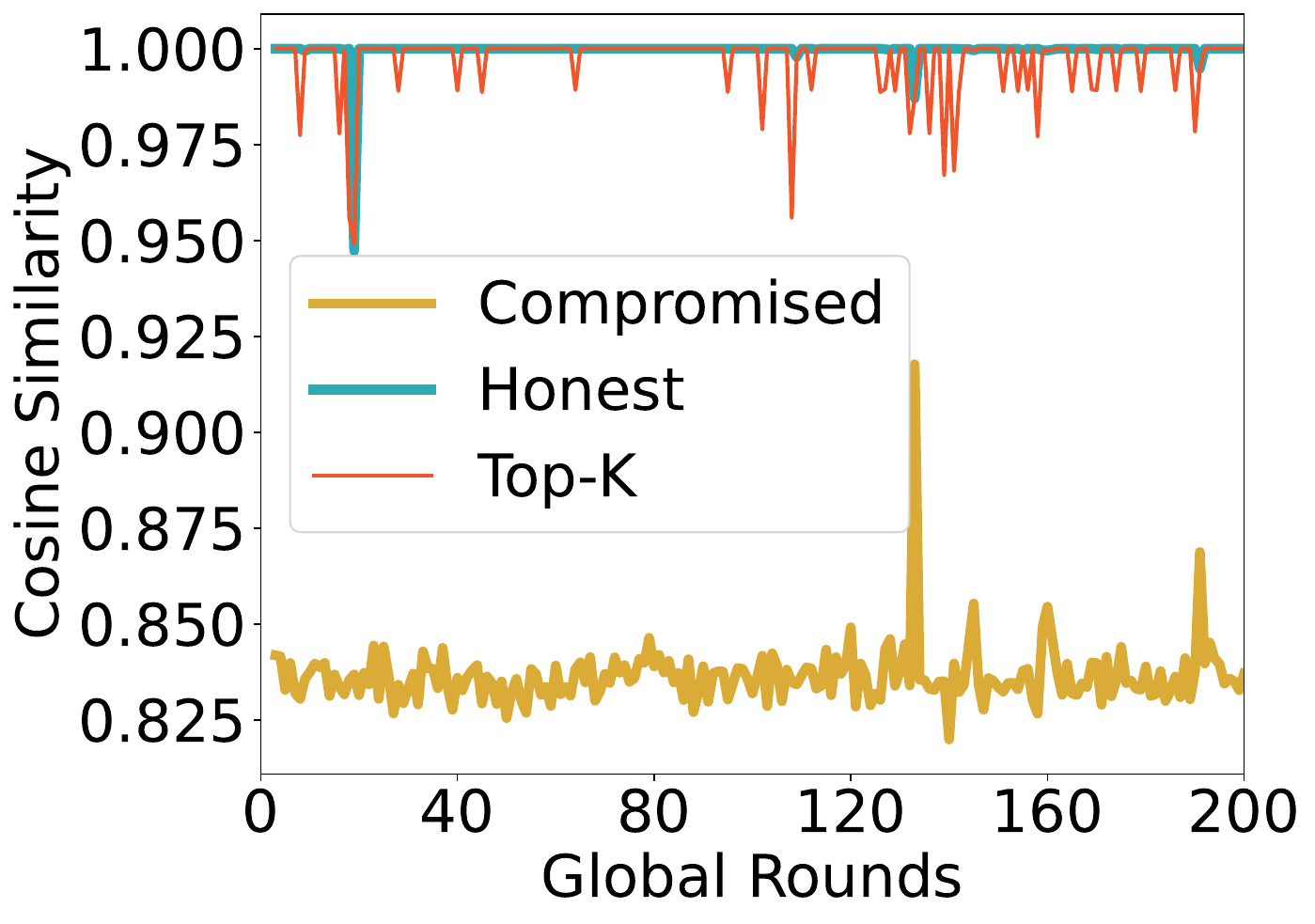}}
	\subfloat[MNIST,90,90\%,8]{
		\includegraphics[scale=0.18]{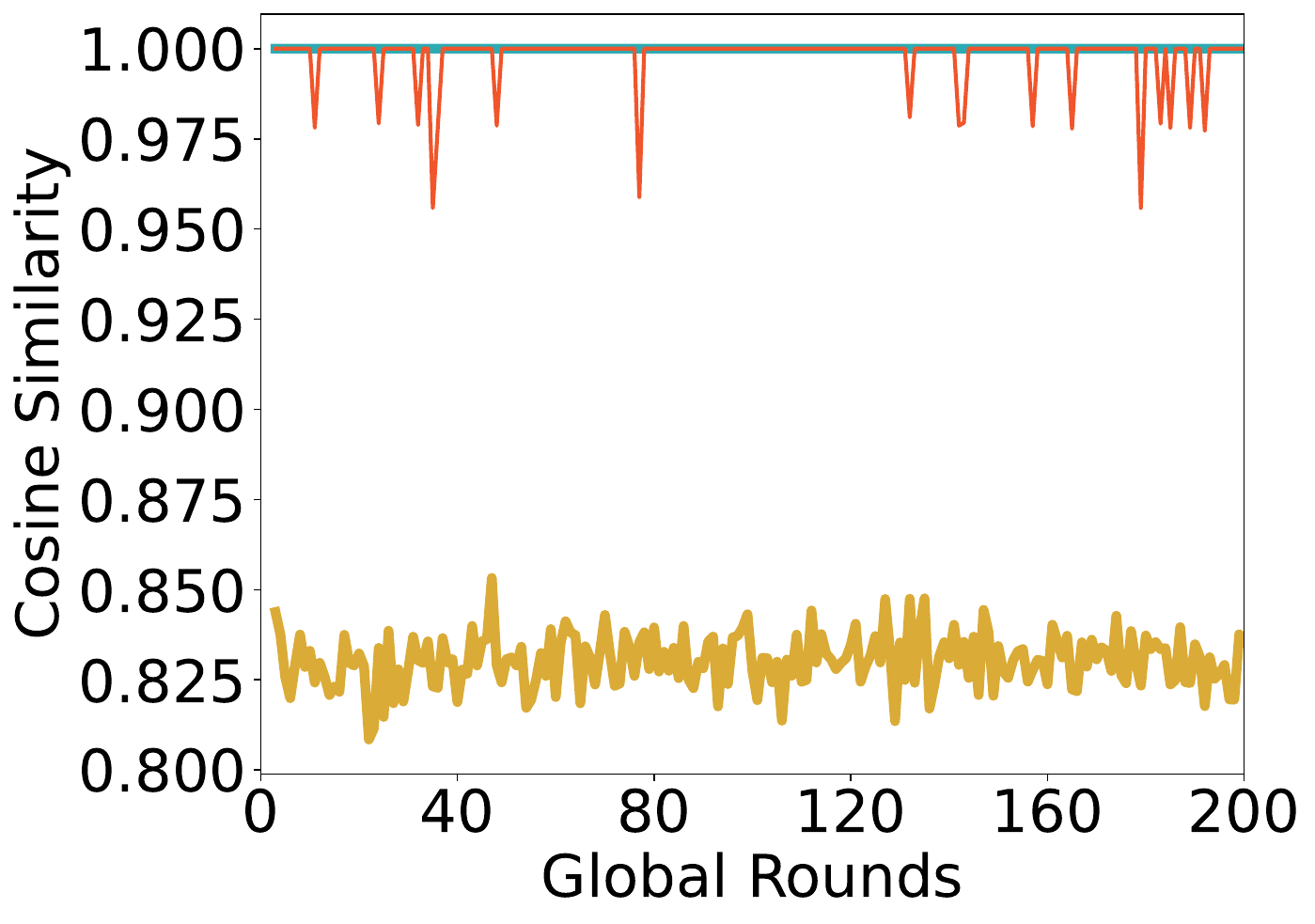}}
	\\
	\subfloat[CIFAR10,80,80\%,15]{
		\includegraphics[scale=0.18]{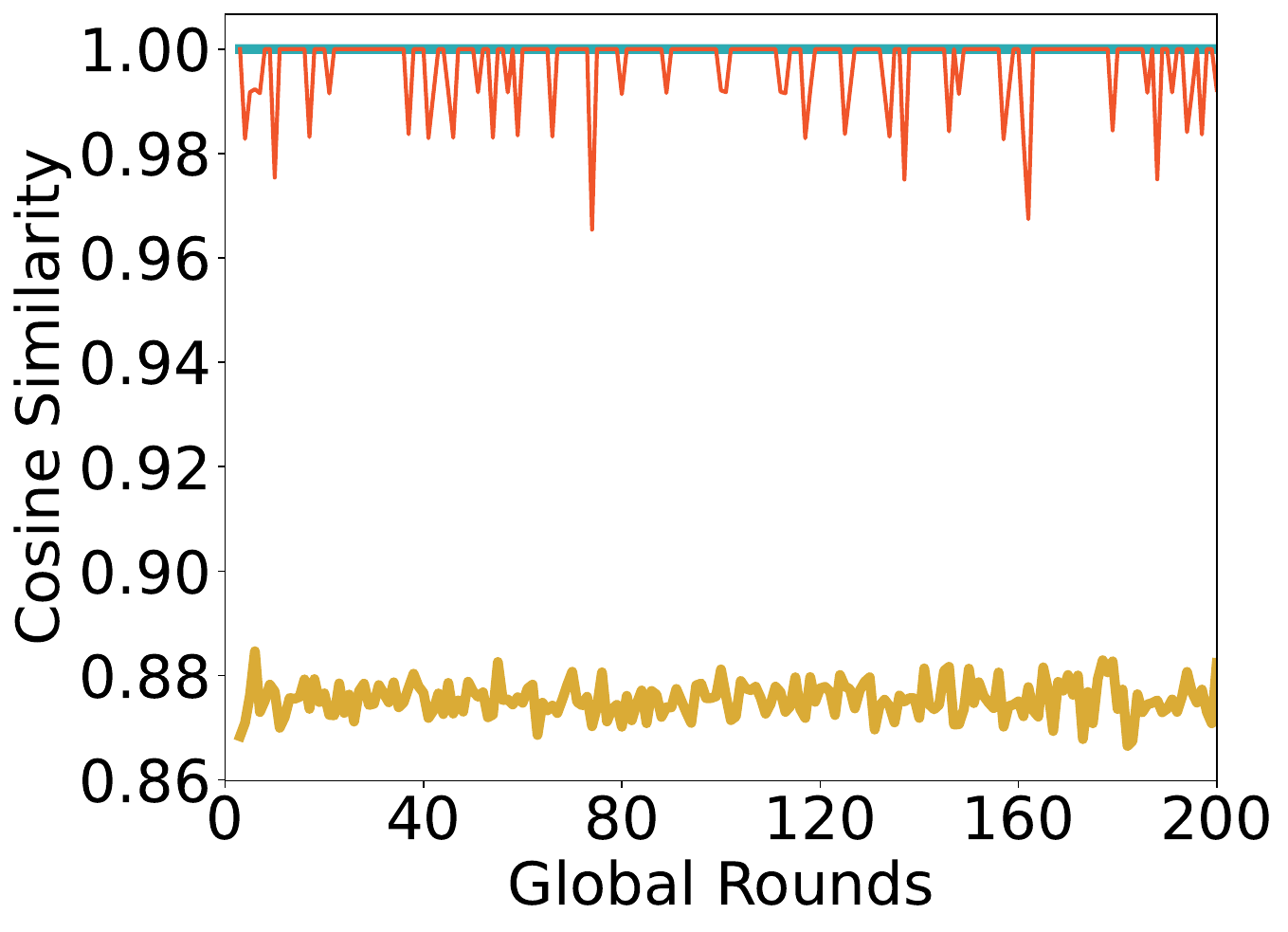}}
	\subfloat[CIFAR10,90,90\%,8]{
		\includegraphics[scale=0.18]{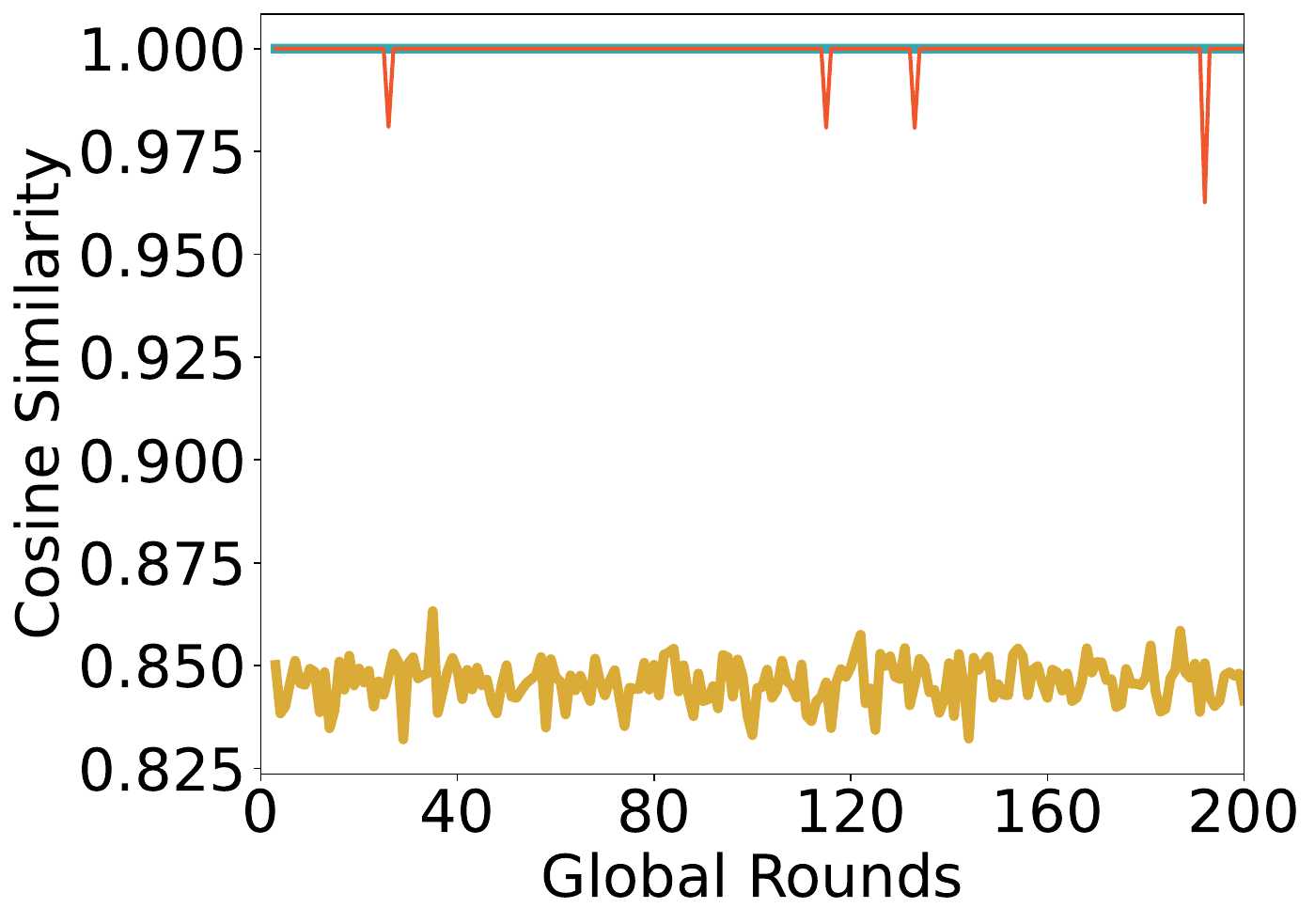}}
	\caption{The performance of VERT in predicting user gradients in the face of large-scale ALIE attacks in non-IID scenarios.}
\label{figure10}
\end{figure}

Figure~\ref{figure4}-\ref{figure10} shows the Prediction performance of VERT for  different model poisoning attacks on different datasets.

\subsection{Defense Effectiveness}\label{appendixd}

TABLE~\ref{table4} shows the defense effectiveness of different defenses against large-scale model poisoning attacks in non-IID scenarios.

\begin{table*}[h]
	\centering
	\caption{The defense effectiveness of different defenses against large-scale model poisoning attacks in non-IID scenarios.}
	\begin{tabular}{|c|c|c|c|c|c|c|}
		\toprule
		\textbf{Methods} & \textbf{FedAvg} & \textbf{Krum} & \textbf{Median} & \textbf{FLDetector} & \textbf{VERT+Krum} & \textbf{VERT+FedAvg} \\
		\midrule
		\multicolumn{7}{|c|}{MNIST, $|C_t|$ = 80, 90, $pr$ = 80\%, 90\%} \\
		\midrule
		\textbf{GN} & 15.53\%, 14.29\% & 15.21\%, \colorbox{color2}{32.08\%} & \colorbox{color2}{35.63\%}, 24.20\% & 12.72\%, 11.68\% & 16.50\%, 19.90\% & \colorbox{color1}{74.97\%}, \colorbox{color1}{74.10\%} \\
		\midrule
		\textbf{MR} & 10.00\%, 9.80\% & \colorbox{color2}{13.45\%}, 14.38\% & 11.86\%, 9.80\% & 10.00\%, 10.00\% & 11.36\%, \colorbox{color2}{17.30\%} & \colorbox{color1}{38.24\%}, \colorbox{color1}{63.84\%} \\
		\midrule
		\textbf{AGR} & 10.00\%, 10.00\% & 10.00\%, \colorbox{color2}{10.58\%}  & 9.80\%, 9.80\% & 9.80\%, 9.80\% & \colorbox{color1}{18.92\%}, \colorbox{color1}{17.67\%} & \colorbox{color2}{10.39\%}, 10.37\% \\
		\midrule
		\textbf{ALIE} & 10.00\%, 10.00\% & 10.00\%, 10.00\% & 10.00\%, 10.00\% & 9.80\%, 9.80\% & \colorbox{color1}{11.35\%}, \colorbox{color1}{15.39\%}  & \colorbox{color2}{11.35\%}, \colorbox{color2}{13.61\%} \\
		\midrule
		\multicolumn{7}{|c|}{CIFAR10, $|C_t|$ = 80, 90, $pr$ = 80\%, 90\%} \\
		\midrule
		\textbf{GN} & 13.38\%, 15.21\% & 12.36\%, 15.94\% & \colorbox{color2}{17.17\%}, \colorbox{color2}{20.14\%} & 12.06\%, 13.40\% & 11.12\%, 18.56\% & \colorbox{color1}{33.62\%}, \colorbox{color1}{30.38\%} \\
		\midrule
		\textbf{MR} & \colorbox{color1}{27.86\%}, \colorbox{color2}{21.24\%} & 12.38\%, 12.50\% & 15.86\%, 12.70\% & 9.57\%, 11.25\% & 14.23\%, 13.92\% & \colorbox{color2}{23.96\%}, \colorbox{color1}{24.54\%} \\
		\midrule
		\textbf{AGR} & 10.00\%, 10.00\% & 10.00\%, 10.00\% & 10.00\%, 10.00\% & 10.00\%, 10.00\% & \colorbox{color1}{10.68\%}, \colorbox{color1}{17.30\%} & \colorbox{color2}{10.68\%}, \colorbox{color2}{14.51\%} \\
		\midrule
		\textbf{ALIE}  & 10.00\%, 10.00\% & 10.00\%, 10.00\% & 10.00\%, 10.00\% & 10.00\%, 10.00\% & \colorbox{color1}{16.66\%}, \colorbox{color1}{17.12\%} & \colorbox{color2}{10.67\%}, \colorbox{color2}{11.88\%} \\

		\bottomrule
	\end{tabular}
	\begin{tablenotes}
		\footnotesize
		\item[1] Note: The data in each cell is the highest global model accuracy, on the left is the result of pr=80\%, and the right is 90\%.
	\end{tablenotes}
	\label{table4}
\end{table*}

\end{document}